%% file: main.tex
\documentclass[lettersize,journal]{IEEEtran}
\usepackage{amsmath,amsfonts}
\usepackage[T1]{fontenc}

\usepackage{algorithm}
\usepackage{array}
\usepackage{textcomp}
\usepackage{stfloats}
\usepackage{url}
\usepackage{verbatim}
\usepackage{graphicx}
\usepackage{cite}

\usepackage{relsize}
\usepackage{hyperref} 
\usepackage{multirow} 
\usepackage{algpseudocode}
\usepackage{caption}
\usepackage{subcaption}
\usepackage{float}
\usepackage[,font=normalsize]{subcaption}

\usepackage{booktabs}
\usepackage{arydshln}
\usepackage{makecell}

\begin{document}

\title{SpectralEarth: Training Hyperspectral Foundation Models at Scale}

\author{%
Nassim Ait Ali Braham,
Conrad M. Albrecht,
Julien Mairal,
Jocelyn Chanussot,
Yi Wang,
Xiao Xiang Zhu%
\thanks{Nassim Ait Ali Braham (corresponding author, e-mail:
        Nassim.AitAliBraham@dlr.de), Yi Wang and Xiao Xiang Zhu are with the
        Chair of Data Science in Earth Observation (SiPEO), Technical
        University of Munich (TUM), 80333 Munich, Germany.}%
\thanks{Conrad M. Albrecht and Nassim Ait Ali Braham are with the Remote
        Sensing Technology Institute (IMF), German Aerospace Center (DLR),
        82234 Wessling, Germany.}%
\thanks{Julien Mairal and Jocelyn Chanussot are with Univ.\ Grenoble
        Alpes, Inria, CNRS, Grenoble INP, LJK, 38000 Grenoble, France.}%
}

\markboth{Journal of \LaTeX\ Class Files,~Vol.~14, No.~8, August~2021}%
{Shell \MakeLowercase{\textit{et al.}}: A Sample Article Using IEEEtran.cls for IEEE Journals}


\maketitle

\renewcommand{\thefootnote}{}
\footnotetext{This is the accepted manuscript version of the article published in \textit{IEEE Journal of Selected Topics in Applied Earth Observations and Remote Sensing}. The final version is available at: \url{https://doi.org/10.1109/JSTARS.2025.3581451}}
\renewcommand{\thefootnote}{\arabic{footnote}}

\input{sec/0_abstract}    
\input{sec/1_intro}

\input{sec/2_related_work}

\input{sec/3_datasets}

\input{sec/4_models}

\input{sec/5_experimental_setup}
\input{sec/6_experimental_results}

\input{sec/7_conclusion}

\section{Acknowledgement}
The work of N.\ Ait Ali Braham was supported by the European Commission through the project ``EvoLand'' under the Horizon 2020 Research and Innovation program (Grant Agreement No.\ \texttt{101082130}). The work of Y.\ Wang and C.\ Albrecht was funded by the Helmholtz Association through the Framework of Helmholtz AI, grant ID: \texttt{ZT-I-PF-5-01} --- Local Unit Munich Unit Aeronautics, Space and Transport (MASTr). This work was partially supported by ANR 3IA MIAI\texttt{@}Grenoble Alpes (ANR-19-P3IA-0003), and J.\ Mairal is supported by ERC grant number 101087696 (APHELEIA project). 
The compute was supported by the Helmholtz Association’s Initiative and Networking Fund on the HAICORE\texttt{@}FZJ partition. The authors gratefully acknowledge the computational and data resources provided through the joint high-performance data analytics (HPDA) project “terrabyte” of the German Aerospace Center (DLR) and the Leibniz Supercomputing Center (LRZ).

\newpage

{
    \bibliographystyle{IEEEtran}
    \bibliography{main}
    
}

\begin{IEEEbiography}[{\includegraphics[width=1in,height=1.25in,clip,keepaspectratio]{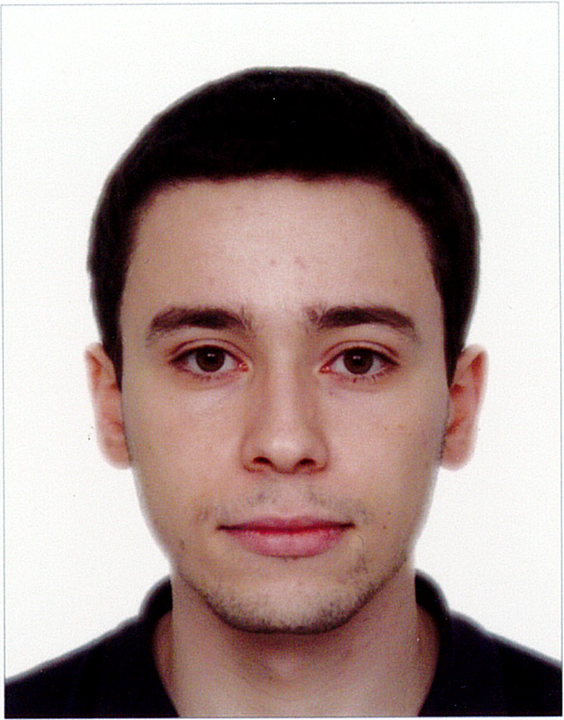}}]{Nassim Ait Ali Braham} received his M.Sc. degree in computer science from Ecole nationale Supérieure d'Informatique (ESI), Algiers, Algeria, in 2019 and a M.Sc. degree in Artificial Intelligence and Data Science from Université Paris Dauphine-PSL, Paris, France, in 2020. He is pursuing a Ph.D. degree at the German Aerospace Center, Wessling, Germany, and at the Technical University of Munich, Munich, Germany. In 2019, he spent six months as a research intern at the LIRIS-CNRS laboratory, Lyon, France. In 2020, he spent six months at the LAMSADE-CNRS laboratory, PSL Research University, Paris, France. In 2023, he spent six months as a visiting researcher at INRIA THOTH, Grenoble, France. His research interests include deep learning, computer vision, self-supervised learning and remote sensing.
\end{IEEEbiography}
\begin{IEEEbiography}[{\includegraphics[width=1in,height=1.25in,clip]{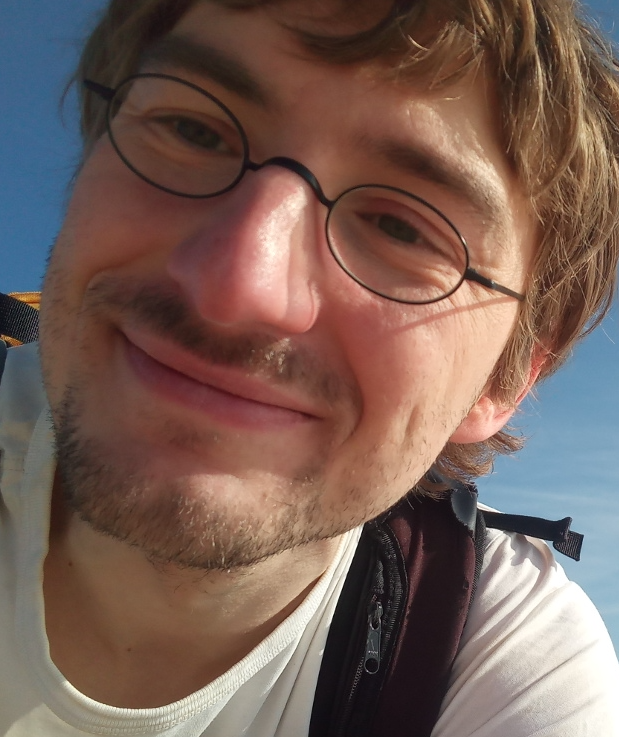}}] {Conrad M Albrecht} (M'17) received his PhD in physics from Heidelberg University, Germany, in 2014. Since April 2021 he is a researcher at the Earth Observation Center of the German Aerospace Center (DLR), Oberpfaffenhofen, Germany. As PI of the HelmholtzAI Young Investigator Group (YIG) ``Large-Scale Data Mining in Earth Observation'' (DM4EO) he advanced weakly-supervised deep learning methodologies for remote sensing. In July 2023, he was appointed a visiting associate professor with the Institute of Nasca at Yamagata University, Yamagata, Japan contributing to research in machine learning for the UNESCO World Heritage of the Nasca culture in Peru. Jointly with his collaborators, his work has been awarded the 2024 Cozzarelli Prize by the National Academy of Sciences.

For over 6 years Conrad was a research scientist in the Physical Sciences department at the IBM T.J.\ Watson Research Center in Yorktown, NY, USA. While at the Institute for Theoretical Physics, Heidelberg, Germany, he graduated in physics (International Max-Planck Research School for Quantum Dynamics in Physics, Chemistry and Biology) with an extra certification in computer science (Cluster- \& Detector Management team at CERN, Switzerland). He received a corresponding Ph.D. degree from Heidelberg University, Germany in 2014 working on distributed computing to study physics at low temperatures.

Conrad's research agenda interconnect physical models and numerical analysis, employing Big Data technologies and machine learning through open-science research, \url{https://conrad-m-albrecht.github.io}. Conrad co-organized workshops at the IEEE BigData conference, IGARSS conferences, the EGU General Assembly, the CVPR conference, and the AAAS annual meeting. Conrad is home in Europe and the United States. Some of his transatlantic initiatives include: In 2023, he was invited by the Alexander-von-Humboldt Foundation to present at the German-American Frontiers of Engineering Symposium. Beginning in 2024, and initiated by Helmholtz Imaging and Data Science, he is in close touch with the German Department at Princeton University for cross-atlantic undergraduate internships. In fall 2024, Conrad taught as Adjunct Professor of Earth and Environmental Engineering at Columbia University, NY, USA, and in spring 2025 he supervises Columbia University students for environmental monitoring.
\end{IEEEbiography}%
\begin{IEEEbiography}[{\includegraphics[width=1in,height=1.25in,clip,keepaspectratio]{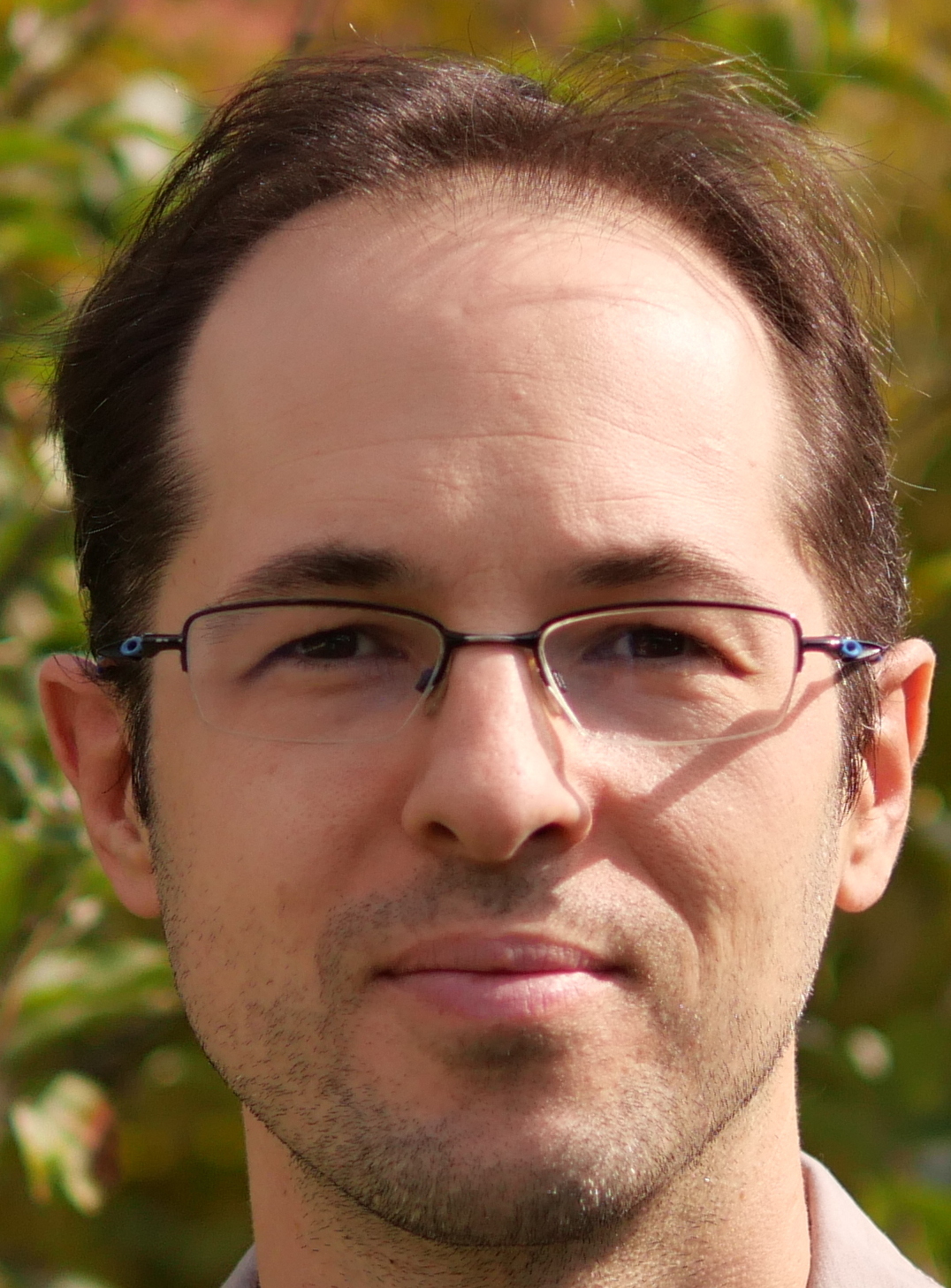}}]{Julien Mairal} received the graduate degree from Ecole Polytechnique, Palaiseau, France, in 2005, and the Ph.D. degree from Ecole Normale Superieure, Cachan, France, in 2010. After that, he joined the statistics department at UC Berkeley as a post-doctoral researcher. In 2012, he joined Inria, Grenoble, France, where he is currently a research director and head of the Thoth team. His research interests include machine learning, computer vision, mathematical optimization, and statistical image and signal processing. He received a starting grant and a consolidator grant from the European Research Council, respectively in 2016 and 2022. He was awarded the Cor Baayen prize in 2013, the IEEE PAMI young research award in 2017 and the test-of-time award at ICML 2019.
\end{IEEEbiography}
\begin{IEEEbiography}
[{\includegraphics[width=1in,height=1.25in,clip,keepaspectratio]{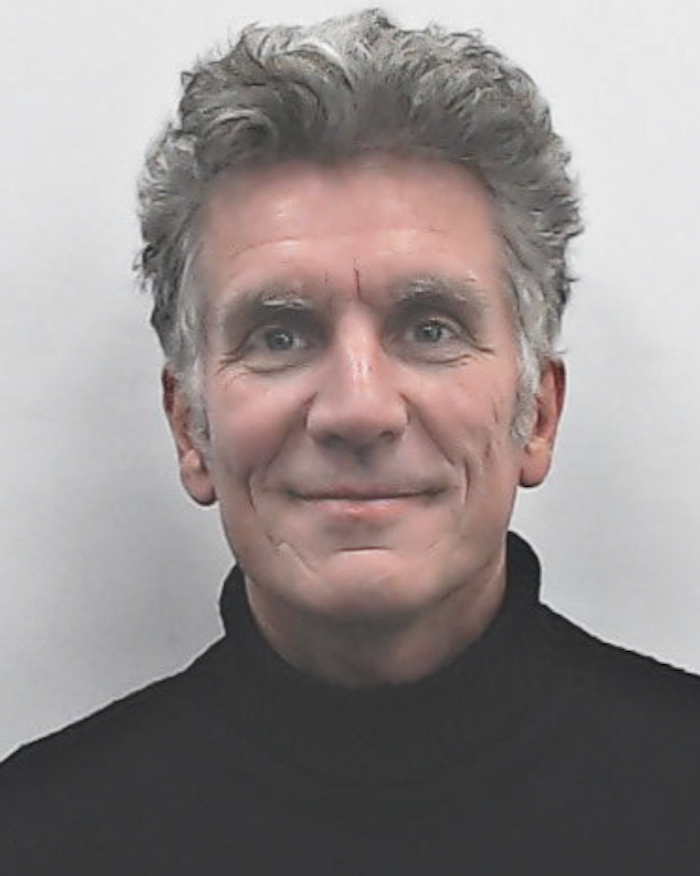}}]
{Jocelyn Chanussot} (M’04–SM’04–F’12) received the M.Sc. degree in electrical engineering from the Grenoble Institute of Technology (Grenoble INP), Grenoble, France, in 1995, and the Ph.D. degree from the Université de Savoie, Annecy, France, in 1998. From 1999 to 2023, he has been with Grenoble INP, where he was a Professor of signal and image processing. He is currently a Research Director with INRIA, Grenoble. His research interests include image analysis, hyperspectral remote sensing, data fusion, machine learning and artificial intelligence. He has been a visiting scholar at Stanford University (USA), KTH (Sweden) and NUS (Singapore). Since 2013, he is an Adjunct Professor of the University of Iceland. In 2015-2017, he was a visiting professor at the University of California, Los Angeles (UCLA).  He holds the AXA chair in remote sensing and is an Adjunct professor at the Chinese Academy of Sciences, Aerospace Information research Institute, Beijing.
Dr. Chanussot is the founding President of IEEE Geoscience and Remote Sensing French chapter (2007-2010) which received the 2010 IEEE GRS-S Chapter Excellence Award. He has received multiple outstanding paper awards. He was the Vice-President of the IEEE Geoscience and Remote Sensing Society, in charge of meetings and symposia (2017-2019). He was the General Chair of the first IEEE GRSS Workshop on Hyperspectral Image and Signal Processing, Evolution in Remote sensing (WHISPERS). He was the Chair (2009-2011) and  Cochair of the GRS Data Fusion Technical Committee (2005-2008). He was a member of the Machine Learning for Signal Processing Technical Committee of the IEEE Signal Processing Society (2006-2008) and the Program Chair of the IEEE International Workshop on Machine Learning for Signal Processing (2009). He is an Associate Editor for the IEEE Transactions on Geoscience and Remote Sensing, the IEEE Transactions on Image Processing and the Proceedings of the IEEE. He was the Editor-in-Chief of the IEEE Journal of Selected Topics in Applied Earth Observations and Remote Sensing (2011-2015). In 2014 he served as a Guest Editor for the IEEE Signal Processing Magazine. He is a Fellow of the IEEE, an ELLIS Fellow, a Fellow of the Asia-Pacific Artificial Intelligence Association, a member of the Institut Universitaire de France (2012-2017) and a Highly Cited Researcher (Clarivate Analytics/Thomson Reuters, since 2018).
\end{IEEEbiography}
\begin{IEEEbiography}
[{\includegraphics[width=1in,height=1.25in,clip,keepaspectratio]{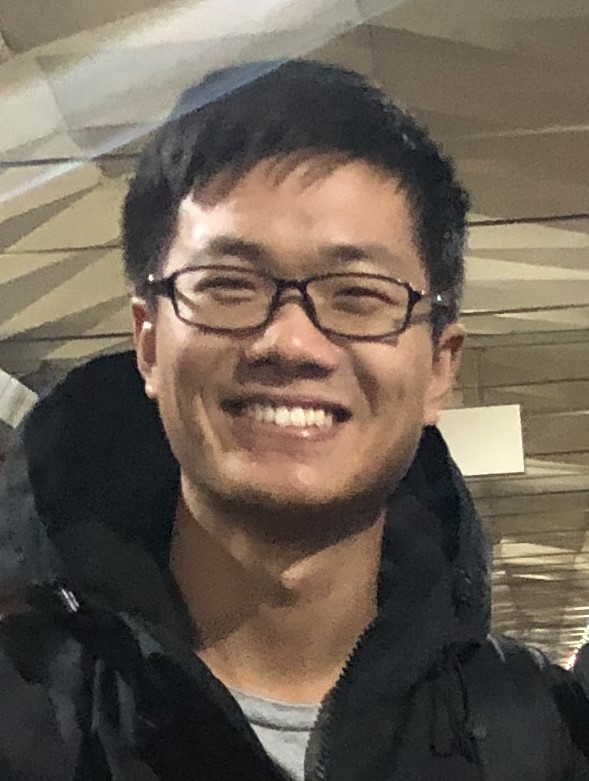}}] {Yi Wang} (S'21) received his B.E. degree in remote sensing science and technology from Wuhan University, Wuhan, China, in 2018 and his M.Sc. degree in geomatics engineering from University of Stuttgart, Stuttgart, Germany, in 2021. He is pursuing his Ph.D. degree at the Technical University of Munich (TUM), Munich, Germany. From 2021 to 2024, he was a research associate at the Remote Sensing Technology Institute, German Aerospace Center (DLR). In 2020, he spent three months at the perception system group, Sony Corporation, Stuttgart, Germany. His research interests include self-supervised learning, weakly-supervised learning, and multimodal representations. 
\end{IEEEbiography}
\begin{IEEEbiography}[{\includegraphics[width=1in,height=1.25in,clip,keepaspectratio]{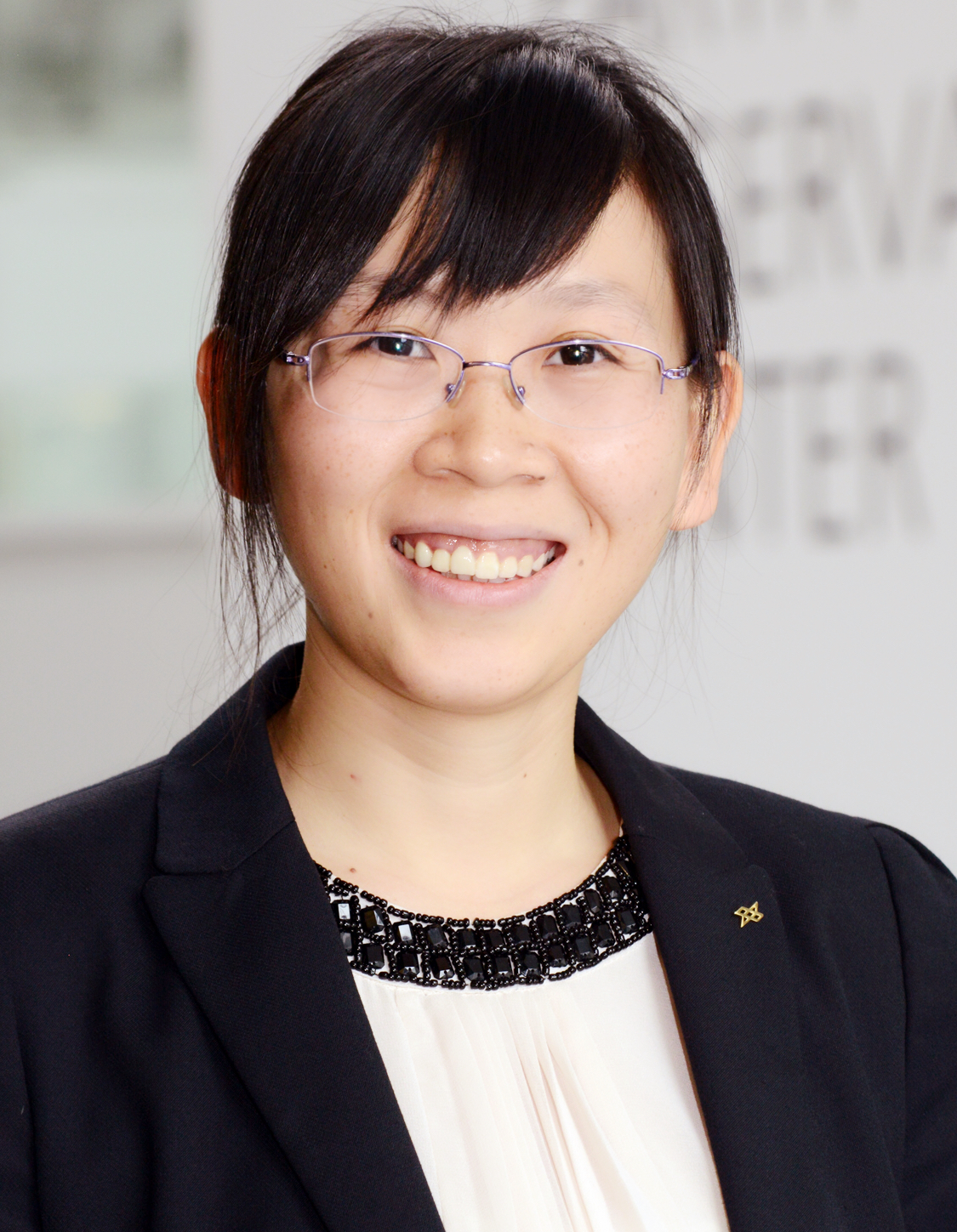}}]{Xiao Xiang Zhu}(S'10--M'12--SM'14--F'21) received the Master (M.Sc.) degree, her doctor of engineering (Dr.-Ing.) degree and her “Habilitation” in the field of signal processing from Technical University of Munich (TUM), Munich, Germany, in 2008, 2011 and 2013, respectively.
\par
She is the Chair Professor for Data Science in Earth Observation at Technical University of Munich (TUM) and was the founding Head of the Department ``EO Data Science'' at the Remote Sensing Technology Institute, German Aerospace Center (DLR). Since May 2020, she is the PI and director of the international future AI lab "AI4EO -- Artificial Intelligence for Earth Observation: Reasoning, Uncertainties, Ethics and Beyond", Munich, Germany. Since October 2020, she also serves as a Director of the Munich Data Science Institute (MDSI), TUM. From 2019 to 2022, Zhu has been a co-coordinator of the Munich Data Science Research School (www.mu-ds.de) and the head of the Helmholtz Artificial Intelligence -- Research Field ``Aeronautics, Space and Transport".  Prof. Zhu was a guest scientist or visiting professor at the Italian National Research Council (CNR-IREA), Naples, Italy, Fudan University, Shanghai, China, the University  of Tokyo, Tokyo, Japan and University of California, Los Angeles, United States in 2009, 2014, 2015 and 2016, respectively. She is currently a visiting AI professor at ESA's Phi-lab, Frascati, Italy. Her main research interests are remote sensing and Earth observation, signal processing, machine learning and data science, with their applications in tackling societal grand challenges, e.g. Global Urbanization, UN’s SDGs and Climate Change.

Dr. Zhu has been a member of young academy (Junge Akademie/Junges Kolleg) at the Berlin-Brandenburg Academy of Sciences and Humanities and the German National  Academy of Sciences Leopoldina and the Bavarian Academy of Sciences and Humanities. She is a Fellow of the Academia Europaea (the Academy of Europe). She serves in the scientific advisory board in several research organizations, among others the German Research Center for Geosciences (GFZ, 2020-2023) and Potsdam Institute for Climate Impact Research (PIK). She is an associate Editor of IEEE Transactions on Geoscience and Remote Sensing, Pattern Recognition and served as the area editor responsible for special issues of IEEE Signal Processing Magazine (2021-2023). She is a Fellow of IEEE, AAIA, and ELLIS.
\end{IEEEbiography}

\end{document}

%% file: sec/0_abstract.tex
\begin{abstract}
Foundation models have triggered a paradigm shift in computer vision and are increasingly being adopted in remote sensing, particularly for multispectral imagery. Yet, their potential in hyperspectral imaging (HSI) remains untapped due to the absence of comprehensive and globally representative hyperspectral datasets. To close this gap, we introduce \textit{SpectralEarth}, a large-scale multi-temporal dataset designed to pretrain hyperspectral foundation models leveraging data from the Environmental Mapping and Analysis Program (EnMAP). SpectralEarth comprises 538,974 image patches covering 415,153 unique locations from 11,636 globally distributed EnMAP scenes spanning two years of archive. Additionally, 17.5\% of these locations include multiple timestamps, enabling multi-temporal HSI analysis. Utilizing state-of-the-art self-supervised learning (SSL) algorithms, we pretrain a series of foundation models on SpectralEarth, integrating a spectral adapter into classical vision backbones to accommodate the unique characteristics of HSI. In tandem, we construct nine downstream datasets for land-cover, crop-type mapping, and tree-species classification, providing benchmarks for model evaluation. Experimental results support the versatility of our models and their generalizability across different tasks and sensors. We also highlight computational efficiency during model fine-tuning. The dataset, pretrained models, and code are publicly available.\footnote{Dataset: \url{https://geoservice.dlr.de/web/datasets/enmap_spectralearth}, Code: \url{https://github.com/AABNassim/spectral_earth}}
\end{abstract}

%% file: sec/1_intro.tex
\section{Introduction}
\label{sec:intro}
      Hyperspectral imaging (HSI) from space provides valuable information about the material composition of the Earth's surface and atmosphere. With hundreds of narrow bands, each a few nanometers in width, hyperspectral images record detailed electromagnetic information across a wide range of wavelengths, from long-wave ultraviolet to short-wave infrared~\cite{khan2018modern}. This rich spectral information provides opportunities for numerous environmental applications such as soil and mineral mapping, pollution tracking, agricultural assessment, and forest monitoring~\cite{chabrillat2019imaging, van2012multi, jia2020status, lu2020recent, upadhyay2018hyperspectral}.
    \begin{figure}
        \centering
        \includegraphics[width=\linewidth]{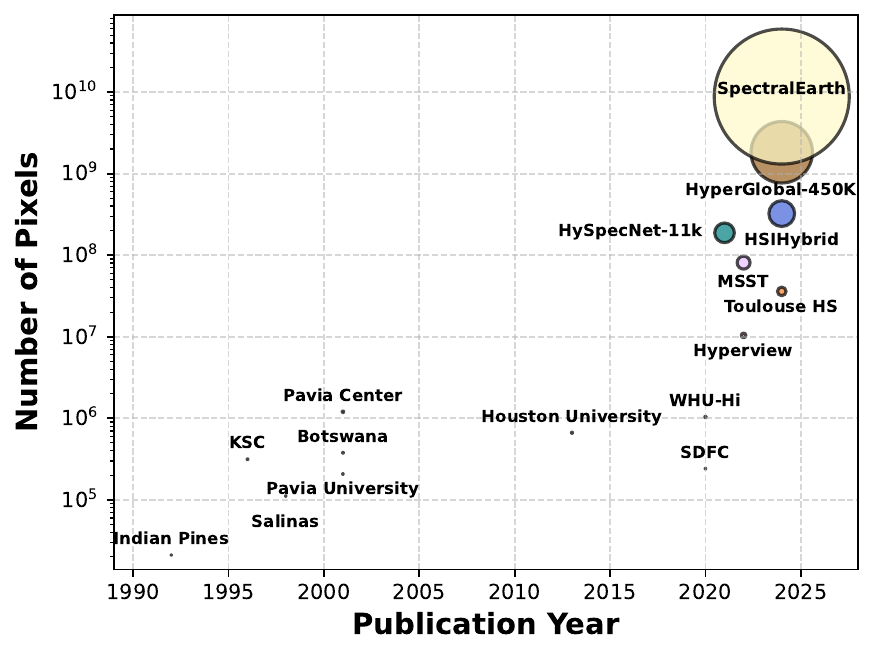}
        \caption{Visualization of scale for various hyperspectral datasets published within the past three decades illustrating the volume of the SpectralEarth dataset (area of circles).}
        \label{fig:dataset-scale}
    \end{figure}
    In recent years, the availability of hyperspectral data has been significantly expanded by the launch of new satellite missions. Notable examples include Germany’s Environmental Mapping and Analysis Program (EnMAP)~\cite{kaufmann2012science} with precursor DLR Earth Sensing Imaging Spectrometer mission (DESIS)~\cite{kerr2016hyperspectral} mounted to the International Space Station (ISS), Italy’s Hyperspectral Precursor and Application Mission (PRISMA)~\cite{pignatti2013prisma}, and the European Space Agency’s upcoming Copernicus Hyperspectral Imaging Mission for the Environment (CHIME)~\cite{rast2021copernicus}. These developments open new avenues to employ deep learning and Self-Supervised Learning (SSL) for large-scale HSI analysis.
    \begin{figure*}[]
        \centering
        \includegraphics[width=0.9\textwidth]{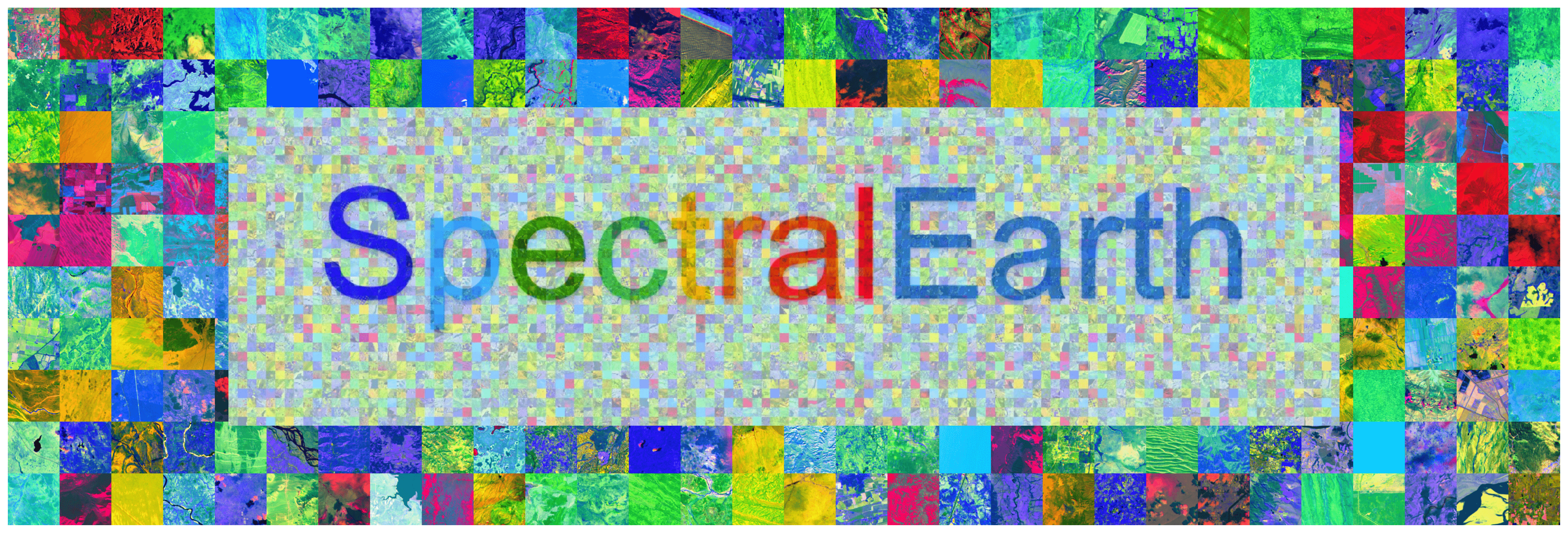}
        \caption{Mosaic of pseudo-RGB representatives from the SpectralEarth dataset to showcase various landscapes, including urban areas, agricultural land, deserts, forests, and water bodies. Each image is presented with a false-color composite using the first three principal components.}
        \label{fig:spectral_earth_main_fig}
    \end{figure*}%
    
    Foundation models pretrained with SSL demonstrated remarkable generalization capabilities with minimal fine-tuning in computer vision~\cite{oquab2023dinov2}. Subsequently, there has been a surge in developing geospatial foundation models, particularly for multispectral and high-resolution RGB imagery~\cite{wang2022self}. This line of research benefits from the abundance of publicly available satellite data from missions like Copernicus Sentinel-2, which provides petabytes of archive data for model training. In contrast, progress in foundation models for the hyperspectral domain has been hindered by the lack of large-scale HSI datasets for pre-training. Historically, access to hyperspectral imagery has been costly, with most benchmarks restricted to small-scale aerial imagery datasets~\cite{audebert2019deep}. Although recent efforts~\cite{fuchs2023hyspecnet,scheibenreif2023masked,wang2024hypersigmahyperspectralintelligencecomprehension} have begun to address this limitation, they still fall short in terms of data volume and geographic diversity to effectively pre-train general-purpose hyperspectral models.
    
    To bridge this gap, we introduce \textit{SpectralEarth}, a large-scale dataset derived from the EnMAP satellite mission, featuring over 538,974 hyperspectral image patches from 415,153 unique locations with global spatial distribution and cloud coverage below 10\%. SpectralEarth, as illustrated in Figure~\ref{fig:dataset-scale}, represents an important leap in scale, being significantly larger than existing HSI datasets, and spans a wide variety of landscapes to reflect the full diversity of spectral signatures. Compiled from 11,636 EnMAP scenes, the dataset volume exceeds 3TB in hyperspectral images. Notably, about 17.5\% of its geospatial locations include time series data to enable multi-temporal HSI analysis. To harness the rich information in hyperspectral data, we modify conventional vision backbones such as ResNet~\cite{he2016deep} and ViT~\cite{dosovitskiy2020image} with a spectral adapter module. This enables such backbones to capture the unique spectral characteristics of HSI. Using three popular SSL algorithms, MoCo-V2~\cite{he2020momentum, chen2020improved}, DINO~\cite{caron2021emerging} and MAE~\cite{he2022masked}, we train these backbones on SpectralEarth, to provide a plurality of pretrained models for hyperspectral image analysis. To benchmark the efficacy of our models, we introduce nine downstream datasets constructed by pairing EnMAP, DESIS, and Hyperion EO-1 imagery with land cover, crop type, and tree species labels.
    Our experiments demonstrate the potential of large-scale pre-training to efficiently generalize across various hyperspectral imaging contexts, including adapting to data from alternative sensors with different spectral characteristics. Our contributions are summarized as follows:
    \begin{itemize}
        \item Assembly of \textit{SpectralEarth}, a large-scale dataset comprising around 3.3TB of nearly cloud-free EnMAP hyperspectral images. SpectralEarth features a global geospatial distribution encompassing over 538,974 image patches from 415,153 unique geo-locations, 73,307 including multiple timestamps.
        \item Construction of nine downstream datasets designed to benchmark classification and semantic segmentation tasks for HSI.
        \item An empirical evaluation of self-supervised learning in hyperspectral imaging, by pre-training various models on SpectralEarth and evaluating their generalizability across downstream tasks and sensors. In addition, we demonstrate the models' computational benefit through faster convergence in fine-tuning.
    \end{itemize}

%% file: sec/2_related_work.tex
\section{Related Work}
    \begin{figure}
      \centering
      \includegraphics[width=\linewidth]{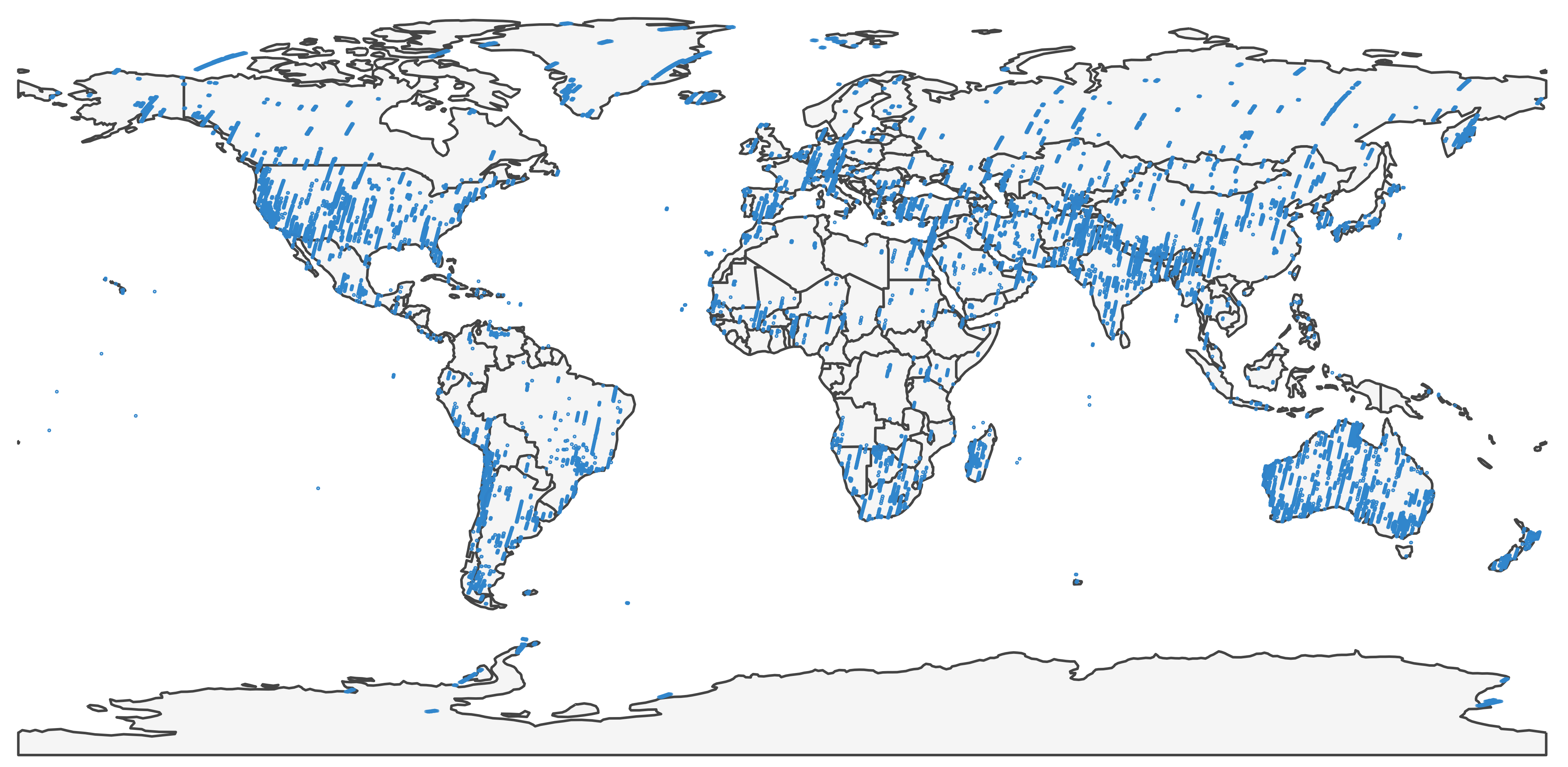}
      \caption{Geographical Distribution of SpectralEarth. The map depicts the global coverage of hyperspectral images within our dataset, demonstrating its extensive geographical scope.}
      \label{fig:spectral_earth_map}
    \end{figure}

\subsection{Hyperspectral Datasets}
    \begin{table*}[t!]
        \centering
        \small 
        \setlength{\tabcolsep}{3pt} 
        \begin{tabular}{@{}lcccccc@{}} 
            \toprule
            \textbf{Dataset} & \textbf{\# Img.} & \textbf{Size} & \textbf{Sensor} & \textbf{Bands} & \textbf{GSD} & \textbf{Multi-temporal} \\
            \midrule
            Indian Pines~\cite{baumgardner2015220}      & 1      & 145$\times$145  & AVIRIS        & 224   & 20m & $\times$ \\
            KSC~\cite{Hyperspectral}         & 1      & 512$\times$614  & AVIRIS        & 224   & 18m & $\times$ \\    
            Salinas~\cite{Hyperspectral}           & 1      & 512$\times$217  & AVIRIS        & 224   & 3.7m & $\times$ \\
            Pavia University~\cite{Hyperspectral}        & 1      & 610$\times$340  & ROSIS         & 103   & 1.3m & $\times$ \\
            Pavia Center~\cite{Hyperspectral}      & 1      & 1096$\times$1096 & ROSIS        & 102   & 1.3m & $\times$ \\
            Botswana~\cite{Hyperspectral}          & 1      & 1476$\times$256 & EO-1 & 242   & 30m & $\times$ \\
            Houston University~\cite{6776408}     & 1      & 349$\times$1905 & ITRES-CASI    & 144   & 2.5m & $\times$ \\
            SDFC~\cite{sun2023sdfc}           & 1      & 1201$\times$201 & HAHS          & 63    & 0.5m & $\times$ \\
            Chikusei~\cite{yokoya2016airborne} & 1 & 2517$\times$2335 & Hyperspec-VNIR-C & 128 & 2.5m & $\times$ \\
            WHU-Hi~\cite{zhong2020whu} & 3 & 1217$\times$303, 940$\times$475, 550$\times$400 & Nano-Hyperspec & 270-274 & 0.043-0.463m & $\times$ \\
            Hyperview~\cite{nalepa2022hyperview}         & $\sim$3k & avg. $\sim$60$\times$60 & HySpex    & 150   & 2m & $\times$ \\
            Toulouse HS~\cite{thoreau2024toulouse}   & 9  & $\sim$4000$\times$1000  & HySpex  & 310  & 1m  & $\checkmark$ \\

            \hdashline
            HySpecNet-11k~\cite{fuchs2023hyspecnet}         & $\sim$11k    & 128$\times$128  & EnMAP         & 202   & 30m & $\times$ \\
            MSST~\cite{scheibenreif2023masked}              & $\sim$20k    & 64$\times$64    & EnMAP         & 200   & 30m & $\times$ \\
            HSIHybrid~\cite{wang2024hsimae}              & $\sim$4m    & 9$\times$9    & Multiple         & Variable   & Variable & $\times$ \\
            HyperGlobal-450K~\cite{wang2024hypersigmahyperspectralintelligencecomprehension}  & $\sim$447k   & 64$\times$64    & EO-1 and Gaofen-5B & 242 and 330 & 30m & $\times$ \\
            SpectralEarth     & $\sim$538k   & 128$\times$128  & EnMAP         & 202   & 30m & $\checkmark$ \\
            \bottomrule
        \end{tabular}
        \caption{\textit{Summary of Hyperspectral Remote Sensing Datasets}: SpectralEarth contains nearly nine gigapixels of 202 EnMAP bands each at 30m spatial resolution. This equates to an area of a little over six million square kilometers. The rows above the dashed line concern manually annotated datasets, while the part below lists unlabeled datasets for SSL.}
        \label{tab:dataset_comparison}
    \end{table*}
    Traditionally, hyperspectral benchmark datasets have been limited in geographical coverage, often confined to a single or a limited number of scenes~\cite{hyperspectral_scenes_ehu,audebert2019deep,nalepa2022hyperview,hong2023cross}. This limitation is primarily rooted in the expenses associated with airborne hyperspectral sensor surveys on the one hand and the collection of ground truth annotations and in-situ measurements on the other hand. While collecting quality labels remains a challenge today, the increased accessibility to more cost-effective sensors and the launch of hyperspectral satellite missions enabled coverage of larger regions with HSI. Developments in SSL and foundation models have also driven interest in creating large-scale unlabeled HSI datasets. For instance, the HySpecNet-11k~\cite{fuchs2023hyspecnet} collected about 11,500 unlabeled hyperspectral patches from the EnMAP satellite. HySpecNet-11k is designed to benchmark (unsupervised) image compression techniques. Similarly, MSST~\cite{scheibenreif2023masked} compiled a dataset of 19,792 EnMAP patches for SSL pre-training. While HySpecNet-11k and MSST have contributed to increasing the scale of existing HSI datasets, they remain limited in size to train foundation models. The recently introduced HyperGlobal-450K dataset~\cite{wang2024hypersigmahyperspectralintelligencecomprehension} encompasses more than 450,000 HSI patches derived from 486 Hyperion EO-1 and 215 Gaofen-5B scenes. While this dataset significantly increases the scale of publicly available HSI data, it presents certain limitations. First, its geographical distribution is less diverse, with a substantial part of the data concentrated in a few provinces of China. Moreover, it relies on imagery from the decommissioned EO-1 satellite. In contrast, SpectralEarth offers more extensive global coverage using data from the currently operational EnMAP satellite, which is continuously collecting data. In addition, SpectralEarth is approximately five times larger in pixel count and includes a temporal dimension, increasing its value for pre-training of hyperspectral foundation models.
    A comparison of existing labeled and unlabeled HSI datasets is provided in Table~\ref{tab:dataset_comparison}.

\subsection{Self-Supervised Learning for Remote Sensing}
    SSL emerged as a powerful paradigm for learning representations from unlabeled data~\cite{jing2020self}. Early SSL methods were based on engineered pretext tasks to encourage the model to learn distinct features by solving auxiliary problems such as image colorization~\cite{zhang2016colorful}, jigsaw puzzles~\cite{misra2020self}, or rotation angle prediction~\cite{gidaris2018unsupervised}. Recently, progress in SSL has been primarily focused on two families of methods:
    \begin{enumerate}
        \item \textbf{Joint-Embedding Architectures:} These methods train models to generate consistent representations for augmented views of the same input, enforcing invariance to these augmentations. To avoid convergence to trivial solutions, i.e., \textit{model collapse} to a constant representation, several strategies have been developed: Contrastive approaches like MoCo~\cite{he2020momentum} and SimCLR~\cite{chen2020simple} use negative pairs to encourage learning discriminative representations. Clustering-based methods, such as SwAV~\cite{caron2020unsupervised}, enforce balanced cluster assignment. Distillation-based frameworks like BYOL~\cite{grill2020bootstrap} and DINO~\cite{caron2021emerging} use an asymmetric teacher-student architecture. Methods like Barlow-Twins~\cite{zbontar2021barlow} and VICReg~\cite{bardes2022vicreg} explicitly use a loss regularization term to control the variance and feature correlation in the embedding space.
        \item \textbf{Masked Image Modeling (MIM):} Inspired by masked language modeling, MIM methods predict missing parts of an image from a masked input. Existing methods differ in their reconstruction targets, loss functions, or network architectures. For example, Masked Autoencoder (MAE)~\cite{he2022masked} and SimMIM~\cite{xie2022simmim} reconstruct raw pixels, BEiT~\cite{bao2021beit} predicts discrete visual tokens, while Data2Vec~\cite{baevski2022data2vec} and MSN~\cite{assran2022masked} employ a teacher-student architecture to regress targets generated from the teacher branch.
    \end{enumerate}

    In remote sensing, SSL techniques have been applied and adapted to multispectral and high-resolution RGB imagery. SeCo~\cite{manas2021seasonal} compiled a 200K-location dataset comprising Sentinel 2 images with multiple temporal views (one per season) and introduced seasonal contrast augmentation. The results supported the benefit of incorporating seasonal invariance in the training objective of contrastive methods. Building on this approach, SSL4EO-S12~\cite{wang2022ssl4eo} provided co-located Sentinel 2 and Sentinel 1 images and pre-trained models for both sensors. Subsequently, SSL4EO-L~\cite{stewart2024ssl4eo} extended this approach to Landsat imagery, further expanding the available pre-training data, downstream datasets, and pre-trained models. Several studies have adapted MAE for remote sensing imagery. 
    SatMAE~\cite{cong2022satmae} tokenized inputs across spectral and temporal dimensions, while Scale-MAE~\cite{reed2023scale} introduced ground-sampled distance positional encoding for multi-scale imagery. GFM~\cite{mendieta2023gfm} assembled GeoPile, a pre-training dataset comprising high-resolution RGB images, by merging various existing datasets. GeoPile was used to train an MAE model with an additional distillation loss guided by an ImageNet-22K pre-trained teacher. SpectralGPT~\cite{hong2024spectralgpt} and S2MAE~\cite{li2024s2mae} proposed a hierarchical MAE pre-training approach based on a spatial-spectral vision transformer for Sentinel 2 imagery. 
    Prithvi~\cite{jakubik2023foundation} and Prithvi-EO-2.0~\cite{szwarcman2024prithvi} built large-scale pretraining datasets from Harmonized Landsat-Sentinel-2 time-series and spatial/spectral/temporal ViT backbones using MAE. OmniSat~\cite{astruc2024omnisat} proposed a multi-sensor framework that combines masked autoencoding with contrastive learning, using high-resolution aerial imagery and multi-modal time series from Sentinel-1 and Sentinel-2.
    CROMA~\cite{fuller2024croma} combined contrastive learning and MIM for SAR-Optical self-supervised learning. msGFM~\cite{han2024bridgingremotesensorsmultisensor} proposed a multi-sensor MAE model with a cross-sensor reconstruction loss for paired images on RGB, Sentinel 1, Sentinel 2, and DSM data.  SenPa-MAE~\cite{prexl2024senpa} introduced a sensor-aware MAE pretraining strategy by incorporating metadata such as ground sampling distance and spectral response functions to allow sensor generalization. The model was trained with MAE on Sentinel-2, Planet-SuperDove, and Landsat-8/9.
    AnySat~\cite{astruc2024anysat} leveraged I-JEPA~\cite{assran2023self} to learn joint representations across several sensors. The model was trained on a combination of multiple datasets, including high-resolution imagery, Sentinel 1 and 2, MODIS, Landsat-6/7/8, and ALOS-2 time series.

    In the hyperspectral domain, most SSL works focus on pixel-level classification (with or without spatial context) where the pre-training scene is the same as the test image~\cite{li2023few,hou2021hyperspectral,liu2023self,braham2024enhancing,cao2023transformer,ibanez2022masked,tian2024swin}. While this approach is effective for classical HSI datasets, the learned representations cannot generalize across scenes. To address this issue, HSIMAE~\cite{wang2024hsimae} proposed HSIHybrid, a compilation of existing HSI datasets, and pre-trained an MAE-like model. This approach effectively increases data diversity but poses challenges when combining sensors with different characteristics, requiring pre-processing steps such as PCA to unify the number of spectral bands. Additionally, the scale of HSIHybrid remains limited when training HSI foundation models. Recent studies have explored collecting unlabeled HSI data for pre-training. For example, MSST~\cite{scheibenreif2023masked} used MAE on EnMAP imagery, introducing decoupled spectral and spatial attention blocks to reduce the computational cost associated with spatial-spectral tokenization. DOFA~\cite{xiong2024neural} proposed a general-purpose architecture that dynamically generates patch embedding weights from sensor wavelength metadata. The model was trained using MAE combined with a distillation loss on RGB, multispectral, SAR, and EnMAP hyperspectral imagery. HyperSIGMA~\cite{wang2024hypersigmahyperspectralintelligencecomprehension} proposed an MAE-inspired approach with independent spectral and spatial transformers and evaluated their model on several hyperspectral downstream tasks. However, the data used for pre-training these models is less diverse than SpectralEarth. Additionally, these studies focus exclusively on ViTs and MAE, excluding ConvNets and other pre-training methodologies. In contrast, our work explores a more diverse set of architectures and pre-training strategies and provides a library of pretrained models for HSI applications.

%% file: sec/3_datasets.tex
\section{SpectralEarth \& Benchmarks}
This section introduces the SpectralEarth dataset and related downstream datasets for benchmarking.  

\begin{figure*}[h!]
    \centering
    \begin{subfigure}[t]{0.191\linewidth}
         \centering
         \includegraphics[width=\linewidth]{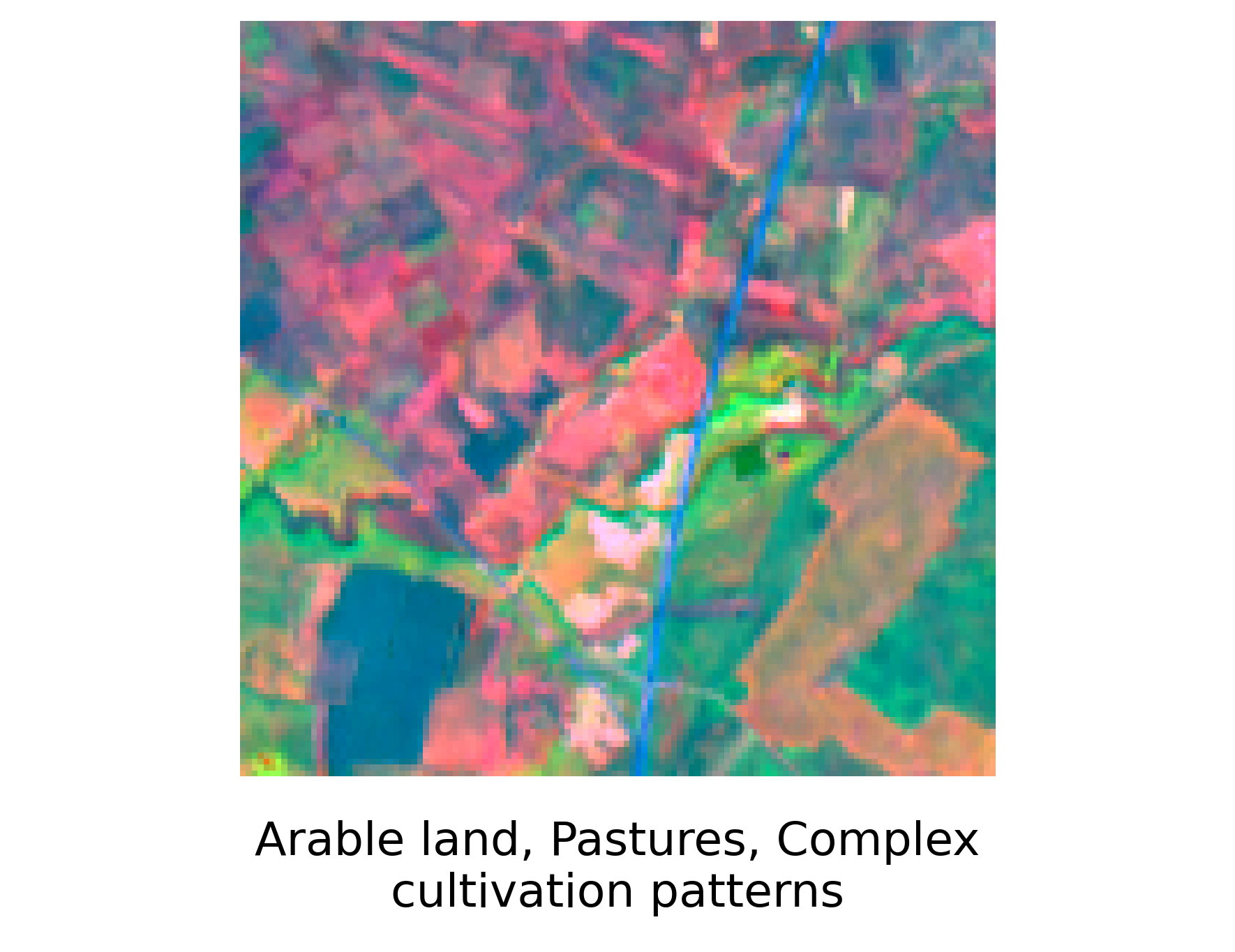}
         \captionsetup{width=\linewidth}
         \caption{\smaller\textbf{CORINE}}
    \end{subfigure}
    \hfill
    \begin{subfigure}[t]{0.275\linewidth}
         \centering
         \includegraphics[width=\linewidth]{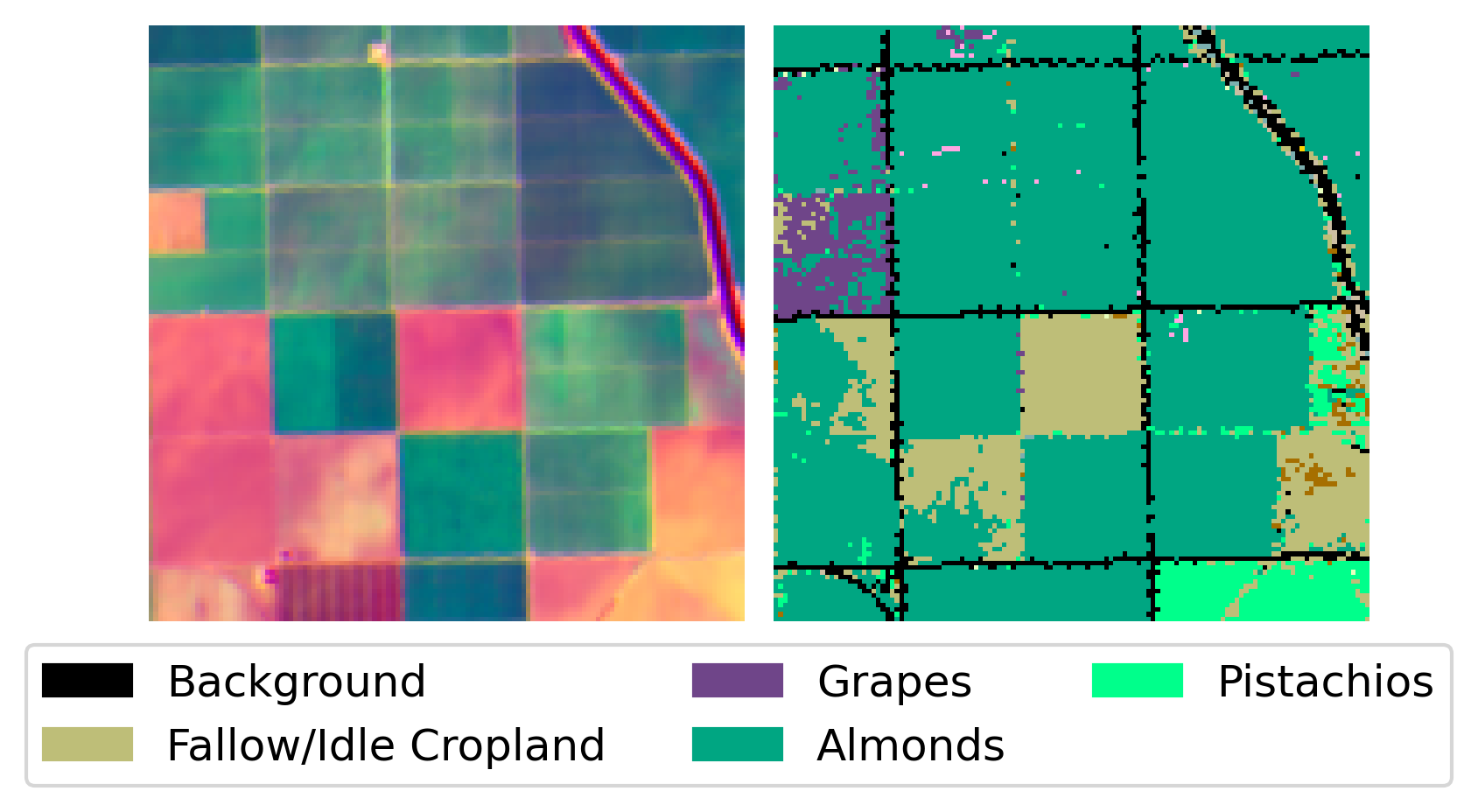}
         \captionsetup{width=\linewidth}
         \caption{\smaller\textbf{CDL}}
    \end{subfigure}
    \hfill
    \begin{subfigure}[t]{0.245\linewidth}
         \centering
         \includegraphics[width=\linewidth]{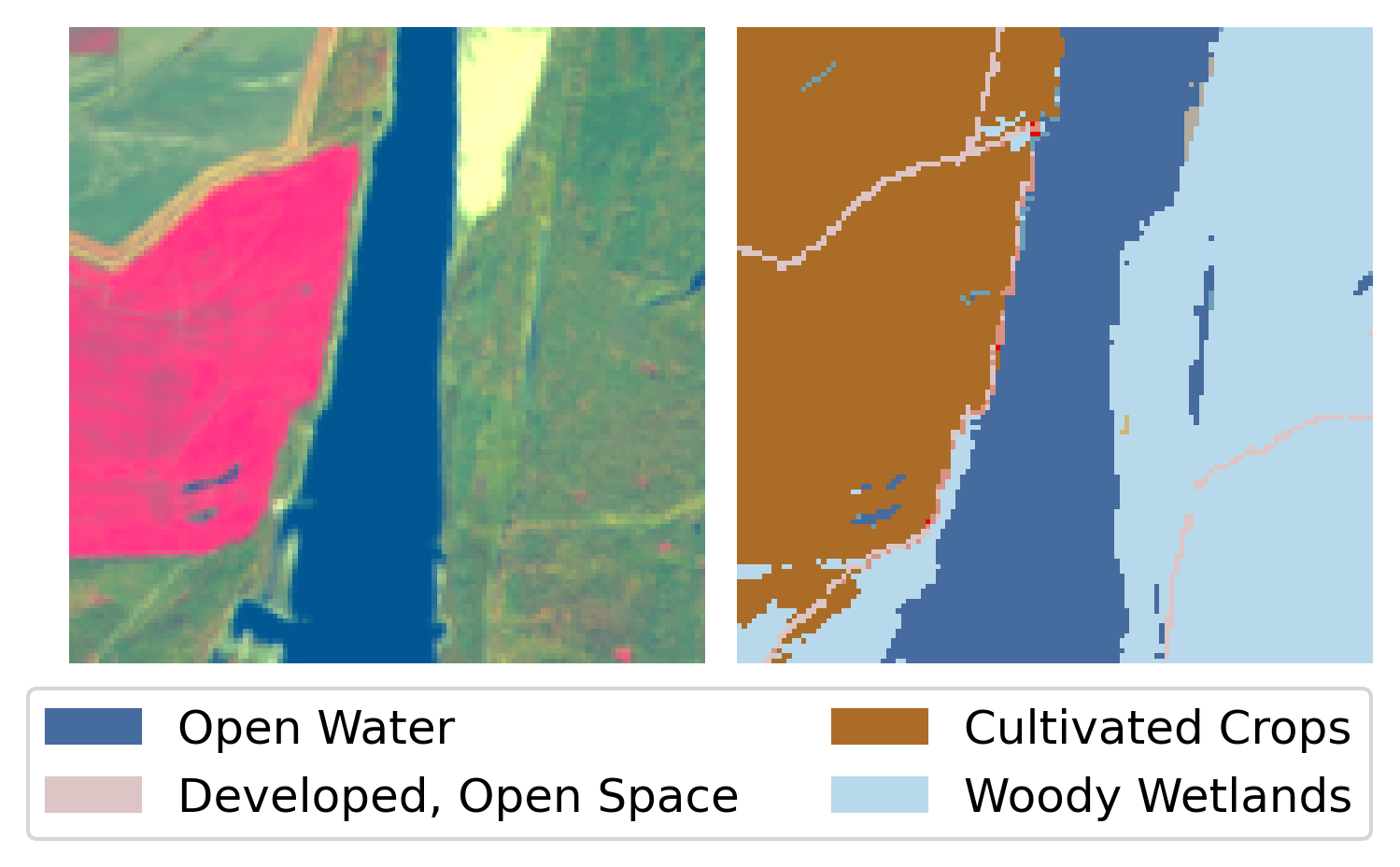}
         \captionsetup{width=\linewidth}
         \caption{\smaller\textbf{NLCD}}
    \end{subfigure}
    \hfill
    \begin{subfigure}[t]{0.270\linewidth}
         \centering
         \includegraphics[width=\linewidth]{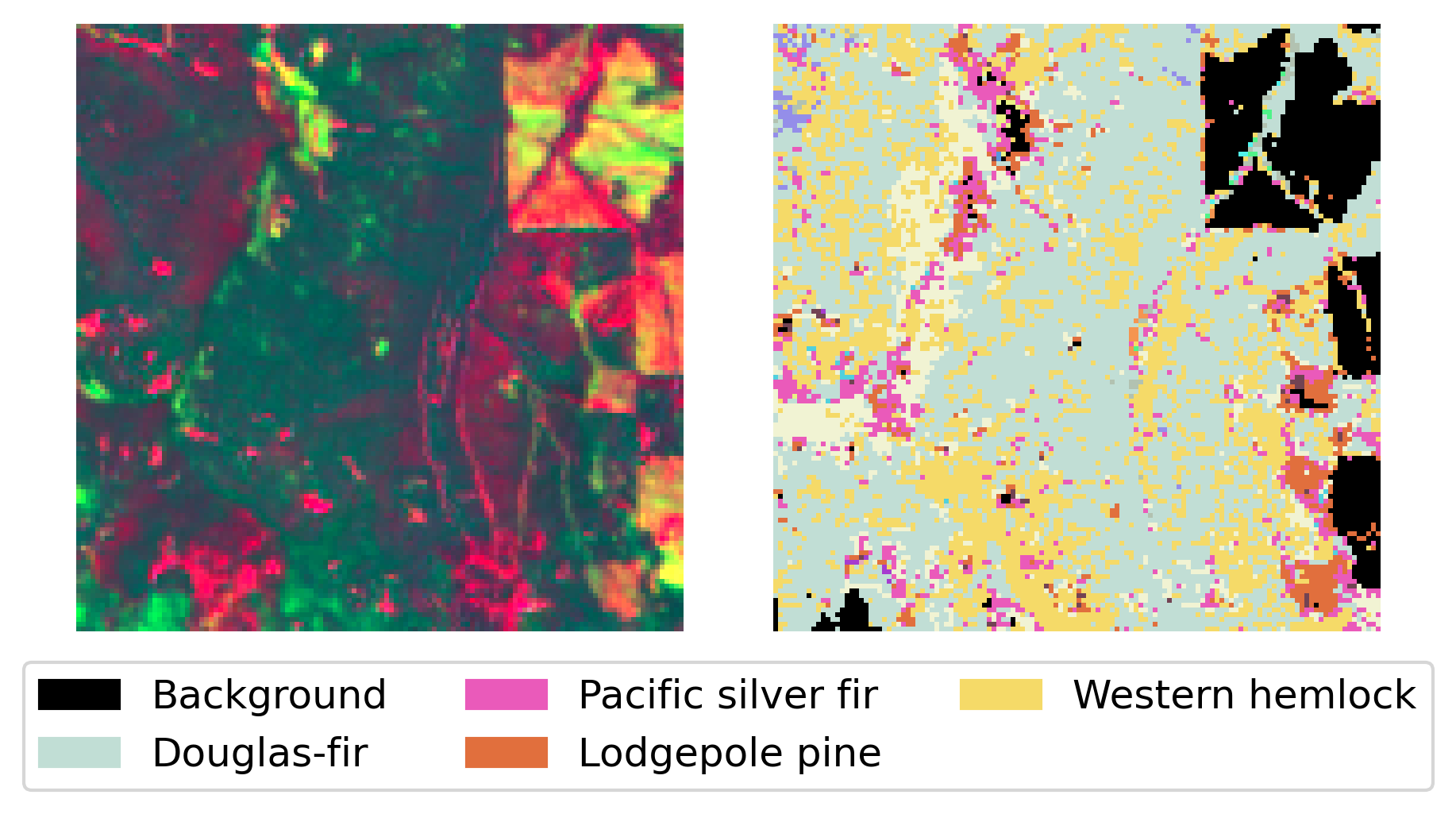}
         \captionsetup{width=\linewidth}
         \caption{\smaller\textbf{TreeMap}}
    \end{subfigure}
    
    \vspace{2mm}
    
    \begin{subfigure}[t]{0.29\linewidth}
         \centering
         \includegraphics[width=\linewidth]{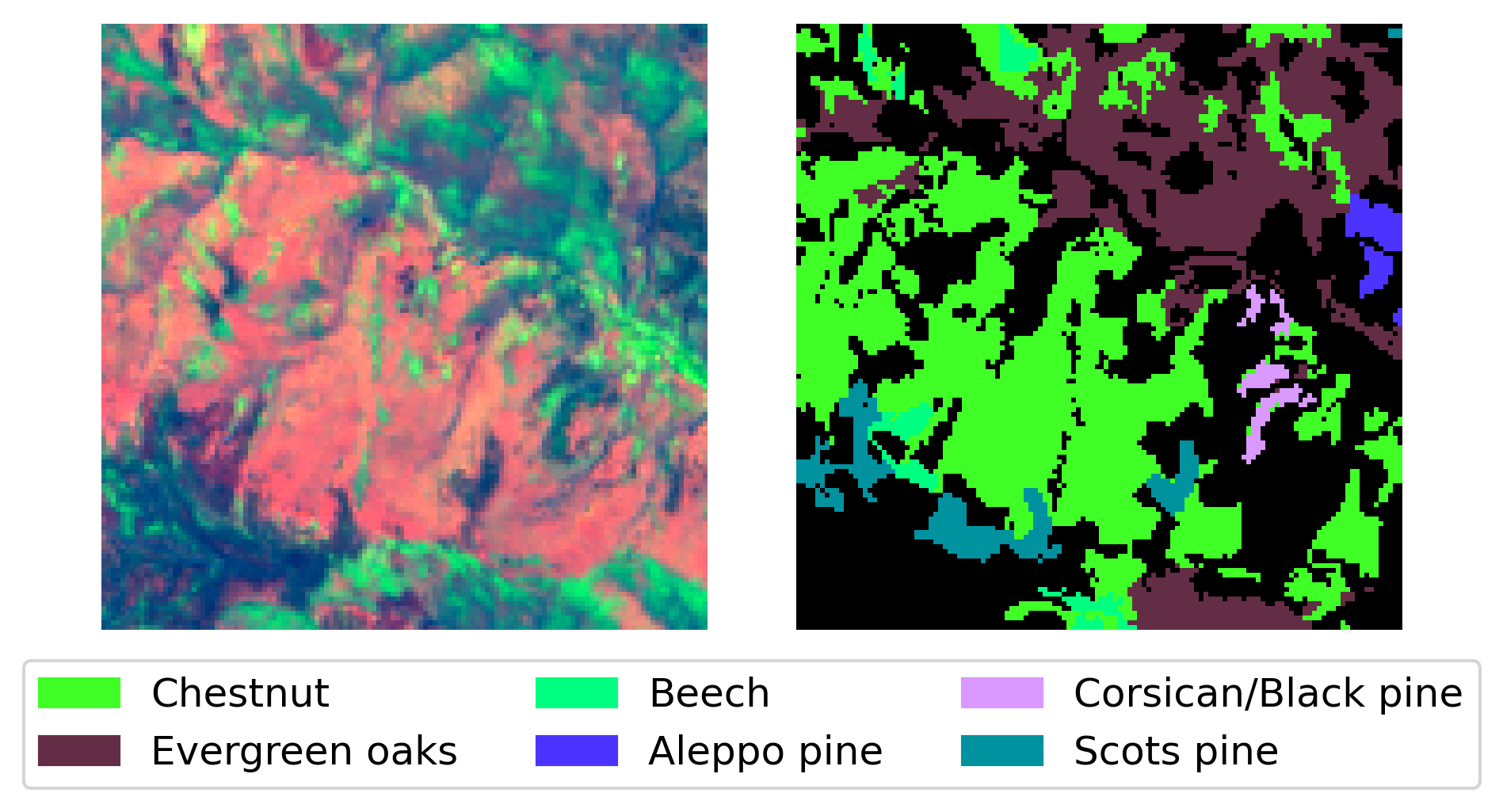}
         \captionsetup{width=\linewidth}
         \caption{\smaller\textbf{BD-Foret}}
    \end{subfigure}
    \hfill
    \begin{subfigure}[t]{0.33\linewidth}
         \centering
         \includegraphics[width=\linewidth]{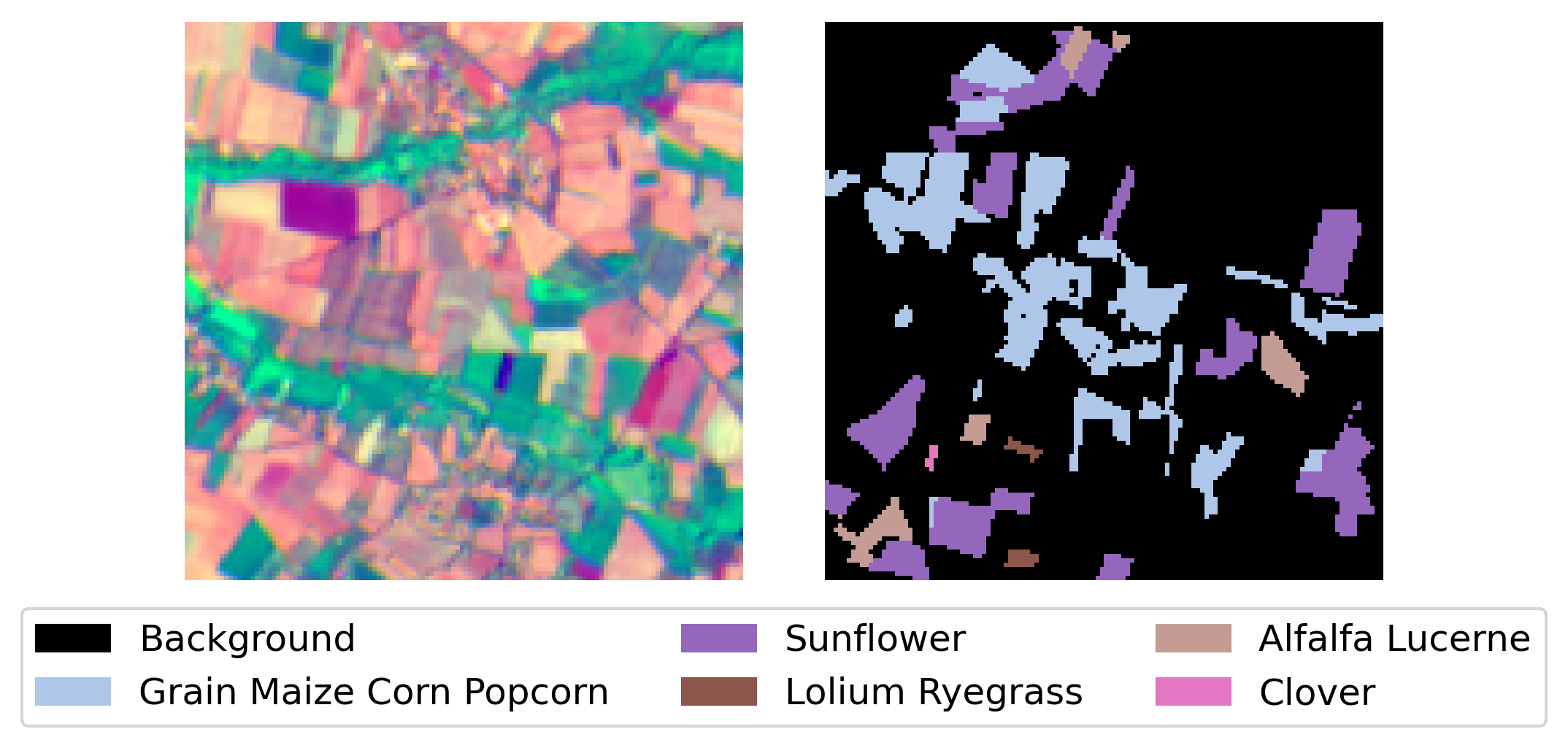}
         \captionsetup{width=\linewidth}
         \caption{\smaller\textbf{EuroCrops}}
    \end{subfigure}
    \hfill
    \begin{subfigure}[t]{0.33\linewidth}
         \centering
         \includegraphics[width=\linewidth]{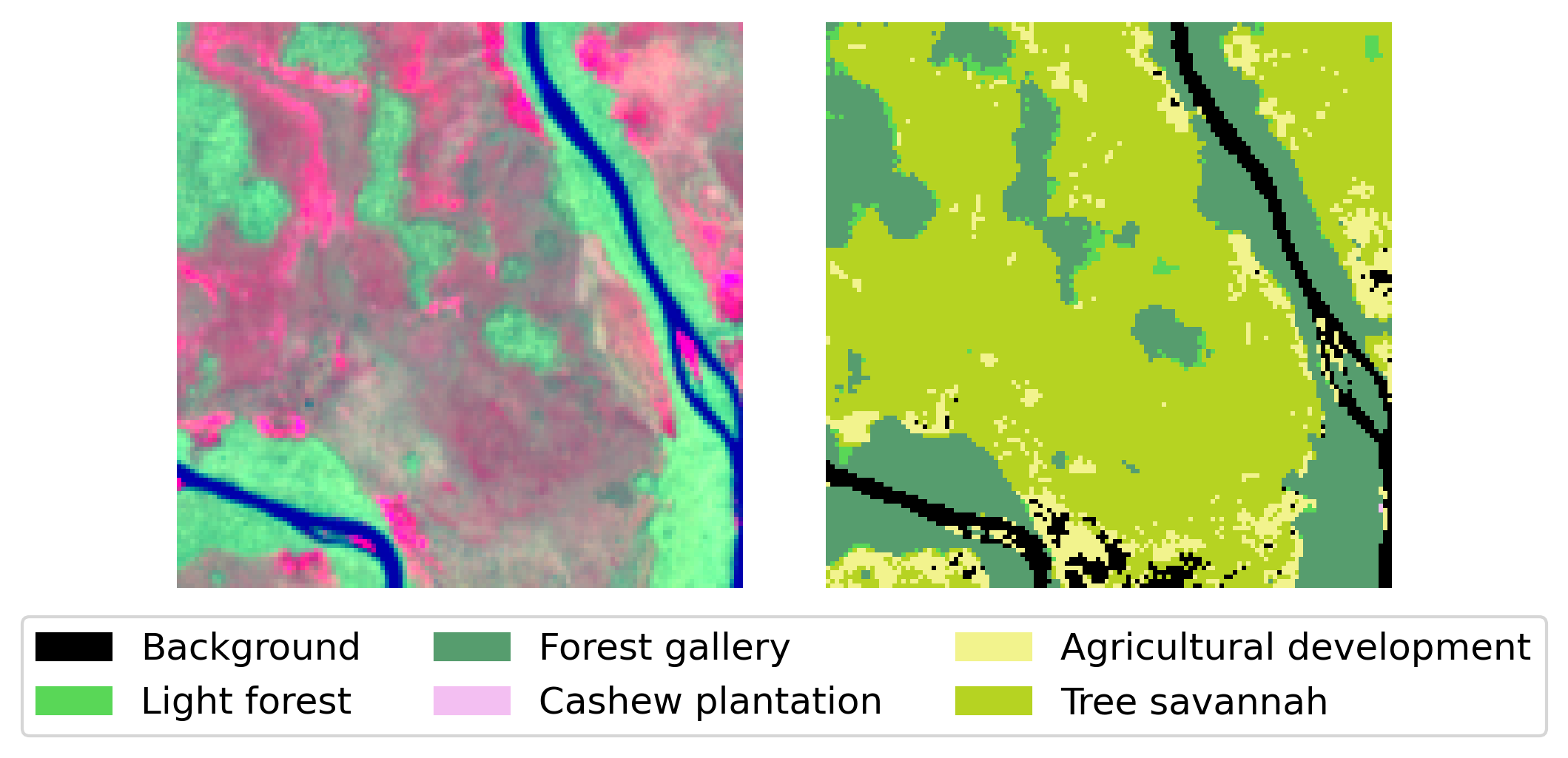}
         \captionsetup{width=\linewidth}
         \caption{\smaller\textbf{BNETD}}
    \end{subfigure}
    
    \caption{Sample pseudo-RGB images from the proposed EnMAP benchmarks.}
    \label{fig:enmap_benchmarks}
\end{figure*}

\subsection{SpectralEarth}
    \label{sec:PretrainingDataset}
    \paragraph{Data Acquisition}
        The EnMAP data was sourced from the \href{https://eoweb.dlr.de/egp}{DLR GeoPortal}. A total of 11,636 tiles were retrieved via the portal's graphical interface through a significant manual effort. 
        All tiles were carefully selected based on human visual inspection. Cloud coverage was kept below $\sim$10\%. The selection process encompassed the entire EnMAP archive from April 27, 2022 to April 24, 2024. All acquired tiles underwent radiometric, geometric, and atmospheric corrections using the L2A processor from the EnMAP mission~\cite{de2023enmap}. 
    \paragraph{Data Pre-processing}
            The EnMAP tiles have been split into geospatial patches---each of size 128$\times$128 pixels where an individual pixel includes 224 bands. Bands dominated by water absorption, specifically [127–141] and [161–167], were excluded due to the frequent presence of missing values (\texttt{no-data}). Removing those bands is a common practice in hyperspectral dataset creation~\cite{fuchs2023hyspecnet,hyperspectral_scenes_ehu}. As a result, the total number of bands per image was reduced to 202. 
            
    \paragraph{Temporal Views Extraction}
        To exploit all available tiles, we utilized overlaps between EnMAP data acquisitions to generate time-series of EnMAP patches\footnote{EnMAP records hyperspectral imagery based on a schedule generated from user requests \cite{kaufmann2012science}, i.e.\ the identification of spatially aligned patches for multiple timestamps is a valuable asset our data curation provides.}. At the same time, SpectralEarth patches for fixed geo-locations never overlap. This process required (a) the identification of tiles that overlap spatially and (b) extracting patches from the intersection of those tiles. Time series include valuable information on landscape evolution and provide characteristic signals for seasonal variation---an asset for contrastive learning \cite{manas2021seasonal,wang2022ssl4eo}.
        Algorithm~\ref{alg:patchify} outlines the steps taken to generate temporal positives by high-level pseudo-code. An \emph{overlap graph} is first constructed to identify spatial intersections between tiles. For each tile, the algorithm forms candidate subsets of overlapping tiles, computes their joint intersection, and extracts patches of fixed size from these regions. We use an R-tree to efficiently track and store patch locations and avoid overlaps.

        \begin{algorithm}
        \caption{Temporal Views Extraction}
        \label{alg:patchify}
        \begin{algorithmic}[1]
        \Procedure{Main Procedure}{}
            \State \textit{tiles} $\gets$ EnMAPData
            \State \textit{overlap\_graph} $\gets$ \textsc{GetOverlaps}(\textit{tiles})
            \State \textit{R\_tree} $\gets$ $K_0$\Comment{\small empty tree, for SpectralEarth patches}
            \For{\textit{tile} \textbf{in} \textit{tiles}}
                \State \textit{combs} $\gets$ \textsc{BuildCombinations}(\textit{tile}, \textit{overlap\_graph})
                \For{\textit{tile\_subset} \textbf{in} \textit{combs}}
                    \State \textit{intersection} $\gets$ \textsc{Intersection}(\textit{tile\_subset})
                    \State \textit{patches} $\gets$ \textsc{Patchify}(\textit{intersection})
                    \State \textsc{INSERT}(\textit{R\_tree}, \textit{patches})
                \EndFor
            \EndFor
        \EndProcedure
        \newline
        \Function{BuildCombinations}{\textit{tile}, \textit{overlap\_graph}}
            \State \textit{combinations} $\gets$ \textsc{GetEdges}(\textit{tile}, \textit{overlap\_graph})\footnotemark
            \For{subset size \textit{n} \textbf{in} $[3, 4, \dots]$}
                \State get \textit{n}-tuples from (\textit{n}-1)-tuples \textbf{in} \textit{combinations}
                \State compute intersections of all \textit{n}-tuples
                \State keep largest \textit{n}-tuples by area
                \If{no valid \textit{n}-tuple found}
                    \State \textbf{break}
                \EndIf
                \State add \textit{n}-tuples to \textit{combinations}
            \EndFor
            \State \Return \textit{combinations}
        \EndFunction
        \end{algorithmic}
        
        \end{algorithm}

        As a result of Algorithm \ref{alg:patchify}, we identified a few long time series in the EnMAP archive. For those involving more than 25 intersecting tiles, calculating intersections of all subsets was computationally intractable due to combinatorial explosion. To address this, we implemented several heuristics to optimize the computation:
        \begin{enumerate}
            \item Consider only a subset of possible intersections using a breadth-first search (BFS) approach.
            \item Avoid redundant computations across tiles.
            \item Parallelization of the code across all connected components of the overlap graph.
        \end{enumerate}

         \begin{figure*}[t]
        \centering
        \begin{subfigure}[t]{0.31\textwidth}
            \centering
            \includegraphics[width=\linewidth]{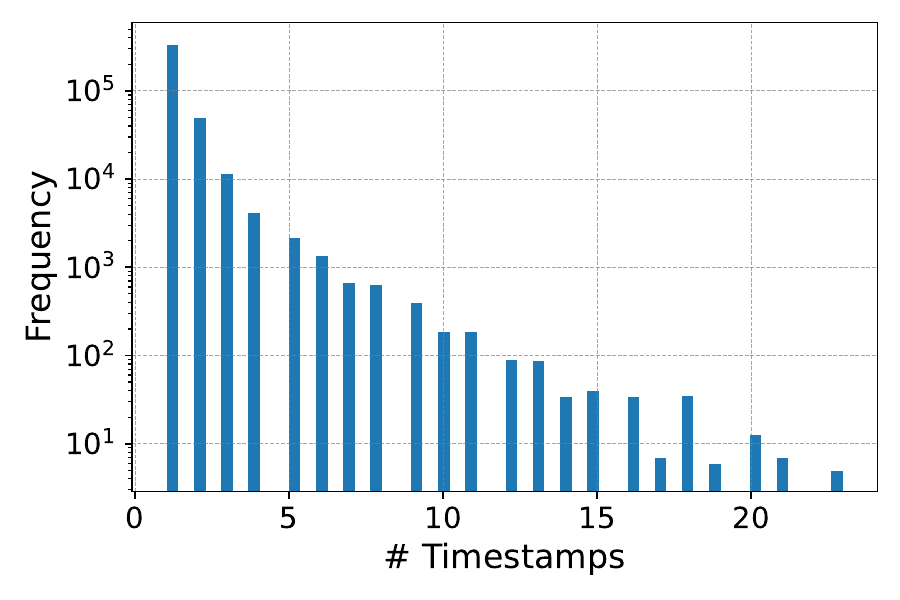}
            \caption{Number of timestamps per geo-location (log scale).}
            \label{fig:n_timestamps}
        \end{subfigure}
        \hfill
        \begin{subfigure}[t]{0.31\textwidth}
            \centering
            \includegraphics[width=\linewidth]{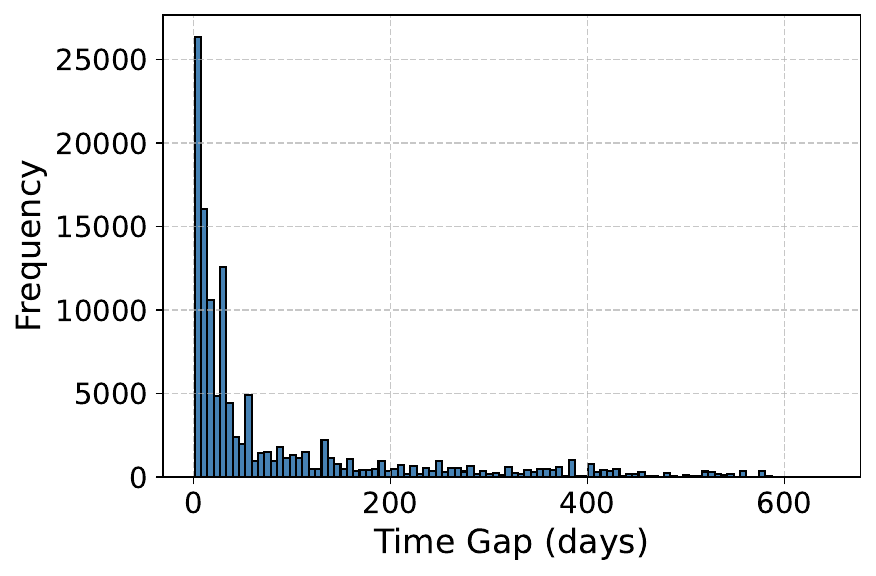}
            \caption{Gap between consecutive timestamps.}
            \label{fig:timestamp_gap}
        \end{subfigure}
        \hfill
        \begin{subfigure}[t]{0.31\textwidth}
            \centering
            \includegraphics[width=\linewidth]{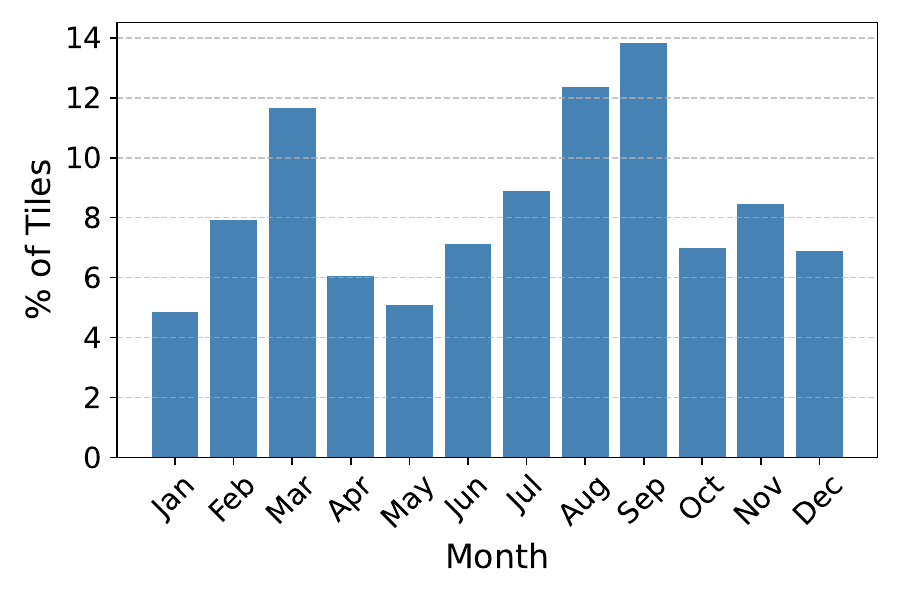}
            \caption{Monthly distribution aggregated over 2 years.}
            \label{fig:spectral_earth_temp}
        \end{subfigure}
        \caption{Temporal distribution of SpectralEarth: (a) number of timestamps per location, (b) time gaps between acquisitions, and (c) monthly acquisition frequency.}
        \label{fig:temporal_distribution}
    \end{figure*}

        Following this approach, we extracted 73,307 non-overlapping locations with multiple timestamps. For areas with a single timestamp, we patchified avoiding any spatial overlap. Correspondingly, in SpectralEarth, we extracted 415,153 locations with a spatial extent of 128$\times$128 pixels, totaling 538,974 patches. The spatial coverage of SpectralEarth showcases the dataset’s global diversity in land-cover types (Figure~\ref{fig:spectral_earth_main_fig}) and geographical distribution (Figure~\ref{fig:spectral_earth_map}). Notably, the data distribution varies across geo-locations, reflecting the flight request-based archive composition of EnMAP.
        
        We analyze the temporal coverage of our dataset in Figure~\ref{fig:temporal_distribution}. Figure~\ref{fig:n_timestamps} illustrates the distribution of timestamp counts in SpectralEarth. Over the two-year period, 17.5\% of locations were covered by more than one timestamp in the EnMAP archive---with the majority having only two timestamps. This highlights the challenge of collecting extensive hyperspectral time series data with broad geographical coverage today. Missions such as ESA's CHIME satellite constellation \cite{rast2021copernicus} will eventually resolve the issue. Figure~\ref{fig:timestamp_gap} shows the histogram of time gaps between consecutive acquisitions, restricted to samples with at least two timestamps. We observe that the distribution is skewed towards smaller values. Notably, the highest peak is 4 days, matching EnMAP's revisit period. The median gap between two consecutive acquisitions across all samples is 27 days.
        
        We also analyze the seasonal distribution of EnMAP acquisitions, as shown in Figure~\ref{fig:spectral_earth_temp}. SpectralEarth includes data from all months, with a relatively balanced distribution. However, certain disparities can be observed, e.g., January is underrepresented while July, August, and March are over-represented. This can be attributed to several factors:
        \begin{itemize}
            \item Increased cloud cover during winter months, which inherently limits the availability of clear imagery;
            \item A satellite outage in \href{https://www.enmap.org/news/2022-12-15}{December-January 2023} and \href{https://www.enmap.org/news/2022-12-15}{December-January 2024}, which resulted in a temporary interruption of data acquisition;
            \item The task scheduling nature of the EnMAP mission. As EnMAP is operationally tasked based on specific use cases, the data distribution inherently reflects real-world demands of satellite imagery, leading to a natural variation in seasonal coverage.
        \end{itemize}

    \begin{figure*}[ht!]
    \centering
    \begin{subfigure}[b]{0.3\linewidth}
        \centering
        \includegraphics[width=\linewidth]{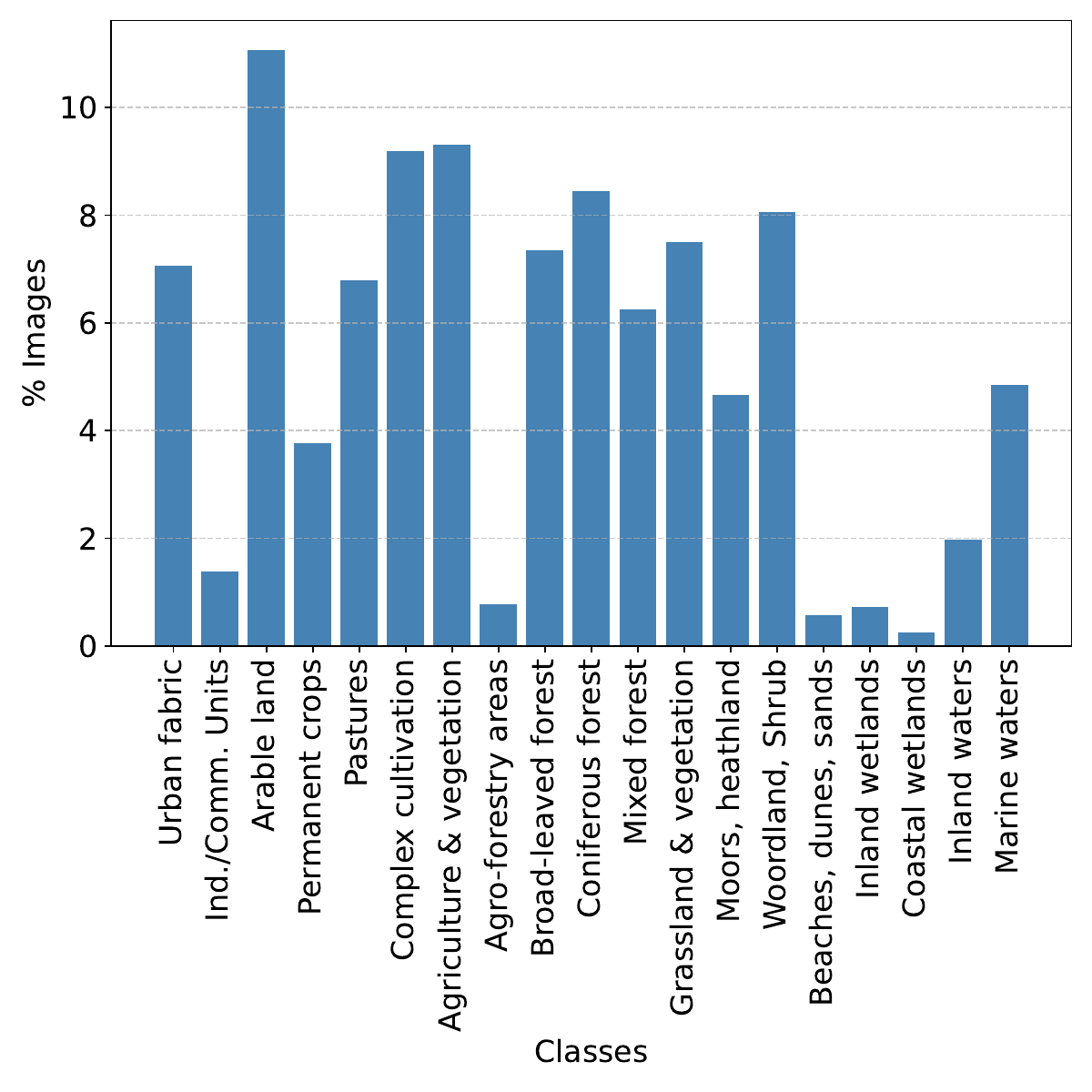}
        \caption{\smaller\textbf{EnMAP-CORINE}}
        \label{fig:enmap_corine_class_dist}
    \end{subfigure}
    \hfill
    \begin{subfigure}[b]{0.3\linewidth}
        \centering
        \includegraphics[width=\linewidth]{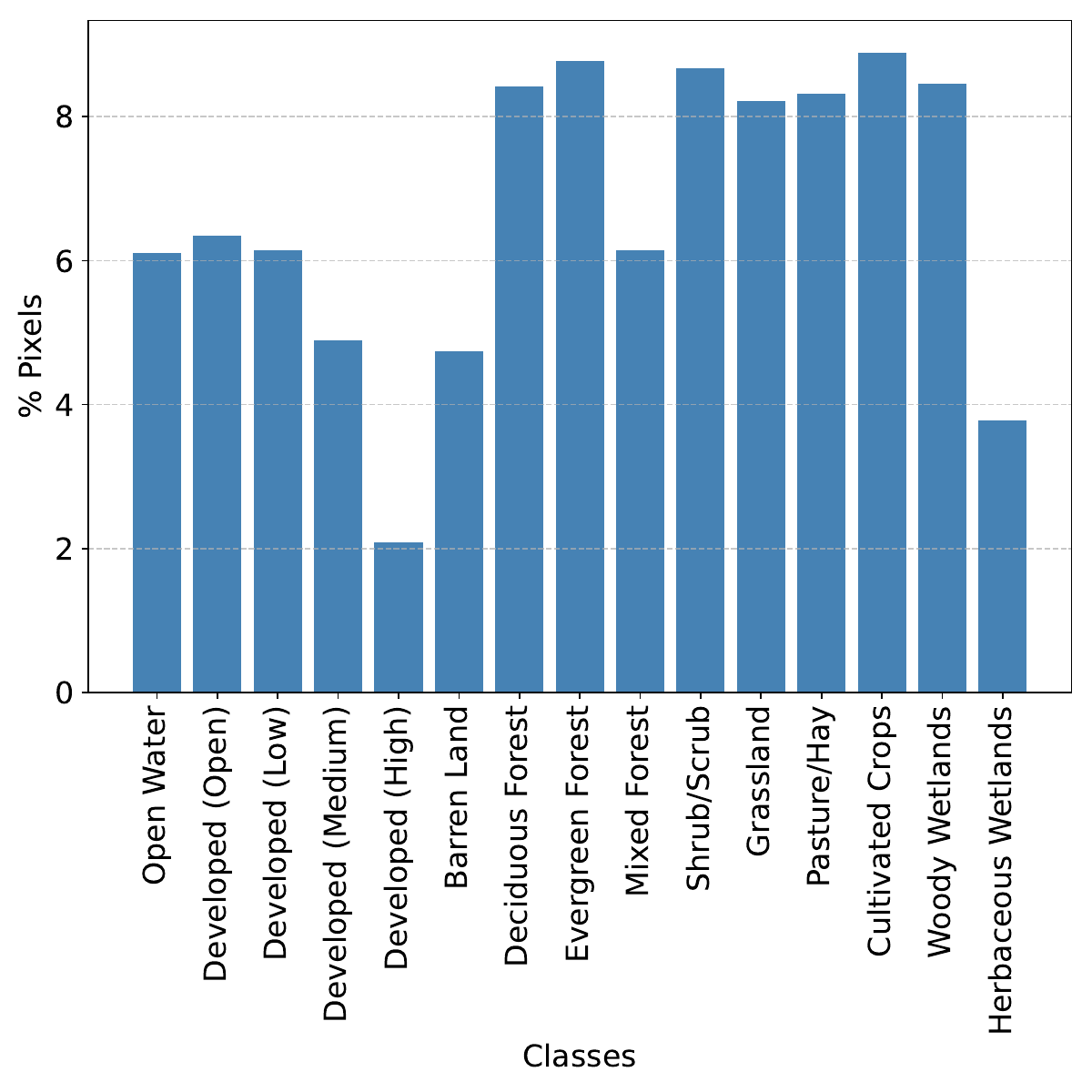}
        \caption{\smaller\textbf{EnMAP-NLCD}}
        \label{fig:enmap_nlcd_class_dist}
    \end{subfigure}
    \hfill
    \begin{subfigure}[b]{0.3\linewidth}
        \centering
        \includegraphics[width=\linewidth]{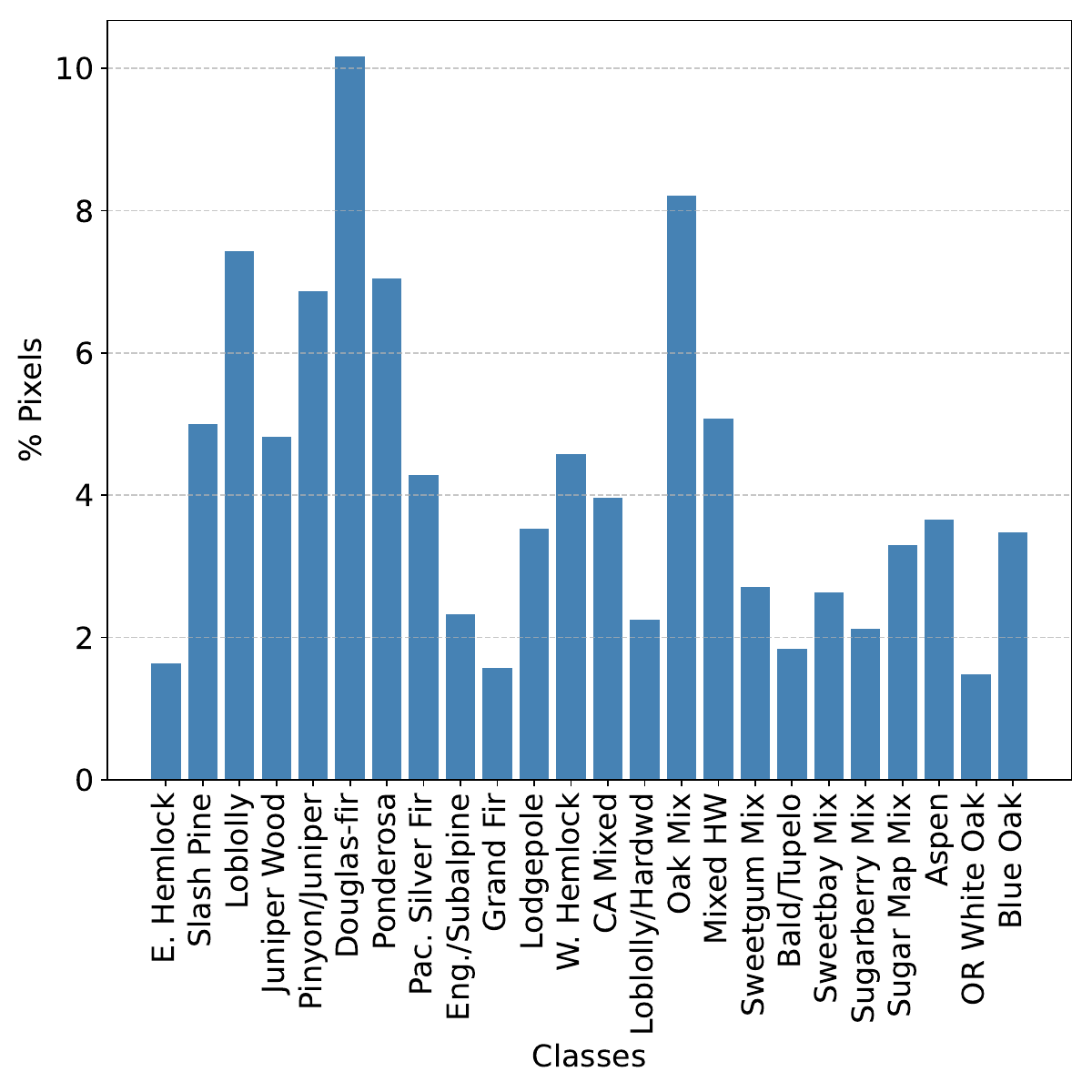}
        \caption{\smaller\textbf{EnMAP-Treemap}}
        \label{fig:treemap_class_dist}
    \end{subfigure}
    
    \vspace{2mm}
    
    \begin{subfigure}[b]{0.3\linewidth}
        \centering
        \includegraphics[width=\linewidth]{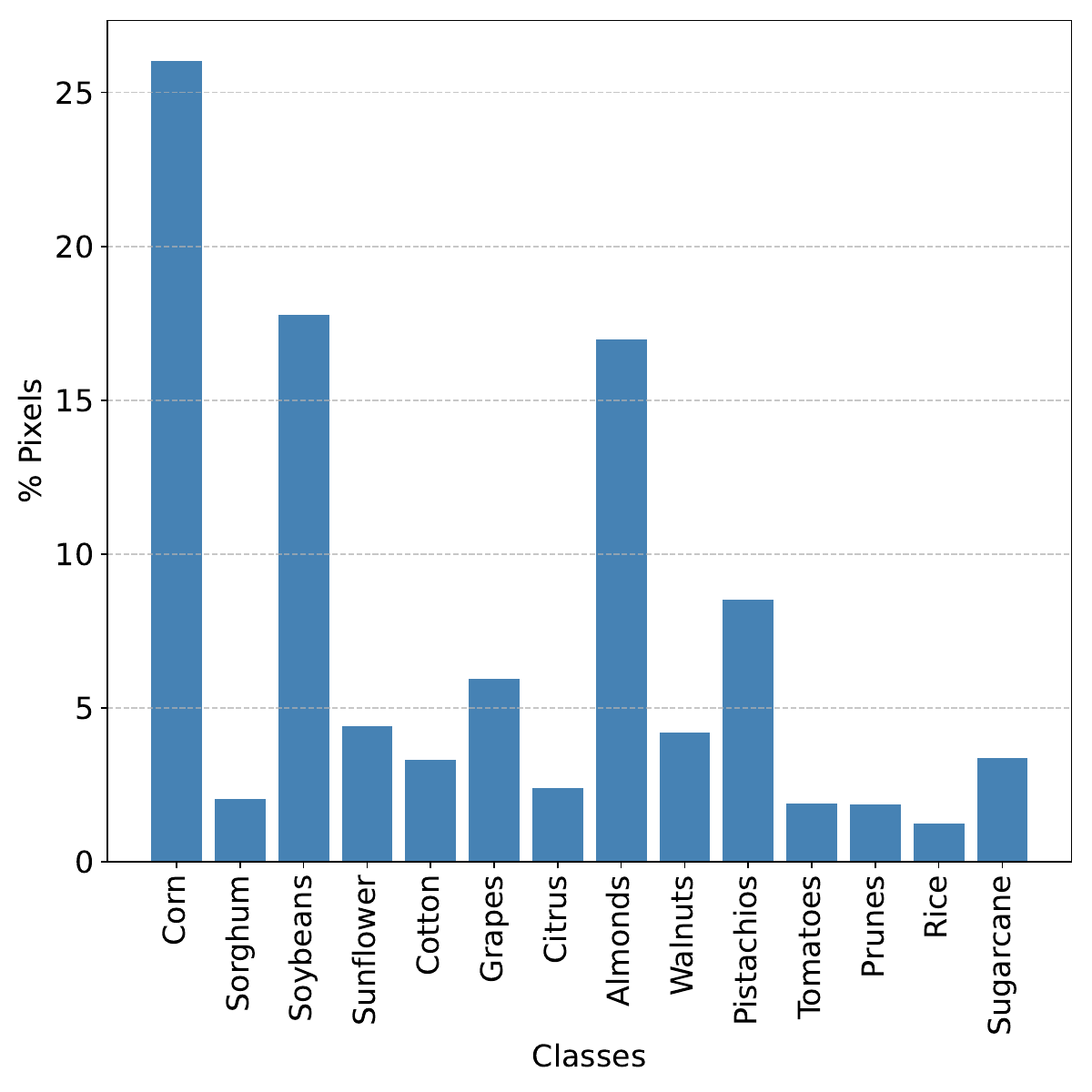}
        \captionsetup{width=\linewidth}
        \caption{\smaller\textbf{EnMAP-CDL}}
        \label{fig:enmap_cdl_class_dist}
    \end{subfigure}
    \hfill
    \begin{subfigure}[b]{0.3\linewidth}
        \centering
        \includegraphics[width=\linewidth]{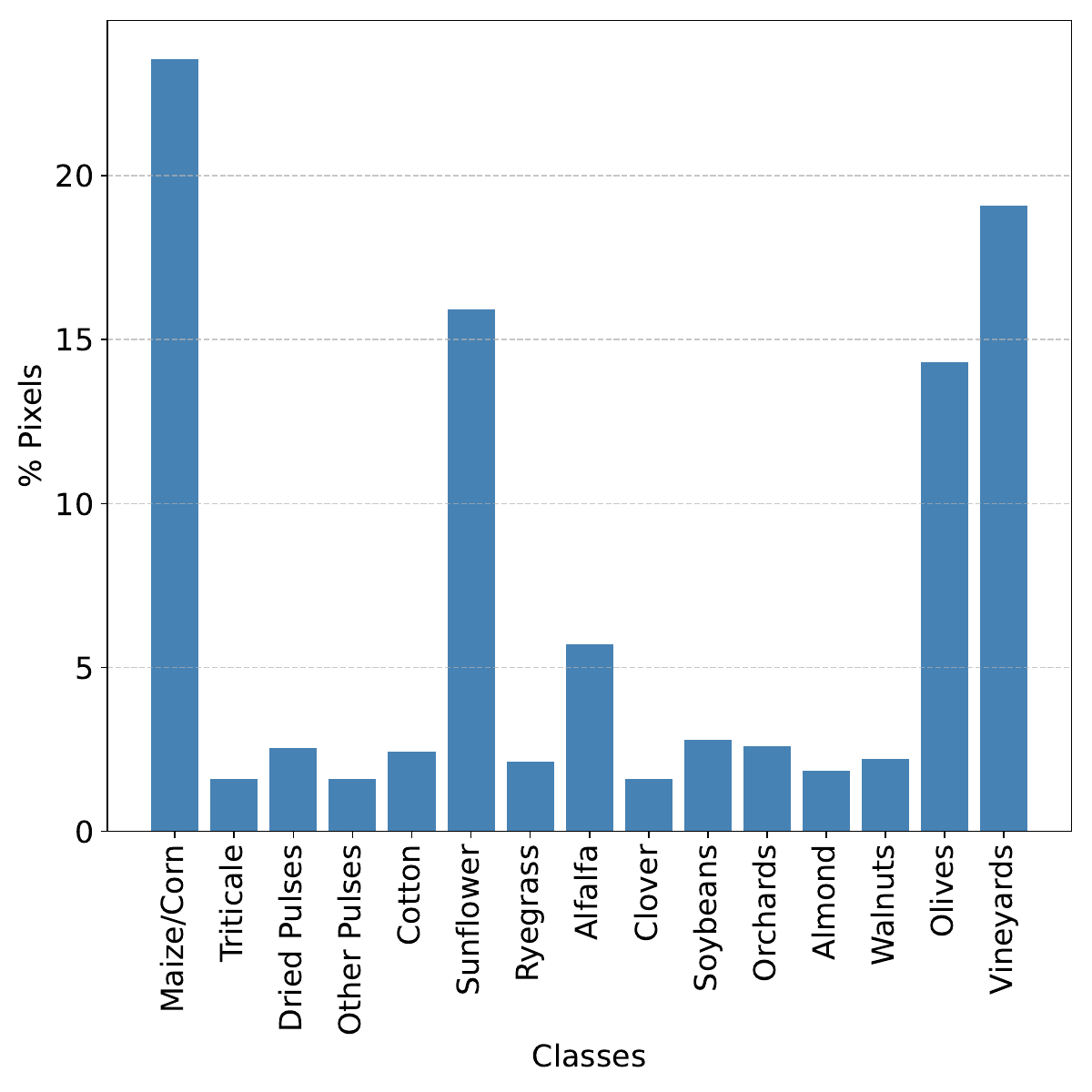}
        \captionsetup{width=\linewidth}
        \caption{\smaller\textbf{EnMAP-EuroCrops}}
        \label{fig:eurocrops_class_dist}
    \end{subfigure}
    \hfill
    \begin{subfigure}[b]{0.3\linewidth}
        \centering
        \includegraphics[width=\linewidth]{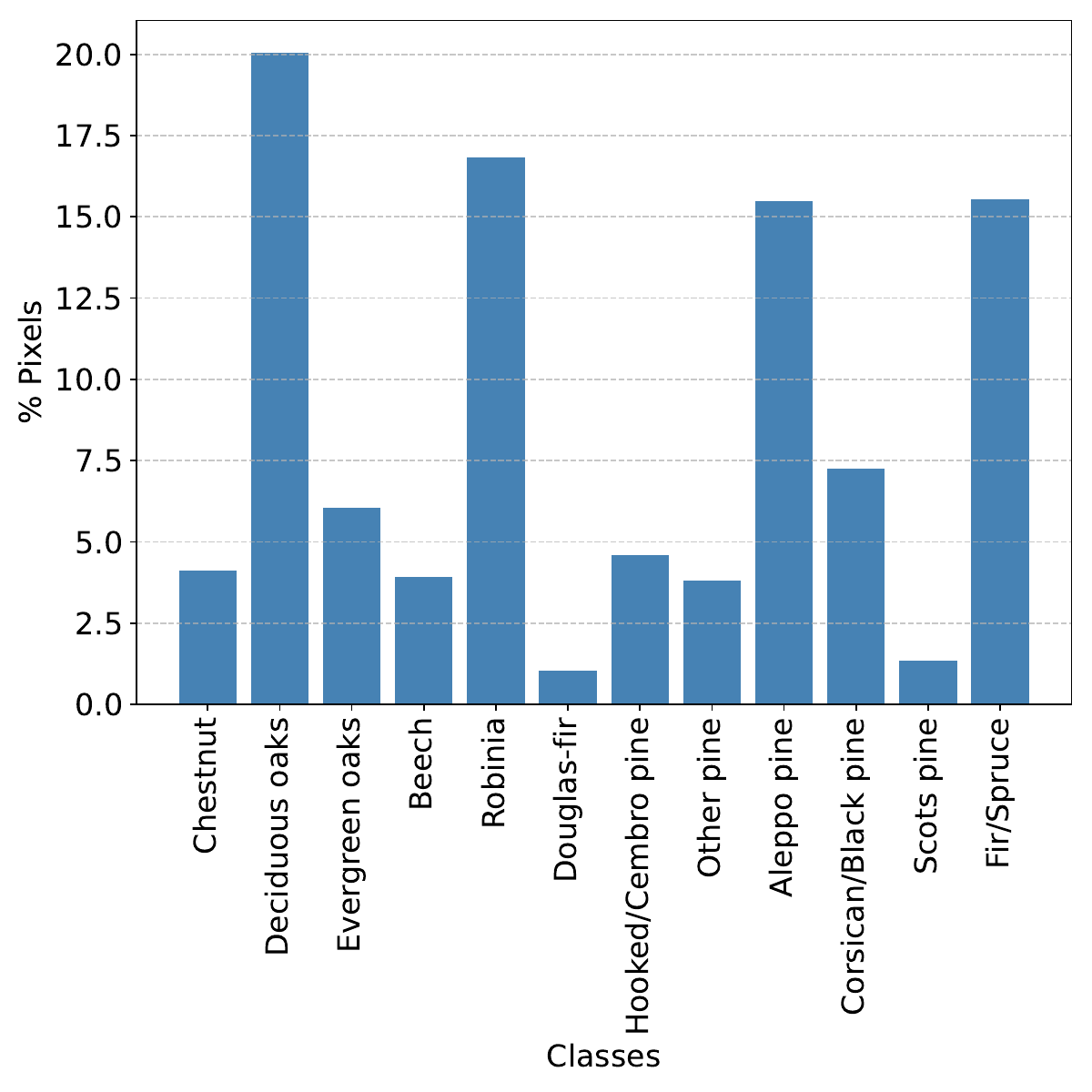}
        \captionsetup{width=\linewidth}
        \caption{\smaller\textbf{EnMAP-BDForet}}
        \label{fig:bdforet_class_dist}
    \end{subfigure}
    
    \vspace{2mm}
    
    \begin{subfigure}[b]{0.3\linewidth}
        \centering
        \includegraphics[width=\linewidth]{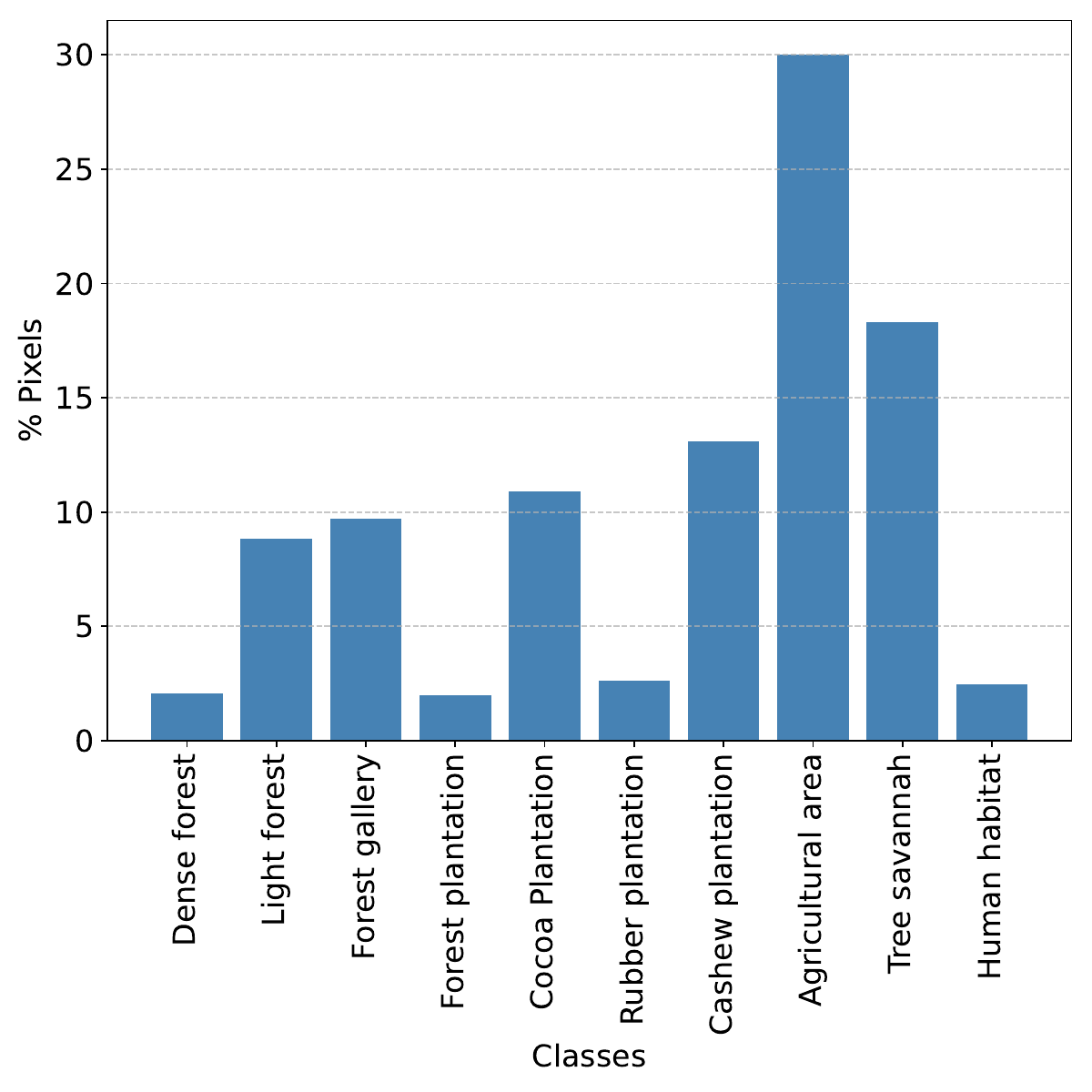}
        \captionsetup{width=\linewidth}
        \caption{\smaller\textbf{EnMAP-BNETD}}
        \label{fig:bnetd_class_dist}
    \end{subfigure}
    \hfill
    \begin{subfigure}[b]{0.3\linewidth}
        \centering
        \includegraphics[width=\linewidth]{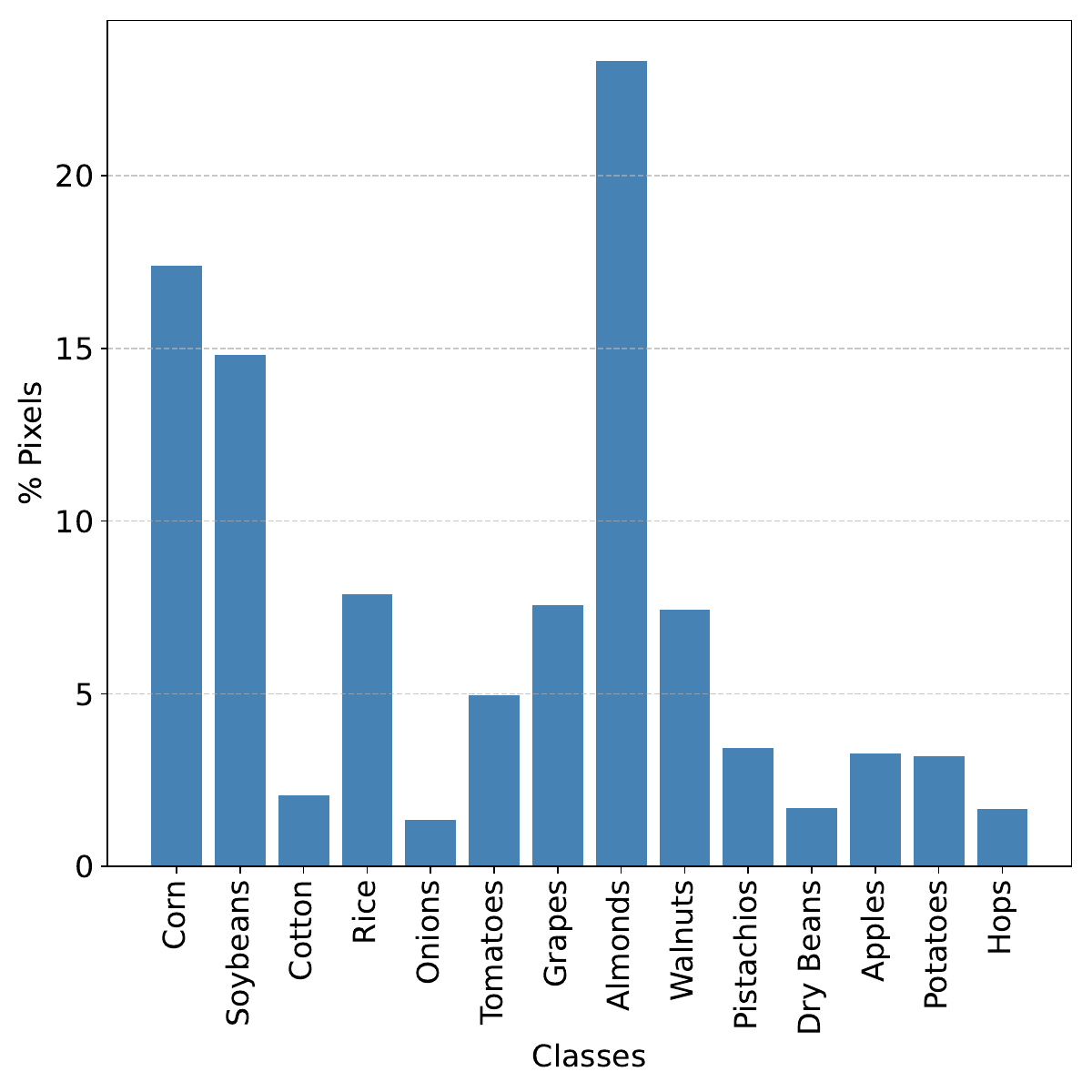}
        \captionsetup{width=\linewidth}
        \caption{\smaller\textbf{DESIS-CDL}}
        \label{fig:desis_cdl_class_dist}
    \end{subfigure}
    \hfill
    \begin{subfigure}[b]{0.3\linewidth}
        \centering
        \includegraphics[width=\linewidth]{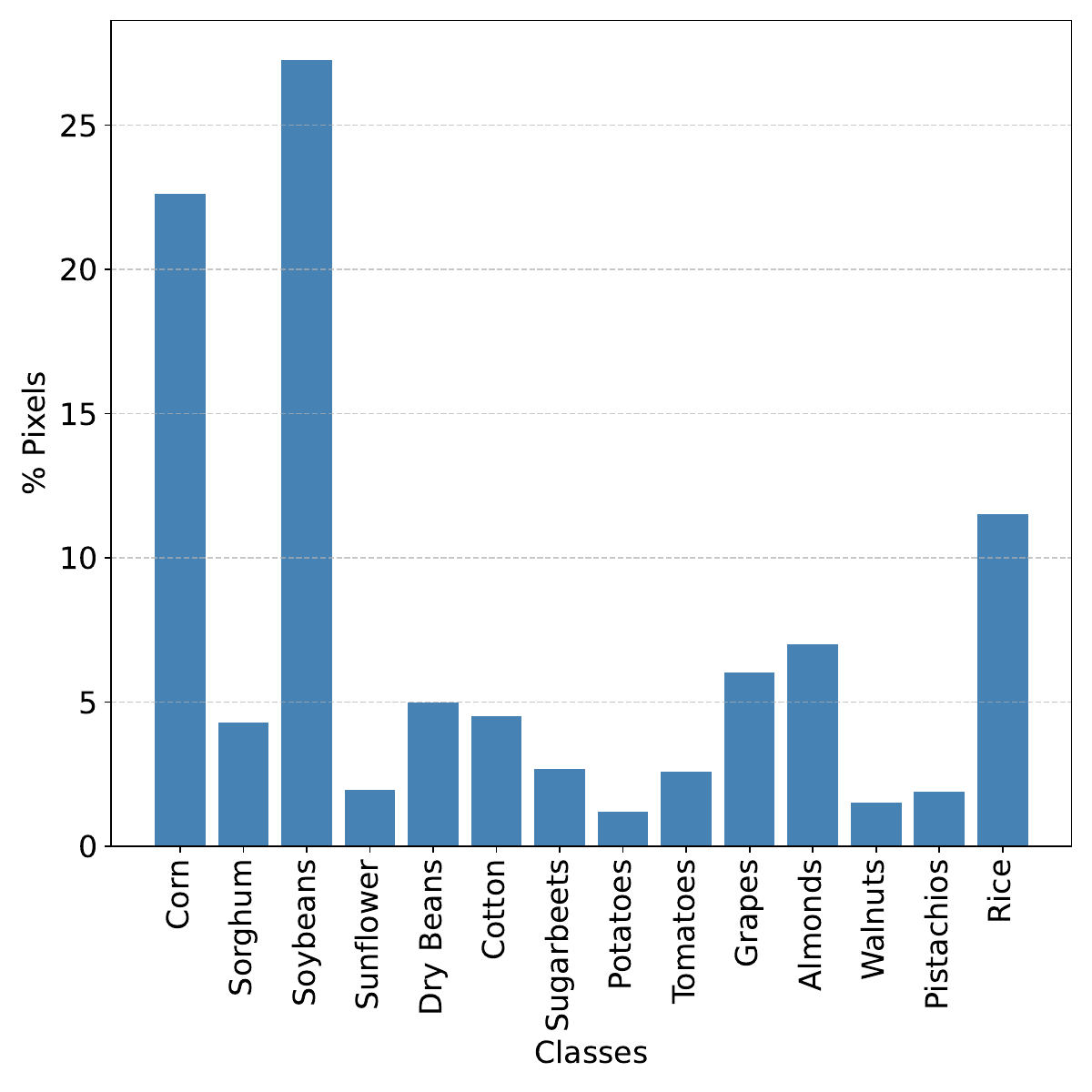}
        \captionsetup{width=\linewidth}
        \caption{\smaller\textbf{EO1-CDL}}
        \label{fig:eo1_cdl_class_dist}
    \end{subfigure}
    
    \caption{Class distributions for the downstream datasets introduced in SpectralEarth.}
    \label{fig:combined_class_distributions}
\end{figure*}

    \subsection{Downstream Tasks}
    \label{sec:downstream_tasks}
        To evaluate the models pretrained on SpectralEarth, we assembled nine downstream datasets for benchmarking. Each benchmark involves a subset of EnMAP/DESIS/EO-1 images aligned with specific geospatial products focusing on different aspects of land cover, agricultural, and forest analysis. While the labels in these datasets carry some inherent uncertainty, they are sufficiently accurate for model evaluation. Similar strategies have previously been employed to create remote sensing datasets and geospatial foundation model benchmarking~\cite{sumbul2019bigearthnet,sumbul2019bigearthnet,jakubik2023foundationmodelsgeneralistgeospatial}. An overview of the SpectralEarth downstream tasks is provided in Figure~\ref{fig:enmap_benchmarks}.
        \paragraph{EnMAP-CORINE - Land Cover Classification}
            This dataset pairs EnMAP imagery with the \href{https://land.copernicus.eu/en/products/corine-land-cover/clc2018}{\textit{CORINE} land cover database} for Europe. The CORINE land cover map has a 100m spatial resolution, which is coarse-grained relative to EnMAP's 30m resolution. The CORINE product reports an overall thematic accuracy exceeding 85\%. Our SpectralEarth image patches correspond to approximately 38x38 CORINE land cover pixels. Consequently, we create a multi-label classification benchmark where each SpectralEarth patch is annotated with the set of CORINE classes covered. The resulting subset, called EnMAP-CORINE, includes 11,000 patches, each multi-labeled with 19 distinct classes. These classes have been aggregated from the 44 categories defined in CORINE following the taxonomy proposed by BigEarthNet-MM~\cite{sumbul2021bigearthnet}. The EnMAP-CORINE dataset exhibits a diverse range of land cover types. As Figure~\ref{fig:enmap_corine_class_dist} demonstrates, some classes such as 'Arable Land' and 'Complex cultivation patterns' are more frequent, while others like 'Coastal/Inland wetlands' and 'Beaches, dunes, sand' are comparatively rare. This class imbalance reflects the distribution of natural land cover in Europe. 
        \paragraph{EnMAP-CDL - Crop Type Segmentation}
            This dataset aligns EnMAP images with the Cropland Data Layer (CDL)~\cite{boryan2011monitoring} product for the United States. 
            The CDL map is published annually by the \href{https://www.usda.gov/}{USDA}'s National Agricultural Statistics Service since 2008. It targets agricultural crop classification. In addition to cultivated crops such as Corn and Soybeans, CDL encompasses non-cultivated areas like Grassland and Shrubland. According to the USDA metadata, classification accuracy is generally between 85\% and 95\% for major crop-specific classes. To minimize the impact of seasonal variations on crop classification, the dataset utilizes SpectralEarth patches from the summers of 2022 and 2023. We mask out all non-agricultural classes, as well as winter crops (e.g., Winter Wheat), and crops that are likely to be harvested before the end of summer (e.g., Oats). Given the strong imbalance among CDL classes, we resampled the class distribution for a more balanced representation. The resulting dataset comprises 1,600 patches with a total of 14 classes. The class distribution of the EnMAP-CDL dataset is shown in Figure~\ref{fig:enmap_cdl_class_dist}. Overall, the dataset exhibits a reasonable level of class balance among the different crop types, albeit some over-represented classes such as Corn. 
        \paragraph{EnMAP-NLCD - Land Cover Segmentation}
            This downstream dataset leverages the National Land Cover Database (NLCD)~\cite{dewitz2019nlcd} from the U.S.\ Geological Survey (USGS) to provide pixel-level land cover annotations. The NLCD map, a comprehensive land cover product, has been published every 2–3 years since 2001, with data available back to 2019. The NLCD 2019 product reports an overall accuracy of approximately 91\%. We pair EnMAP patches over the US to the NLCD map, and manually filter out cloudy and snowy images.
            The resulting dataset contains 13,500 patches, each annotated with pixel-wise land cover class labels from one out of the 15 land cover classes. As depicted in Figure~\ref{fig:enmap_nlcd_class_dist}, the class distribution is well-balanced.
        \paragraph{EnMAP-TreeMap - Tree Species Segmentation} The EnMAP-TreeMap dataset maps EnMAP imagery from 2022 and 2023 to TreeMap~\cite{riley2022treemap} product, a 30\,m resolution map of forest attributes across the continental United States. We use the forest-type layer, which encodes the dominant species group based on live tree stocking. To mitigate the temporal mismatch between TreeMap (2016) and EnMAP acquisitions, we mask regions affected by deforestation using the Hansen Global Forest Change dataset~\cite{hansen2013high}. The resulting dataset includes 10,000 patches covering 24 tree species, enabling fine-grained classification of forest types. The class distribution is depicted in Figure~\ref{fig:treemap_class_dist}.
        \paragraph{EnMAP-BDForet - Tree Species Segmentation} The EnMAP-BDForet dataset combines EnMAP hyperspectral imagery acquired between 2022 and 2024 with the BDForet V2 dataset from the Institut Géographique National (IGN) France~\footnote{\url{https://inventaire-forestier.ign.fr/spip.php?article646}}. We project the dominant tree species attribute onto the EnMAP grid to generate the labels. To mitigate land cover changes, we filter out areas affected by deforestation using the latest IGN forest mask from 2023. The resulting dataset comprises 2,550 patches with 12 tree species. The class distribution is shown in Figure~\ref{fig:bdforet_class_dist}. 

        \paragraph{EnMAP-EuroCrops - Crop Type Segmentation} The EnMAP-EuroCrops dataset aligns EnMAP imagery with crop type labels derived from the EuroCrops dataset~\cite{schneider2023eurocrops}, which is sourced from national agricultural inventories across European countries. This dataset focuses on crop-type classification for four countries—France, Germany (Brandenburg), Czechia, and Spain—for the year 2023, with additional coverage for France in 2022. Similar to CDL, we exclude crops which are likely to be harvested before our EnMAP acquisition. The resulting dataset consists of 1,800 patches with 15 crop-type classes. Approximately 68\% of the patches are located in France, 27\% in Spain, and the remaining 5\% in Germany and Czechia. The class distribution is illustrated in Figure~\ref{fig:eurocrops_class_dist}.
        \paragraph{EnMAP-BNETD - Land Cover Segmentation} To explore the geographical generalizability of our models beyond pre-training regions, we introduce the EnMAP-BNETD dataset, covering land cover classification in Ivory Coast. This dataset uses the BNETD 2020 Land Cover Map~\footnote{\url{https://africageoportal.maps.arcgis.com/apps/webappviewer/index.html?id=88c2493e722546c09c2a0a8b394c4454}}, produced by the BNETD-CIGN with support from the European Union, and validated through extensive field campaigns in 2022 and 2023. The reported overall accuracy is 91\%. We pair EnMAP hyperspectral imagery with land cover labels from this product, resulting in a dataset of 2,100 patches with 10 classes.  The class distribution is shown in Figure~\ref{fig:bnetd_class_dist}.
        \paragraph{DESIS-CDL - Crop Type Segmentation} The DESIS-CDL dataset contains images captured for the summers of 2018-2023 from the DESIS (DLR Earth Sensing Imaging Spectrometer Mission) instrument, matched with the corresponding CDL mask for crop segmentation. DESIS images are generated at 30m spatial resolution with 235 bands covering wavelengths from 400 nanometers through a micron. The resulting dataset comprises 1,000 images with 14 classes. Due to the limited availability of DESIS tiles, it was not possible to pair this dataset with EnMAP-CDL. Therefore, the class distribution (see Figure~\ref{fig:desis_cdl_class_dist}) and geographical coverage differ. 
        \paragraph{EO1-CDL - Crop Type Segmentation} The EO1-CDL dataset includes EO-1 Hyperion hyperspectral data paired with crop-type labels from the CDL. EO-1 Hyperion provides 220 spectral bands covering 0.357 to 2.576 micrometers, out of which we use 198. The dataset includes Hyperion scenes from 2002 to 2016, focusing on agricultural regions over the US. Due to the limited availability of Hyperion imagery in Google Earth Engine, the dataset consists of 550 patches distributed across 14 crop classes.

%% file: sec/4_models.tex
\section{Models}
\label{sec:models}
    \begin{figure}[t!]
      \centering
      \includegraphics[width=\linewidth]{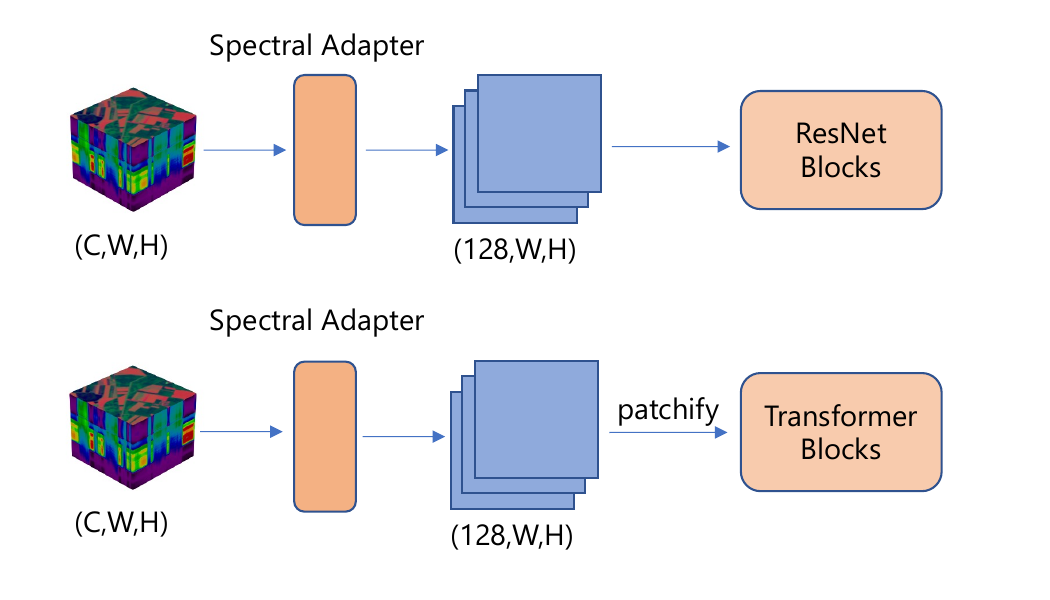}
      \caption{Schematic representation of the proposed backbones tailored for HSI. We augment classical ResNet and ViT with a spectral adapter to process spectral information effectively.}
      \label{fig:backbones}
    \end{figure}
        This section presents the design choices behind our models, including the choice of SSL algorithms and network architecture.
        \subsection{Pre-training Algorithms}
            \label{sec:ssl_algorithms}
            Hyperspectral data has very different characteristics compared to natural images. Therefore, traditional SSL leaderboards in computer vision may not directly translate to the hyperspectral domain. Therefore, we explore a broad range of SSL methods to establish a set of baseline pretrained models for hyperspectral imagery. We use MoCo-V2~\cite{he2020momentum} and DINO~\cite{caron2021emerging} for their effectiveness in frozen encoder performance~\cite{oquab2023dinov2} and their compatibility with both CNNs and ViTs.
            Furthermore, we include MAE~\cite{he2022masked} for its strong performance in fine-tuning scenarios. Together, these methods form a representative set of algorithms to investigate SSL in the hyperspectral domain.

        \subsection{Network Architectures}
        \label{sec:network_arch}
            Since we aim for large-scale coverage and pre-training with spaceborne imagery, our models must operate on large patches instead of pixels or small patches, as traditionally done in the hyperspectral literature. Most models developed at the pixel level in HSI literature do not scale well with large patches~\cite{zhong2017spectral,hong2020graph,hong2021spectralformer}. On the other hand, models like ResNet~\cite{he2016deep} and ViT~\cite{dosovitskiy2020image} are designed to handle larger spatial contexts. Therefore, we seek to adapt these architectures for hyperspectral imaging. A key challenge is that classical ResNet and ViTs treat spectral bands independently, failing to capture the correlations between adjacent bands and the overall characteristics of the spectrum. Our models should address the following requirements for hyperspectral imaging:

            \begin{itemize}
                \item \textbf{Spectral Feature Extraction:} We aim to extract features from both the spatial and the spectral domain. 
                \item \textbf{Adaptability to Diverse Sensors:} The architecture should accommodate variability in the number of spectral bands. This property is important when transferring pretrained models to different sensors.
                \item \textbf{Preserving Fine-Grained Details:} Natural images often deal with large objects compared to medium-resolution remote sensing data. Therefore, the spatial downscaling in classical vision models can lead to a significant loss of details. Our architecture needs to preserve this fine-grained information.
                \item \textbf{Computational Efficiency:} 3D convolution or spectral-spatial tokenization becomes computationally expensive for large patches, with many spectral bands, and deep architectures. In particular, self-attention scales quadratically with the number of tokens~\cite{vaswani2017attention}, and 3D convolutions scale linearly with the number of spectral bands~\cite{tran2015learning}. We therefore seek lightweight adaptation of classical architectures with minimal overhead.  
            \end{itemize}
                
            To meet these requirements, we follow a pragmatic approach and implement a simple modular design, integrating 1D convolutional layers as a Spectral Adapter at the onset of the standard vision backbones, as depicted in Figure~\ref{fig:backbones}. These layers perform convolutions across the spectral dimension, generating feature maps for the core backbone. The Spectral Adapter consists of three (1D Conv + BN + ReLu) layers. The kernel sizes for the 1D convolutions are 7, 7, and 5, with strides of 5, 5, and 3, respectively. The resulting feature maps have 128 channels. To accommodate different sensors and feed the input into standard 2D backbones, we use a global pooling layer to aggregate the remaining spectral dimensions (if any). Specifically:
                
            \begin{itemize}
                \item \textit{Spectral ResNet:} We replace the stem layers with the spectral adapter. By removing the original stem layers in the ResNet architecture, we avoid the 4$\times$downscaling factor of the image to better capture fine-grained details. Additionally, the first bottleneck block is adjusted to take an input feature map of 128 channels instead of the ResNet's 64-channel input at this stage. 
                \item \textit{Spectral ViT:} The spectral adapter is placed before the patch embedding layer, ensuring the transformer receives fixed-size spectral features as input. We use 4$\times$4 patches instead of the typical 16$\times$16 patches, which helps maintain fine spatial details and better preserves spectral information at the patch projection layer. Like the spectral ResNet model, we set the number of input channels in the projection layer to 128.
            \end{itemize}
                
            Although these simplifications may not capture all spectral and spatial interactions as explicitly as 3D convolutions or spectral-spatial tokenization, they offer a practical trade-off that scales better to large datasets and diverse sensors.
            Similar designs have been used effectively in HSI classification tasks~\cite{ahmad2024traditional, audebert2019deep}. Our approach adapts this idea to large-scale SSL, providing a simple scalable solution for hyperspectral representation learning across sensors.

%% file: sec/5_experimental_setup.tex
\begin{figure}[t!]
    \centering
    \includegraphics[width=\linewidth]{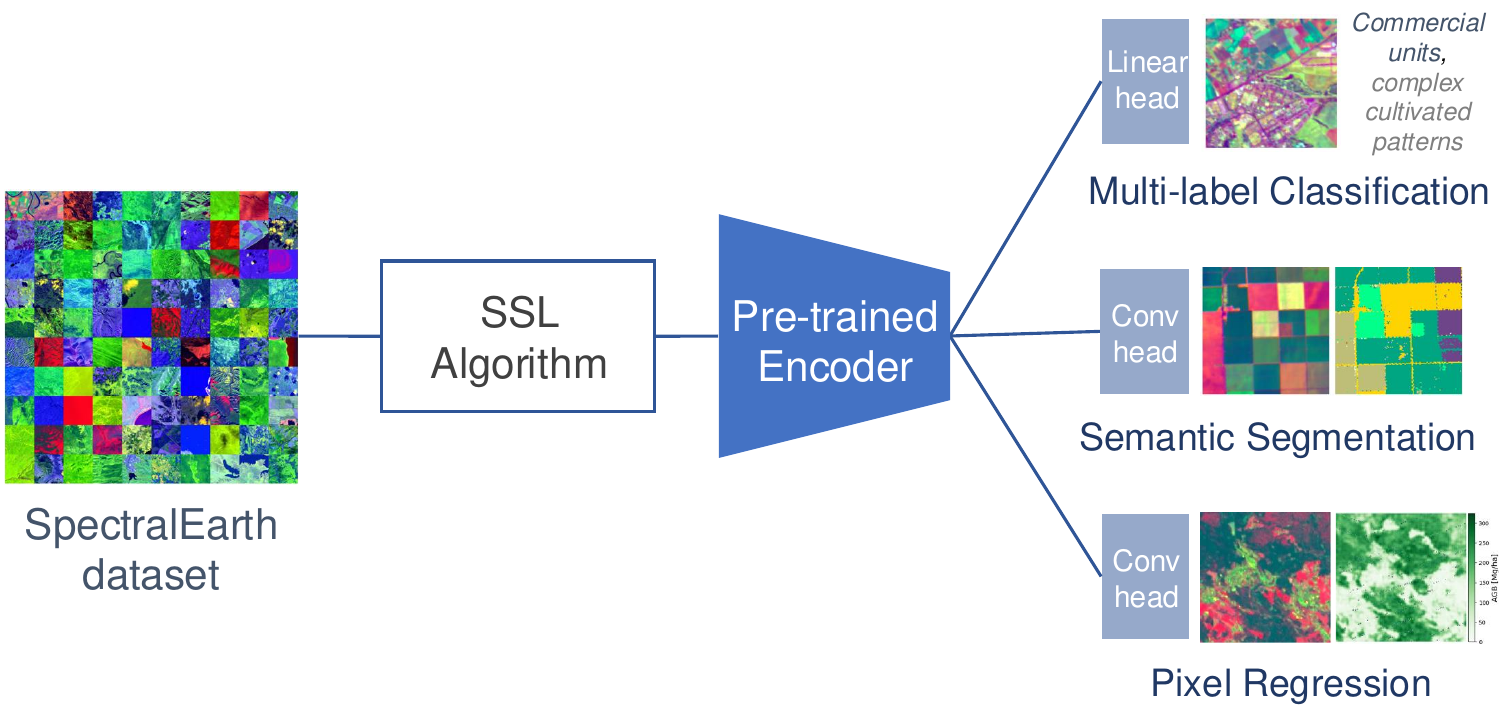}
    \caption{Schematic overview of the pre-training/fine-tuning pipeline. An encoder is pre-trained on SpectralEarth, and adapted to the several downstream tasks with task-specific heads.}
    \label{fig:main_pipeline}
\end{figure}

\section{Experimental Setup}
    This section introduces the details of pre-training and evaluation protocols. A schematic overview of the pipeline is provided in Figure~\ref{fig:main_pipeline}.
    \subsection{Self-Supervised Pre-training}
         We pre-train a spectral ResNet-50 (Spec.\ RN50) and a spectral ViT-S (Spec.\ ViT-S) with MoCo-V2, DINO, and MAE. For larger models, Spec.\ ViT-Base(B)/Large(L)/Huge(H)/Giant(g), we restrict the experiments to MAE, given the high computational cost. For ViT-g, we follow the implementation from~\cite{oquab2023dinov2}.
         We use the augmentations from SimCLR~\cite{chen2020simple} without color-jittering and gray-scale for DINO and MoCo-V2 to avoid distorting spectral information. When available, we use the temporal views as positive pairs for MoCo-V2 and DINO. For MAE, we mask $90\%$ of the tokens. We pretrain over 100 epochs for MoCo-V2 and DINO, and 200 epochs for MAE with a batch size of 256 images. We use a \texttt{StepLR} scheduler for MoCo-V2 and \texttt{CosineAnnealingLR} for DINO and MAE.
        
    \subsection{Downstream Tasks}
        \paragraph{Multi-label classification} We append a linear layer to the pretrained encoders and train the models with Binary Cross-Entropy loss for 100 epochs with a batch size of 128. AdamW optimizer was employed with a cosine annealing scheduler. Data augmentation included random resized crop, horizontal and vertical flipping.
        \paragraph{Semantic segmentation} We use a lightweight transfer protocol for the segmentation tasks, following previous works \cite{he2020momentum, grill2020bootstrap}. The pretrained encoders are converted into segmentation models as follows:
        \begin{itemize}
            \item \textit{ResNet}: We remove all strides from convolutions, replace them with atrous convolutions, and append two final convolutional layers.
            \item \textit{ViT}: We reshape the tokens from the last layer and add two additional convolutional layers with upscaling to recover the full spatial resolution.
        \end{itemize}
        We train for 100 epochs with a cross-entropy loss. AdamW optimizer was used with a batch size of 64 for EnMAP-NLCD/TreeMap and 32 for EnMAP-CDL/BDForet/EuroCrops and DESIS-CDL.  Data augmentation included random resized crop, horizontal and vertical flipping. 
        \paragraph{Parameter regression - Hyperview} We use the dataset from the Hyperview challenge \cite{nalepa2022hyperview} to evaluate cross-sensor transferability. Hyperview is a soil parameter estimation dataset comprising 1,732 hyperspectral patches for training and 1,154 for testing, each with 150 spectral bands. Each patch is annotated with four soil parameters, including potassium ($K$), phosphorus pentoxide ($P_2O_5$), magnesium ($Mg$), and acidity $pH$. We resized all images in the Hyperview dataset to 128$\times$128 pixels.
        For this downstream task, we append two fully connected layers to the backbones. The models are optimized based on the normalized MSE criterion defined in Section~\ref{sec:metrics} with the AdamW optimizer for 400 epochs and a batch size of 32. As for the other tasks, we used random resized crop, horizontal and vertical flipping during training.
        \paragraph{Pixel-wise regression - HyBiomass} We evaluate our models on the above ground forest biomass dataset HyBiomass~\cite{banze2025hybiomassglobalhyperspectralimagery}, which is a pixel regression task. The dataset consists of more than 34,000 patches distributed globally over forest areas. The above-ground biomass estimates are derived from the Global Ecosystem Dynamics Investigation (GEDI) lidar mission. We set the batch size to 128 and use the AdamW optimizer for 50 epochs. We use the same data augmentations as for the segmentation downstream tasks.

    \subsection{Evaluation Protocols}
        We consider multiple evaluation protocols with various levels of fine-tuning of the pretrained models:
        \begin{itemize}
        \item \textbf{Frozen Encoder}: Evaluates the quality of learned representations without fine-tuning the backbone. It is the most common evaluation protocol in the SSL literature~\cite{bardes2022vicregl,oquab2023dinov2}.
        \item \textbf{Fine-tuning}: Fine-tuning of all model parameters on downstream datasets to evaluate the transferability of the pretrained weights.
        \item \textbf{Fine-tuning Adapter}: 
        While frozen encoder evaluation is a common measure of the quality of learned representations, it does not yield optimal results.
        In remote sensing, it is often beneficial to adapt the model to the specific distribution of the downstream dataset, which can be influenced by several factors: geo-location, atmospheric conditions, seasonal changes, etc. Additionally, the relevance of individual spectral bands may vary depending on the downstream task. Therefore, we consider an intermediate setting where only the initial spectral adapter block is fine-tuned.

        \end{itemize}
    
    \subsection{Evaluation Metrics}
    \label{sec:metrics}
        Depending on the nature of the downstream task, we report the following metrics:
        \begin{itemize}
            \item \textbf{Classification:} The multi-label F1-score is reported for the EnMAP-CORINE dataset.
            \item \textbf{Semantic Segmentation:} The mean Intersection over Union (mIoU) is reported for all semantic segmentation datasets.
            \item \textbf{Regression:} For the Hyperview dataset, we follow the metric defined in the challenge~\cite{nalepa2022hyperview}: let $MSE_i, i \in [1, 4]$ be the mean squared error of the predictor for soil parameter $i$. Let $MSE_i^{base}$ be the corresponding MSE of a base predictor returning the mean value of the soil parameter, calculated over the training set. The reported normalized MSE is defined as: $MSE_{norm} = \sum_{i = 1}^4 MSE_i / MSE_i^{base}$.
            \item \textbf{Pixel-wise Regression:} we report the R² value for the HyBiomass dataset.
        \end{itemize}

   \subsection{Comparison Methods}
    
    We compare our pretrained models against several baselines and recent foundation models for hyperspectral imagery:
    
    \begin{itemize}
        \item \textbf{Random Frozen Encoder:} We use this baseline as a sanity check to assess whether the learned representations provide useful features for downstream tasks.
    
        \item \textbf{Training from Scratch:} A strong baseline that has shown competitive performance when trained with sufficient data and epochs~\cite{he2019rethinking}.
    
        \item \textbf{DOFA-(B/L)~\cite{xiong2024neural}:} A foundation model capable of adapting to various sensors given the input wavelengths. The model includes EnMAP data in its pretraining corpus. We evaluate both the base (DOFA-B) and large (DOFA-L) variants.
    
        \item \textbf{SpatSigma-(B/L)~\cite{wang2024hypersigmahyperspectralintelligencecomprehension}:} HyperSigma is a recent foundation model for hyperspectral imagery pretrained on Gaofen and Hyperion EO-1 data. We use only the spatial encoder for simplicity, as the original paper reports limited benefits when combining the spatial and spectral branches. We report results for the base and large variants and refer to them as SpatSigma-B and SpatSigma-L respectively. To mitigate the mismatch in the number of spectral bands, we resample the patch embedding layers with linear interpolation for our experiments.
    \end{itemize}
    

%% file: sec/6_experimental_results.tex
\section{Benchmark Results}
\label{sec:results}
    Our main results are summarized in Tables~\ref{tab:combined_performance} and \ref{tab:large_vits}. 
    We include random initialization baselines for each downstream task, along with our pretrained models. Due to hundreds of spectral channels of the SpectralEarth patches, a comparison with RGB-ImageNet weights or multispectral pretrained models is of little use.  

    \begin{table*}[htbp]
    \centering
    \resizebox{\textwidth}{!}{%
    \small
    \setlength{\tabcolsep}{5pt} 
    \renewcommand{\arraystretch}{1.25} 
    \begin{tabular}{@{}lllcccccccc@{}}
    \toprule
    \multicolumn{3}{c}{\textbf{Model Configuration}} & \multicolumn{1}{c}{\textbf{Classif. (F1)}} & \multicolumn{6}{c}{\textbf{Segmentation (mIoU)}} & \textbf{Pix Reg. (R²)} \\
    \cmidrule(lr){4-4}\cmidrule(lr){5-10}\cmidrule(l){11-11}
    Arch. & Weights & Protocol & CORINE & CDL & NLCD & EUROCROPS & TREEMAP & BDFORET & BNETD & HYBIOMASS \\
    \midrule
    \multirow{2}{*}{} 
      & \multirow{2}{*}{Random} 
          & Frozen    & 69.92 $\pm$ 0.51 & 62.65 $\pm$ 0.80 & 35.94 $\pm$ 0.38 & 54.83 $\pm$ 0.83 & 35.87 $\pm$ 0.23 & 57.11 $\pm$ 0.80 & 41.35 $\pm$ 0.30 & 0.30 $\pm$ 0.01 \\
      &                             
          & Full FT   & \textbf{77.77} $\pm$ \textbf{0.38} & 77.28 $\pm$ 0.13 & \underline{47.95} $\pm$ \underline{0.04} & 68.93 $\pm$ 1.41 & 46.47 $\pm$ 0.04 & \underline{75.50} $\pm$ \underline{0.42} & 48.97 $\pm$ 0.10 & 0.48 $\pm$ 0.00 \\
    \cmidrule(lr){2-11}
    \multirow{3}{*}{Spec. RN50} 
      & \multirow{3}{*}{MoCo-V2} 
          & Frozen    & 74.07 $\pm$ 0.14 & 70.97 $\pm$ 0.19 & 41.87 $\pm$ 0.06 & 61.70 $\pm$ 0.08 & 41.67 $\pm$ 0.05 & 67.60 $\pm$ 0.34 & 44.84 $\pm$ 0.05 & 0.39 $\pm$ 0.00 \\
      &                             
          & Full FT   & 77.43 $\pm$ 0.46 & \textbf{77.53} $\pm$ \textbf{0.25} & \textbf{48.11} $\pm$ \textbf{0.03} & \textbf{70.95} $\pm$ \textbf{0.63} & \underline{46.94} $\pm$ \underline{0.03} & \textbf{78.56} $\pm$ \textbf{0.32} & \underline{49.54} $\pm$ \underline{0.04} & \textbf{0.50} $\pm$ \textbf{0.00} \\
      &                             
          & Adapter   & 76.82 $\pm$ 0.39 & 75.84 $\pm$ 0.11 & 44.62 $\pm$ 0.03 & 66.71 $\pm$ 0.23 & 44.84 $\pm$ 0.04 & 73.86 $\pm$ 0.19 & 47.46 $\pm$ 0.04 & 0.45 $\pm$ 0.00 \\
    \cmidrule(lr){2-11}
    \multirow{3}{*}{} 
      & \multirow{3}{*}{DINO} 
          & Frozen    & 76.61 $\pm$ 0.31 & 71.02 $\pm$ 0.13 & 41.83 $\pm$ 0.03 & 60.29 $\pm$ 0.29 & 42.03 $\pm$ 0.03 & 67.07 $\pm$ 0.11 & 44.32 $\pm$ 0.02 & 0.40 $\pm$ 0.00 \\
      &                             
          & Full FT   & \underline{77.71} $\pm$ \underline{0.55} & \underline{77.49} $\pm$ \underline{0.23} & 47.62 $\pm$ 0.07 & \underline{69.52} $\pm$ \underline{1.00} & \textbf{47.08} $\pm$ \textbf{0.13} & 74.68 $\pm$ 0.90 & \textbf{49.74} $\pm$ \textbf{0.31} & \underline{0.50} $\pm$ \underline{0.00} \\
      &                             
          & Adapter   & 76.96 $\pm$ 0.37 & 75.51 $\pm$ 0.15 & 44.38 $\pm$ 0.03 & 66.62 $\pm$ 0.35 & 44.70 $\pm$ 0.07 & 73.90 $\pm$ 0.11 & 47.21 $\pm$ 0.07 & 0.45 $\pm$ 0.00 \\
    \cmidrule(lr){1-11}
    \multirow{2}{*}{} 
      & \multirow{2}{*}{Random} 
          & Frozen    & 70.59 $\pm$ 0.62 & 65.96 $\pm$ 0.72 & 36.94 $\pm$ 0.30 & 55.78 $\pm$ 0.52 & 38.72 $\pm$ 0.25 & 61.89 $\pm$ 0.85 & 41.55 $\pm$ 0.22 & 0.34 $\pm$ 0.00 \\
      &                             
          & Full FT   & 77.63 $\pm$ 0.37 & 74.77 $\pm$ 0.22 & 45.31 $\pm$ 0.62 & 66.87 $\pm$ 0.62 & 44.43 $\pm$ 0.13 & 74.00 $\pm$ 0.49 & 46.57 $\pm$ 0.11 & 0.40 $\pm$ 0.02 \\
    \cmidrule(lr){2-11}
    \multirow{3}{*}{} 
      & \multirow{3}{*}{MoCo-V2} 
          & Frozen    & 73.99 $\pm$ 0.04 & 71.40 $\pm$ 0.32 & 39.99 $\pm$ 0.08 & 60.21 $\pm$ 0.39 & 41.76 $\pm$ 0.06 & 68.50 $\pm$ 0.12 & 43.50 $\pm$ 0.05 & 0.40 $\pm$ 0.00 \\
      &                             
          & Full FT   & 78.37 $\pm$ 0.34 & 75.59 $\pm$ 0.14 & \underline{45.92} $\pm$ \underline{0.05} & \underline{67.27} $\pm$ \underline{0.29} & 45.03 $\pm$ 0.10 & \underline{75.24} $\pm$ \underline{0.28} & 47.02 $\pm$ 0.07 & 0.43 $\pm$ 0.01 \\
      &                             
          & Adapter   & 76.72 $\pm$ 0.35 & 75.28 $\pm$ 0.19 & 43.18 $\pm$ 0.08 & 66.46 $\pm$ 0.21 & 44.09 $\pm$ 0.07 & 73.85 $\pm$ 0.23 & 45.73 $\pm$ 0.10 & 0.44 $\pm$ 0.00 \\
    \cmidrule(lr){2-11}
    \multirow{3}{*}{Spec. ViT-S} 
      & \multirow{3}{*}{DINO} 
          & Frozen    & 75.11 $\pm$ 0.13 & 71.56 $\pm$ 0.35 & 40.47 $\pm$ 0.06 & 58.76 $\pm$ 0.27 & 41.89 $\pm$ 0.13 & 68.16 $\pm$ 0.43 & 43.12 $\pm$ 0.10 & 0.40 $\pm$ 0.00 \\
      &                             
          & Full FT   & \underline{78.67} $\pm$ \underline{0.34} & \underline{76.10} $\pm$ \underline{0.31} & 45.64 $\pm$ 0.15 & 65.56 $\pm$ 0.76 & \underline{45.44} $\pm$ \underline{0.15} & 73.56 $\pm$ 0.41 & \underline{47.11} $\pm$ \underline{0.14} & \underline{0.48} $\pm$ \underline{0.00} \\
      &                             
          & Adapter   & 77.00 $\pm$ 0.50 & 74.83 $\pm$ 0.07 & 42.86 $\pm$ 0.08 & 64.78 $\pm$ 0.65 & 43.54 $\pm$ 0.11 & 72.38 $\pm$ 0.71 & 44.99 $\pm$ 0.17 & 0.44 $\pm$ 0.00 \\
    \cmidrule(lr){2-11}
    \multirow{4}{*}{} 
      & \multirow{3}{*}{MAE} 
          & Frozen    & 72.85 $\pm$ 0.20 & 72.01 $\pm$ 0.28 & 41.08 $\pm$ 0.05 & 62.14 $\pm$ 0.18 & 41.19 $\pm$ 0.15 & 68.92 $\pm$ 0.73 & 43.34 $\pm$ 0.26 & 0.42 $\pm$ 0.00 \\
      &                             
          & Full FT   & \textbf{79.17} $\pm$ \textbf{0.55} & \textbf{77.16} $\pm$ \textbf{0.31} & \textbf{47.61} $\pm$ \textbf{0.10} & \textbf{69.87} $\pm$ \textbf{0.16} & \textbf{45.80} $\pm$ \textbf{0.15} & \textbf{76.54} $\pm$ \textbf{0.23} & \textbf{49.36} $\pm$ \textbf{0.07} & \textbf{0.51} $\pm$ \textbf{0.01} \\
      &                             
          & Adapter   & 76.75 $\pm$ 0.35 & 75.21 $\pm$ 0.21 & 43.64 $\pm$ 0.07 & 67.18 $\pm$ 0.58 & 43.45 $\pm$ 0.14 & 74.89 $\pm$ 0.47 & 45.89 $\pm$ 0.12 & 0.45 $\pm$ 0.01 \\
    \bottomrule
    \end{tabular}%
    }
    \caption{Benchmark results of the SpectralEarth downstream tasks for Spec. RN50 and Spec. ViT-S under different evaluation protocols: frozen backbone, full fine-tuning, and fine-tuning the spectral adapter only. The best score for \textit{each dataset-backbone pair} is in \textbf{bold}, and the second best is \underline{underlined}.}
    \label{tab:combined_performance}
    \end{table*}

    \subsection{Comparing SSL Algorithms}
    Table~\ref{tab:combined_performance} summarizes our results for three SSL algorithms with the Spec. RN50 and Spec. ViT-S backbones.
    
    \paragraph{Multi-label classification} 
    Pretrained models outperform random weights in linear evaluation for MoCo-V2 and DINO on EnMAP-CORINE, with DINO yielding the best results, demonstrating the ability of joint-embedding methods to learn robust representations, even in the absence of fine-tuning.  

    Additionally, we observe that MAE under-performs in linear probing, which is consistent with existing literature~\cite{lehner2023contrastive}. This can be attributed to the reconstruction pre-training objective, which focuses on low-level features, whereas joint-embedding methods learn more global and semantic representations. For full fine-tuning, CNN pretrained models with MoCo and DINO are on par with/slightly worse than training from scratch. In contrast, the Spec. ViT-S model benefits from pre-training in all settings compared to random initialization, with MAE yielding the best results.  

    Fine-tuning the spectral adapter yields a notable performance boost compared to linear evaluation. This observation aligns with reports in the literature~\cite{lee2022surgical}: fine-tuning early network layers is an efficient compromise between frozen weights vs. complete fine-tuning.  
    In terms of F1 score, only fine-tuning the adapter ($0.3\%$ of network parameters) is, in most settings, less than one point below training from scratch, highlighting the efficient adaptability of our pretrained models. 

    \paragraph{Semantic segmentation} 
    Pretrained models with a frozen encoder consistently outperform random weights across all segmentation tasks. This shows that, without any fine-tuning, the models have learned useful features for segmentation. MoCo-V2 and DINO pre-trained models yield comparable results for both the CNN and the ViT backbones. However, MAE, as a feature extractor, consistently outperforms MoCo-V2 and DINO for Spec. ViT-S. This can be attributed to the pixel-level pre-training objective, which is more suitable for segmentation tasks (as opposed to image-level problems). 

    For complete fine-tuning, MoCo-V2 pre-trained CNNs consistently outperform training from scratch on all segmentation tasks, while DINO is on par with the random initialization baseline. Notably, the gaps are more pronounced for BDForet and EuroCrops, where MoCo-V2 yields an improvement of 3 and 2 points of mIoU, respectively. For the ViT model, MAE + fine-tuning performs best in all settings, yielding consistent improvements of 2 to 3 points of mIoU. Depending on the dataset, the second-best models are either MoCo-V2 or DINO, with smaller improvements compared to training from scratch.  
    
    Fine-tuning the spectral adapter significantly outperforms frozen encoder performance in all settings, showing the importance of tailoring the spectral feature extraction to the characteristics of the downstream task. Notably, fine-tuning the spectral adapter outperforms training from scratch on CDL, EuroCrops, and BD-Foret.  

    Compared to CNNs, ViTs perform slightly worse on segmentation problems. This can be attributed to multiple factors: 
    
    \begin{enumerate}
    \item Training ViTs from scratch results in substantially lower performance than CNNs due to the lack of strong inductive biases. While pre-training significantly mitigates this gap, CNNs still retain a slight edge after fine-tuning; 
    \item The spatial patterns of the masks, which present high-frequency features partly due to label noise, are harder to learn with a ViT due to the fixed patch size compared to CNNs, which can preserve full-resolution feature maps. This effect can be better observed in the patch size experiment; see Figure~\ref{fig:patch_size_ablation}. 
    \end{enumerate}

    \paragraph{Pixel-wise regression} The results on HyBiomass are in line with the segmentation downstream tasks. Our pre-trained frozen encoders significantly outperform the random encoder baseline for all SSL methods and backbones.
    Fine-tuning the spectral adapter yields a significant performance boost compared to the frozen encoder setting, and full fine-tuning works best. MoCo and DINO provide consistent improvements with fine-tuning over training from scratch for both the CNN and ViT backbones. The largest gain from pre-training is observed for Spec. ViT-S with MAE, which achieves an R² improvement of 0.11 compared to training from scratch. This configuration also yields the overall best result. Moreover, both the frozen encoder and the adapter-only fine-tuning outperform training from scratch for Spec. ViT-S with MAE. This is consistent with our segmentation results: without pre-trained weights, CNNs significantly outperform ViTs, which can be attributed to the lack of strong inductive biases in ViTs, requiring larger datasets to perform well.

    \begin{table*}[htbp]
    \centering
    \resizebox{\textwidth}{!}{%
    \small
    \setlength{\tabcolsep}{5pt} 
    \renewcommand{\arraystretch}{1.25} 
    \begin{tabular}{@{}llcccccccc@{}}
    \toprule
    \multicolumn{2}{c}{\textbf{Model Configuration}} & \multicolumn{1}{c}{\textbf{Classif. (F1)}} & \multicolumn{6}{c}{\textbf{Segmentation (mIoU)}} & \textbf{Pix Reg. (R²)} \\
    \cmidrule(lr){1-2}\cmidrule(lr){3-3}\cmidrule(lr){4-9}\cmidrule(l){10-10}
    Arch. & Protocol & CORINE & CDL & NLCD & EUROCROPS & TREEMAP & BDFORET & BNETD & HYBIOMASS \\
    \midrule
    \multicolumn{10}{c}{\textbf{Spec. ViT Models}} \\
    \midrule
    \multirow{4}{*}{Spec. ViT-B} 
      & Supervised   & 77.17 $\pm$ 0.72 & 74.65 $\pm$ 0.60  & 45.57 $\pm$ 0.14  & 66.14 $\pm$ 0.43  & 44.51 $\pm$ 0.13  & 73.55 $\pm$ 0.39  & 46.76 $\pm$ 0.10 & 0.37 $\pm$ 0.02 \\
      & Frozen     & 75.04 $\pm$ 0.30 & 72.20 $\pm$ 0.16  & 40.82 $\pm$ 0.14  & 61.29 $\pm$ 0.22  & 39.74 $\pm$ 0.27  & 68.39 $\pm$ 0.20  & 43.37 $\pm$ 0.10 & 0.38 $\pm$ 0.01 \\
      & Full FT         & \underline{79.49} $\pm$ \underline{0.26} & \textbf{77.44} $\pm$ \textbf{0.26}  & 47.59 $\pm$ 0.23  & \textbf{69.34} $\pm$ \textbf{0.43}  & \underline{46.10} $\pm$ \underline{0.04}  & \underline{76.30} $\pm$ \underline{0.52}  & \textbf{49.46} $\pm$ \textbf{0.06} & \textbf{0.51} $\pm$ \textbf{0.00} \\
      & Adapter    & 77.63 $\pm$ 0.25 & 75.30 $\pm$ 0.32  & 43.81 $\pm$ 0.04  & 67.31 $\pm$ 0.43  & 43.65 $\pm$ 0.18  & 74.51 $\pm$ 0.15  & 45.52 $\pm$ 0.11 & 0.44 $\pm$ 0.01 \\
    \midrule
    \multirow{4}{*}{Spec. ViT-L} 
      & Supervised   & 76.85 $\pm$ 0.67 & 74.28 $\pm$ 0.34  & 45.49 $\pm$ 0.13  & 65.66 $\pm$ 0.46  & 44.52 $\pm$ 0.08  & 72.93 $\pm$ 0.59  & 46.66 $\pm$ 0.04 & 0.38 $\pm$ 0.02 \\
      & Frozen     & 73.98 $\pm$ 0.21 & 72.72 $\pm$ 0.23  & 42.67 $\pm$ 0.03  & 62.13 $\pm$ 0.18  & 42.62 $\pm$ 0.19  & 71.07 $\pm$ 0.31  & 43.96 $\pm$ 0.18 & 0.42 $\pm$ 0.00 \\
      & Full FT         & 79.25 $\pm$ 0.65 & \underline{77.41} $\pm$ \underline{0.26}  & \underline{47.84} $\pm$ \underline{0.03}  & 68.73 $\pm$ 0.36  & \textbf{46.10} $\pm$ \textbf{0.17}  & 76.26 $\pm$ 0.73  & 48.96 $\pm$ 0.39 & \underline{0.49} $\pm$ \underline{0.00} \\
      & Adapter    & 77.24 $\pm$ 0.07 & 75.94 $\pm$ 0.13  & 44.15 $\pm$ 0.07  & 67.69 $\pm$ 0.32  & 44.09 $\pm$ 0.09  & 74.89 $\pm$ 0.26  & 46.11 $\pm$ 0.12 & 0.45 $\pm$ 0.00 \\
    \midrule
    \multirow{4}{*}{Spec. ViT-H} 
      & Supervised   & 77.18 $\pm$ 0.44 & 73.72 $\pm$ 0.47  & 45.33 $\pm$ 0.12  & 64.77 $\pm$ 0.48  & 44.45 $\pm$ 0.17  & 72.98 $\pm$ 0.45  & 46.62 $\pm$ 0.14 & 0.34 $\pm$ 0.05 \\
      & Frozen     & 73.96 $\pm$ 0.30 & 71.68 $\pm$ 0.21  & 41.87 $\pm$ 0.13  & 61.94 $\pm$ 0.11  & 41.95 $\pm$ 0.17  & 68.90 $\pm$ 0.12  & 43.48 $\pm$ 0.14 & 0.38 $\pm$ 0.00 \\
      & Full FT         & \textbf{79.51} $\pm$ \textbf{0.30} & 77.26 $\pm$ 0.18  & \textbf{47.85} $\pm$ \textbf{0.14}  & \underline{69.13} $\pm$ \underline{0.49}  & 46.05 $\pm$ 0.13  & \textbf{76.57} $\pm$ \textbf{0.35}  & \underline{49.38} $\pm$ \underline{0.23} & 0.46 $\pm$ 0.01 \\
      & Adapter    & 77.23 $\pm$ 0.32 & 75.34 $\pm$ 0.22  & 43.90 $\pm$ 0.08  & 66.51 $\pm$ 0.14  & 43.75 $\pm$ 0.06  & 74.13 $\pm$ 0.23  & 45.56 $\pm$ 0.09 & 0.45 $\pm$ 0.01 \\
    \midrule
    \multirow{4}{*}{Spec. ViT-g} 
      & Supervised  & 76.94 $\pm$ 0.61 & 73.44 $\pm$ 0.38  & 45.03 $\pm$ 0.04  & 64.40 $\pm$ 0.40  & 44.38 $\pm$ 0.11  & 72.83 $\pm$ 0.50  & 46.34 $\pm$ 0.07 & 0.39 $\pm$ 0.01 \\
      & Frozen     & 72.50 $\pm$ 0.26 & 68.39 $\pm$ 0.20  & 39.68 $\pm$ 0.03  & 57.11 $\pm$ 0.13  & 40.21 $\pm$ 0.07  & 62.99 $\pm$ 0.09  & 41.94 $\pm$ 0.08 & 0.36 $\pm$ 0.00 \\
      & Full FT         & 78.20 $\pm$ 0.66 & 76.14 $\pm$ 0.23  & 46.61 $\pm$ 0.17  & 66.75 $\pm$ 0.49  & 45.18 $\pm$ 0.12  & 73.47 $\pm$ 0.58  & 48.26 $\pm$ 0.06 & 0.35 $\pm$ 0.07 \\
      & Adapter    & 77.26 $\pm$ 0.50 & 75.25 $\pm$ 0.24  & 43.17 $\pm$ 0.09  & 66.84 $\pm$ 0.29  & 43.69 $\pm$ 0.05  & 74.06 $\pm$ 0.47  & 45.10 $\pm$ 0.06 & 0.41 $\pm$ 0.01 \\
    \midrule
    \multicolumn{10}{c}{\textbf{SOTA Models}} \\
    \midrule
    \multirow{2}{*}{DOFA-B} 
      & Frozen    & 67.82 $\pm$ 0.15 & 58.99 $\pm$ 0.13 & 33.47 $\pm$ 0.07 & 45.44 $\pm$ 0.51 & 34.60 $\pm$ 0.20 & 53.22 $\pm$ 0.43 & 37.04 $\pm$ 0.10 & 0.28 $\pm$ 0.00 \\
      & Full FT   & 74.47 $\pm$ 0.66 & 69.11 $\pm$ 0.30 & 42.60 $\pm$ 0.06 & 56.99 $\pm$ 0.25 & 41.98 $\pm$ 0.05 & 64.90 $\pm$ 0.17 & 42.31 $\pm$ 0.06 & 0.40 $\pm$ 0.00 \\
    \midrule
    \multirow{2}{*}{DOFA-L} 
      & Frozen    & 67.48 $\pm$ 0.13 & 60.64 $\pm$ 0.27 & 33.42 $\pm$ 0.08 & 46.11 $\pm$ 0.17 & 34.67 $\pm$ 0.14 & 51.26 $\pm$ 0.37 & 37.37 $\pm$ 0.08 & 0.28 $\pm$ 0.00 \\
      & Full FT   & 74.72 $\pm$ 0.32 & 69.40 $\pm$ 0.44 & 43.06 $\pm$ 0.07 & 56.54 $\pm$ 0.55 & 42.18 $\pm$ 0.17 & 64.80 $\pm$ 0.37 & 43.06 $\pm$ 0.05 & 0.41 $\pm$ 0.00 \\
    \midrule
    \multirow{2}{*}{SpatSigma-B} 
      & Frozen    & 68.16 $\pm$ 0.33 & 63.49 $\pm$ 0.09 & 36.02 $\pm$ 0.06 & 50.67 $\pm$ 0.36 & 36.99 $\pm$ 0.09 & 58.20 $\pm$ 0.26 & 38.62 $\pm$ 0.12 & 0.31 $\pm$ 0.00 \\
      & Full FT   & 76.71 $\pm$ 0.32 & 70.63 $\pm$ 0.55 & 43.18 $\pm$ 0.07 & 58.32 $\pm$ 0.19 & 42.93 $\pm$ 0.09 & 69.27 $\pm$ 0.35 & 42.54 $\pm$ 0.06 & 0.41 $\pm$ 0.00 \\
    \midrule
    \multirow{2}{*}{SpatSigma-L} 
      & Frozen    & 68.37 $\pm$ 0.33 & 61.72 $\pm$ 0.22 & 34.42 $\pm$ 0.30 & 47.86 $\pm$ 0.28 & 35.70 $\pm$ 0.09 & 54.94 $\pm$ 0.31 & 37.85 $\pm$ 0.06 & 0.27 $\pm$ 0.03 \\
      & Full FT   & 76.66 $\pm$ 0.29 & 70.29 $\pm$ 0.21 & 43.31 $\pm$ 0.07 & 56.83 $\pm$ 1.01 & 42.60 $\pm$ 0.10 & 67.98 $\pm$ 0.45 & 42.96 $\pm$ 0.08 & 0.42 $\pm$ 0.00 \\
    \bottomrule
    \end{tabular}%
    }
    \caption{Benchmark results for large vision transformers pre-trained on SpectralEarth with MAE. Models are evaluated under three protocols: frozen encoder, full fine-tuning, and fine-tuned adapter. Training from scratch is reported as a baseline. State-of-the-art models DOFA and HyperSigma are reported for comparison. The best score for each dataset is highlighted in \textbf{bold}, and the second best in \underline{underlined}.}
    \label{tab:large_vits}
    \end{table*}

    \subsection{Scaling-up Pretrained Models}
    Recent studies have shown significant performance gains through model scaling in SSL~\cite{he2022masked,oquab2023dinov2}. Consequently, recent geospatial foundation models are exploring ViT architectures with hundreds of millions of parameters~\cite{wang2024hypersigmahyperspectralintelligencecomprehension,hong2024spectralgpt,jakubik2023foundationmodelsgeneralistgeospatial}. We investigated model scaling on SpectralEarth by pre-training Spec. ViT architectures of varying sizes: Base (B), Large (L), Huge (H), and Giant (g). Our findings are presented in Table~\ref{tab:large_vits}. 
    
    We observe that MAE consistently outperforms training from scratch across all model sizes and datasets. Notably, scaling from Spec. ViT-B to ViT-L improves frozen encoder performance on multiple downstream tasks, including NLCD, EuroCrops, TreeMap, BDForet, and HyBiomass. Fine-tuning the adapter — only 56K additional parameters — surpasses training from scratch on CORINE, CDL, EuroCrops, BDForet, and HyBiomass. While scaling provides limited performance gains for some downstream tasks up to Spec. ViT-H, we find that Spec. ViT-B offers the best trade-off between performance and computational cost. Further increases in model size, as seen for Spec. ViT-g, degrades performance, possibly due to the size of the pretraining dataset. Although SpectralEarth is significantly larger than previous HSI datasets, it remains smaller than the datasets typically used to train billion-parameter models in computer vision~\cite{oquab2023dinov2}. This trend can also be observed for models such as DOFA and HyperSigma, where the \textit{large} variants provide limited benefits over the \textit{base} backbones. Our models outperform both DOFA and HyperSigma across all tasks and protocols, which we attribute to the architectural design of our backbones that better preserves fine-grained spatial information and the large EnMAP corpus in SpectralEarth. In particular, HyperSigma was trained on Hyperion EO-1 and Gaofen-5B, making it less competitive in EnMAP downstream tasks.

    \subsection{Efficient Fine-tuning}

\begin{figure}[htbp]
            \centering
            \begin{subfigure}{0.49\linewidth}
                \includegraphics[width=\linewidth]{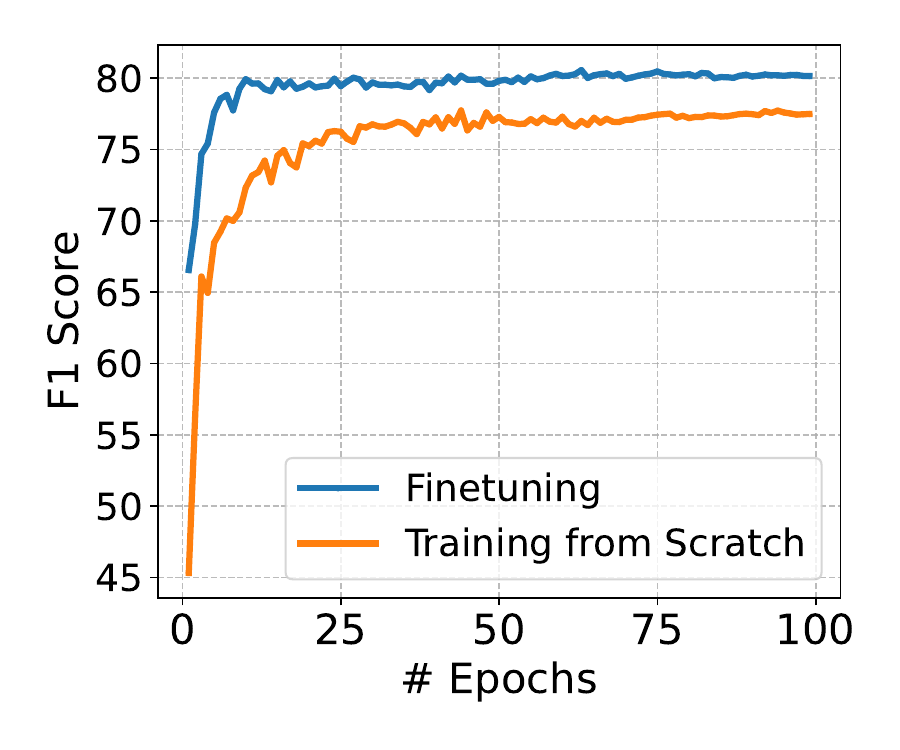}
                \caption{EnMAP-CORINE}
                \label{fig:ablation_convergence_speed_corine}
            \end{subfigure}
            \hfill 
            \begin{subfigure}{0.49\linewidth}
                \includegraphics[width=\linewidth]{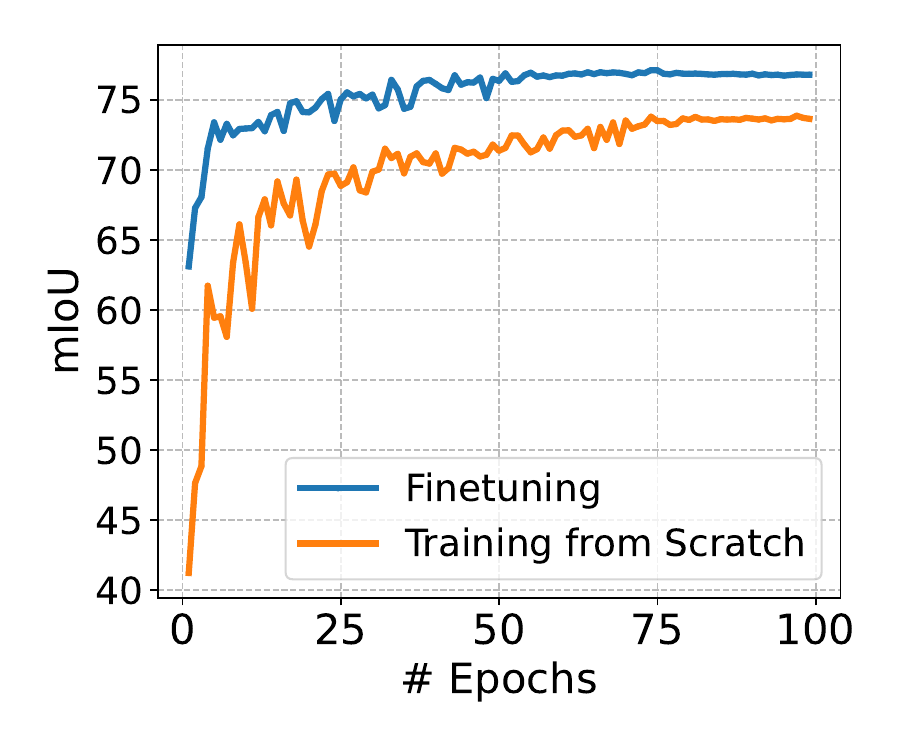}
                \caption{EnMAP-CDL}
                \label{fig:ablation_convergence_speed_cdl}
            \end{subfigure}
            \caption{Comparison of convergence speed for fine-tuning and training from scratch on EnMAP-CORINE and EnMAP-CDL for a Spec. RN50 backbone.}
        \label{fig:combined_convergence_ablation_study}
        \end{figure}

    \paragraph{Model convergence} Using a pretrained model can speed up convergence during fine-tuning, hence reducing the computational cost of training deep neural networks. To assess this for our models, we compare the training efficiency of Spec. RN50 models pretrained with DINO relative to training from scratch on the CORINE and CDL benchmarks. Figure~\ref{fig:combined_convergence_ablation_study} displays the evolution of the validation metrics across training epochs.
    We observe that pretrained models converge more rapidly, particularly in the early stages of training. Notably, fine-tuning matches the 100 epochs performance of training from scratch in $\sim10$  epochs for EnMAP-CORINE and EnMAP-CDL.
    These results highlight the computational advantage of using our pretrained models compared to training from scratch.
         
\begin{figure}[htbp]
            \centering
            \begin{subfigure}[b]{0.49\linewidth}
                \includegraphics[width=\linewidth]{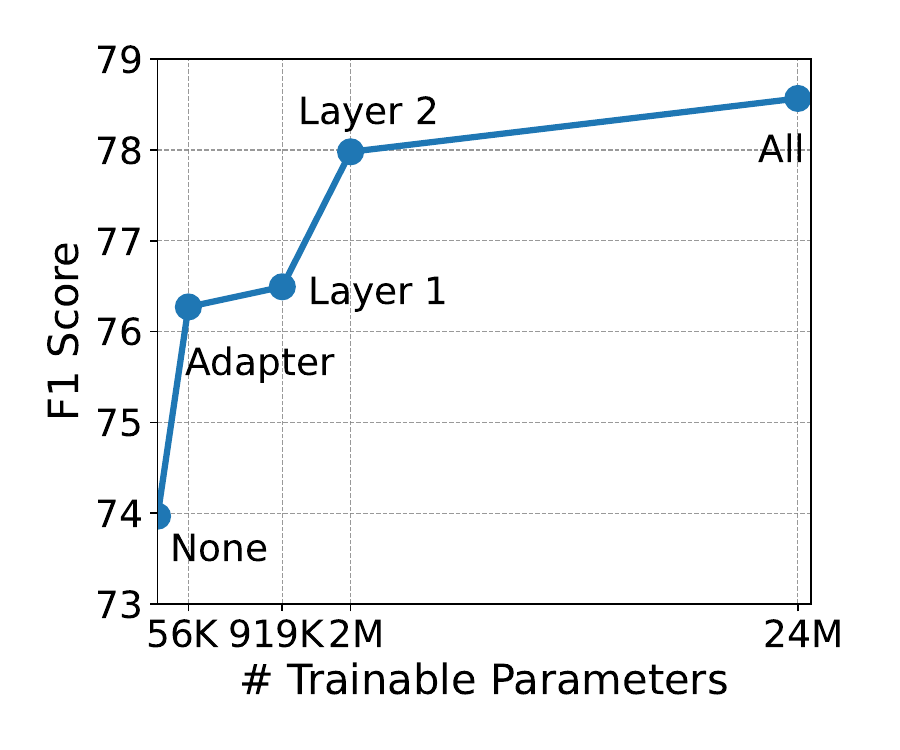}
                \caption{EnMAP-CORINE}
                \label{fig:enmap_corine_ablation}
            \end{subfigure}
            \hfill 
            \begin{subfigure}[b]{0.49\linewidth}
                \includegraphics[width=\linewidth]{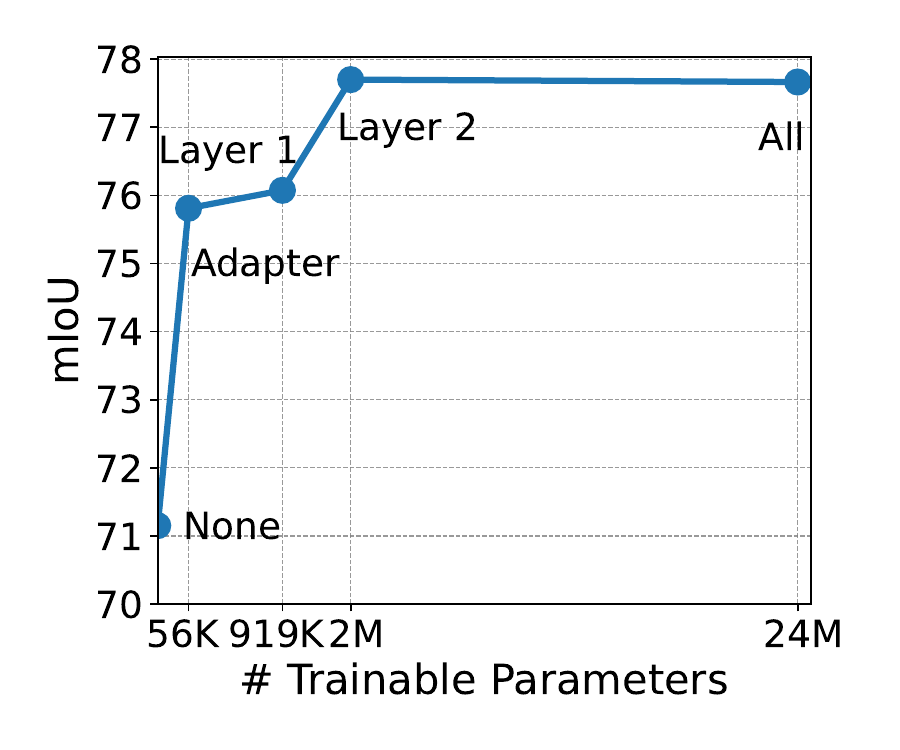}
                \caption{EnMAP-CDL}
                \label{fig:enmap_cdl_ablation}
            \end{subfigure}
            \caption{Impact of incremental parameter unfreezing on EnMAP-CORINE and EnMAP-CDL with a pre-trained Spec. RN50 model.}
            \label{fig:finetuning_ablation_study_combined}
        \end{figure}

    \paragraph{Parameter-efficient fine-tuning} Foundation models are typically evaluated in linear probing to assess the quality of the learnt representations \cite{he2020momentum}. While a frozen encoder can achieve competitive performance, fine-tuning often surpasses it---at a higher computational cost. We experiment with the progressive unfreezing of network layers as a compromise between both scenarios. As depicted in Figure~\ref{fig:finetuning_ablation_study_combined}, fine-tuning the first two blocks of a pretrained Spec. RN50 produces results that closely resemble full fine-tuning. This demonstrates the cost- and energy efficiency of our pretrained models for downstream tasks.

\begin{figure}[htbp]
            \centering
            \begin{subfigure}[b]{0.49\linewidth}
                \includegraphics[width=\linewidth]{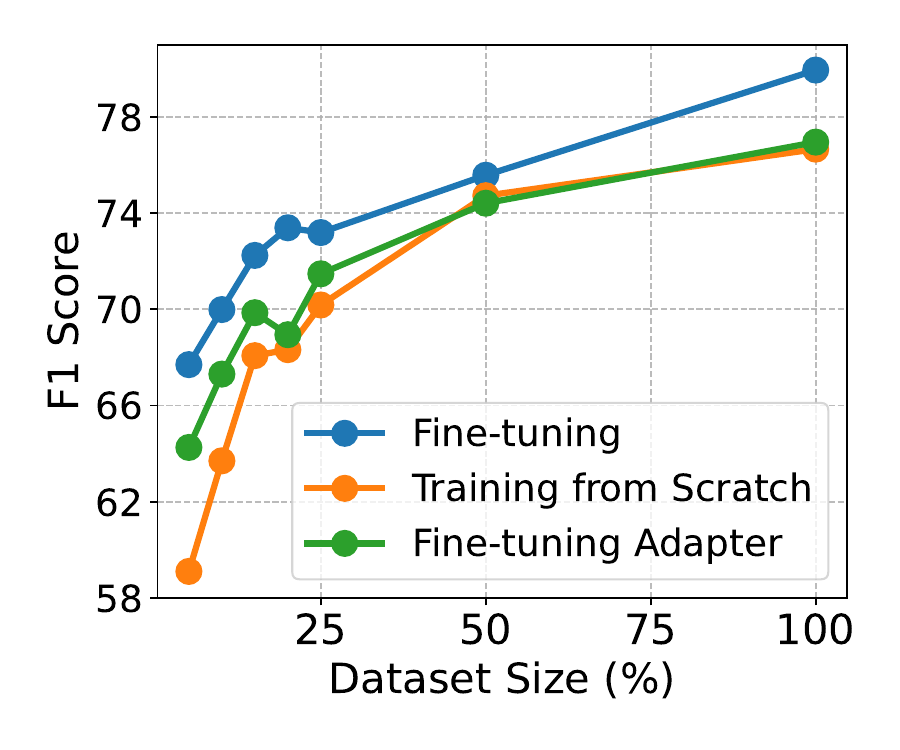}
                \caption{EnMAP-CORINE}
                \label{fig:ablation_subset_size_corine}
            \end{subfigure}
            \hfill 
            \begin{subfigure}[b]{0.49\linewidth}
                \includegraphics[width=\linewidth]{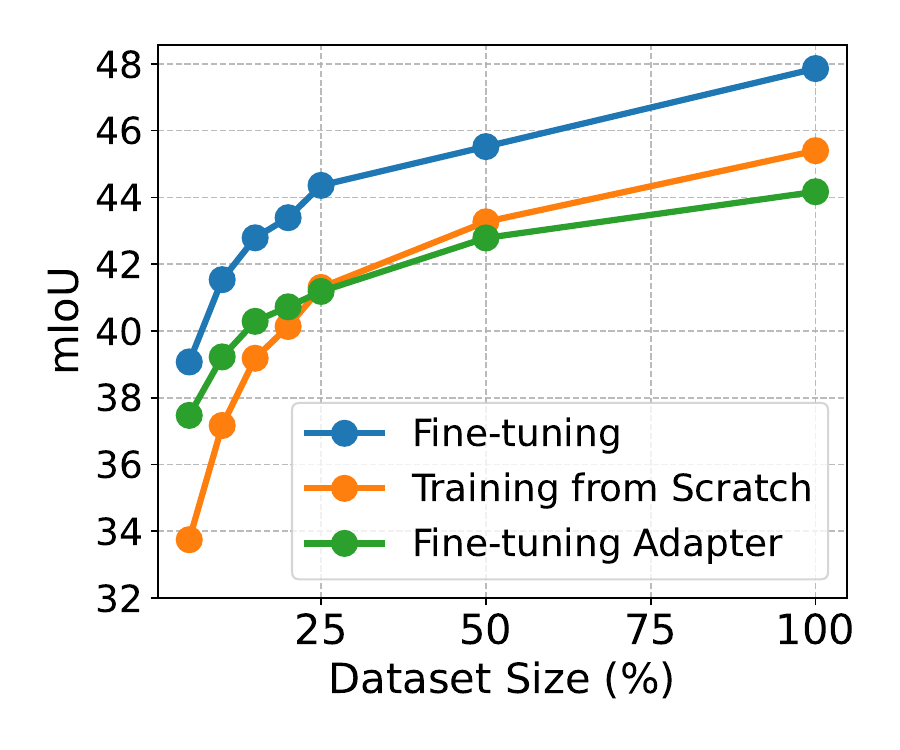}
                \caption{EnMAP-NLCD}
                \label{fig:ablation_subset_size_nlcd}
            \end{subfigure}
            \caption{Impact of the size of the downstream dataset evaluated on EnMAP-CORINE and EnMAP-NLCD for the Spec. ViT-L model.}
            \label{fig:ablation_subset_size}
        \end{figure}

    \paragraph{Training with limited labels} Pre-trained models are very useful when labeled data is scarce~\cite{newell2020useful}. To assess this, we evaluate the performance of the Spec. ViT-L model, pre-trained with MAE, on varying subset sizes of EnMAP-CORINE and EnMAP-NLCD datasets.
    The results, reported in Figure~\ref{fig:ablation_subset_size}, demonstrate the benefits of pre-training across all data regimes. In particular, we observe large gaps of 9 points in F1 score for EnMAP-CORINE, and 5 points in mIoU for EnMAP-NLCD, when using $5\%$ of the labels compared to training from scratch. Interestingly, in lower data regime scenarios, fine-tuning the adapter only suffices to outperform training from scratch. This demonstrates the generalizability of the features learned from pre-training.  

    \subsection{Cross-Sensor Transferability}
    
    To evaluate the generalizability of SpectralEarth-pretrained models across sensors, we assess their performance on three datasets acquired from sensors not seen during pretraining: EO-1 Hyperion (EO1-CDL), DESIS (DESIS-CDL), and Intuition-1 (Hyperview)~\footnote{The data for the hyperview challenge was collected using an airborne sensor that mimics the spectral characteristics of the  Intuition-1 mission.}. These datasets present different spectral characteristics from EnMAP. 
    
    We evaluate Spec. ViT-B and ViT-L models pretrained with MAE and compare them against training from scratch, as well as DOFA and HyperSigma. Our results are summarized in Table~\ref{tab:cross_sensor_transferability}. 

    On the CDL-based tasks (EO-1 and DESIS), our models consistently outperform both DOFA and HyperSigma across all protocols. Notably, our frozen encoders outperform HyperSigma EO-1, despite not being pre-trained on EO-1 imagery. This highlights the transferability of features learned from SpectralEarth. Given the relatively small size of the CDL datasets, fine-tuning the adapter performs on par with full fine-tuning and significantly outperforms training from scratch.

    On the Hyperview dataset, the best performance is obtained using the frozen Spec. ViT-L encoder. We attribute this to severe overfitting when fine-tuning due to the limited training data and the difficulty of the task. This result demonstrates the practicality of using our pre-trained models as frozen feature extractors. 

    \begin{table*}[htbp]
    \centering
    \small
    \setlength{\tabcolsep}{5pt} 
    \begin{tabular}{@{}llccc@{}}
    \toprule
    \multicolumn{2}{c}{\textbf{Model Configuration}} & \multicolumn{1}{c}{\textbf{Hyperview ($MSE_{Norm}$ $\downarrow$)}} & \multicolumn{2}{c}{\textbf{CDL (mIoU $\uparrow$)}} \\
    \cmidrule(lr){1-2}\cmidrule(lr){3-3}\cmidrule(lr){4-5}
    Backbone & Protocol & Intuition-1 & EO-1 & DESIS \\
    \midrule
    \multirow{4}{*}{SpecViT-B}
     & From Scratch   & 0.88 $\pm$ 0.01 & 65.85 $\pm$ 0.78 & 66.18 $\pm$ 0.50 \\
      & Frozen   & 0.85 $\pm$ 0.01 & 59.02 $\pm$ 0.61 & 62.22 $\pm$ 0.61 \\
      & Full FT   & 0.86 $\pm$ 0.00 & \textbf{66.69} $\pm$ \textbf{0.91} & \underline{68.50} $\pm$ \underline{0.78} \\
      & Adapter   & \underline{0.85} $\pm$ \underline{0.00} & 66.17 $\pm$ 0.58 & 67.74 $\pm$ 0.29 \\
    \cmidrule(lr){1-5}
    \multirow{4}{*}{SpecViT-L}
     & From Scratch   & 0.89 $\pm$ 0.01 & 65.09 $\pm$ 0.76 & 66.06 $\pm$ 0.57 \\
    & Frozen   & \textbf{0.81} $\pm$ \textbf{0.02} & 59.46 $\pm$ 1.02 & 62.78 $\pm$ 0.31 \\
      & Full FT   & 0.87 $\pm$ 0.00 & 66.34 $\pm$ 0.77 & \textbf{69.05} $\pm$ \textbf{0.50} \\
      & Adapter   & 0.86 $\pm$ 0.00 & \underline{66.58} $\pm$ \underline{0.41} & 68.45 $\pm$ 0.16 \\
    \midrule
    \multirow{2}{*}{DOFA-B}
          & Frozen   & 0.89 $\pm$ 0.00 & 53.28 $\pm$ 0.82 & 57.92 $\pm$ 0.37 \\
      & Full FT  & 0.87 $\pm$ 0.01 & 61.09 $\pm$ 0.14 & 63.43 $\pm$ 0.72 \\
    \cmidrule(lr){1-5}
    \multirow{2}{*}{DOFA-L}
          & Frozen   & 0.88 $\pm$ 0.00 & 50.20 $\pm$ 0.47 & 58.26 $\pm$ 0.36 \\
      & Full FT  & 0.89 $\pm$ 0.01 & 61.15 $\pm$ 0.68 & 62.88 $\pm$ 0.34 \\
    \cmidrule(lr){1-5}
    \multirow{2}{*}{SpatSigma-B}
          & Frozen   & 0.88 $\pm$ 0.00 & 57.27 $\pm$ 0.24 & 59.15 $\pm$ 0.76 \\
      & Full FT  & 0.88 $\pm$ 0.00 & 62.91 $\pm$ 0.16 & 62.48 $\pm$ 0.29 \\
    \cmidrule(lr){1-5}
    \multirow{2}{*}{SpatSigma-L}
          & Frozen   & 0.87 $\pm$ 0.01 & 51.74 $\pm$ 0.54 & 56.72 $\pm$ 0.35 \\
      & Full FT  & 0.86 $\pm$ 0.01 & 62.90 $\pm$ 0.67 & 62.41 $\pm$ 0.35 \\
    \bottomrule
    \end{tabular}%
    \caption{Cross-sensor transferability experiments. We pretrain Spec. ViT-B and Spec. ViT-L with MAE on SpectralEarth and evaluate on three datasets from different sensors: Hyperview, EO1-CDL and DESIS-CDL. We compare against training from scratch, DOFA and HyperSigma. Best results are in \textbf{bold}, second best are \underline{underlined}.}    \label{tab:cross_sensor_transferability}
    \end{table*}

    \subsection{Ablation Studies}
    \label{sec:ablations}
        \paragraph{Impact of temporal positives} Temporal positives provide a natural data augmentation for joint-embedding methods and have demonstrated their benefits for multi-spectral imagery~\cite{wang2022ssl4eo,manas2021seasonal,stewart2024ssl4eo}. Given that only a subset of SpectralEarth includes temporal positives, it is unclear whether using temporal pairs significantly contributes to the overall performance of the pretrained models. To analyze their effect, we pretrain a Spec. RN50 using DINO with and without temporal pairs. Results in Table~\ref{tab:ablation_study_temporalr} confirm that excluding temporal positives degrades performance, with the effect most pronounced in frozen encoder evaluation.

        \begin{table}[ht]
            \small
            \centering
            \caption{Impact of temporal positives. A Spec. RN50 model is pre-trained on SpectralEarth with DINO and evaluated on EnMAP-CORINE and EnMAP-CDL.}
            \label{tab:ablation_study_temporalr}
            \begin{tabular}{@{}lcccc@{}}
            \toprule
            \textbf{Model} & \multicolumn{2}{c}{\textbf{EnMAP-CORINE (F1)}} & \multicolumn{2}{c}{\textbf{EnMAP-CDL (mIoU)}} \\
            \cmidrule(r){2-3} \cmidrule(l){4-5}
             & \textbf{Frozen} & \textbf{Fine-tune} & \textbf{Frozen} & \textbf{Fine-tune} \\
            \midrule
            w/ TP & $\mathbf{76.64}$ & $\mathbf{77.98}$ & $\mathbf{71.15}$ & $\mathbf{77.66}$ \\
            w/o TP & $74.85$ & $75.36$ & $68.42$ & $76.22$ \\
            \bottomrule
            \end{tabular}
        \end{table}

\begin{figure}[htbp]
            \centering
            \begin{subfigure}{0.49\linewidth}
                \includegraphics[width=\linewidth]{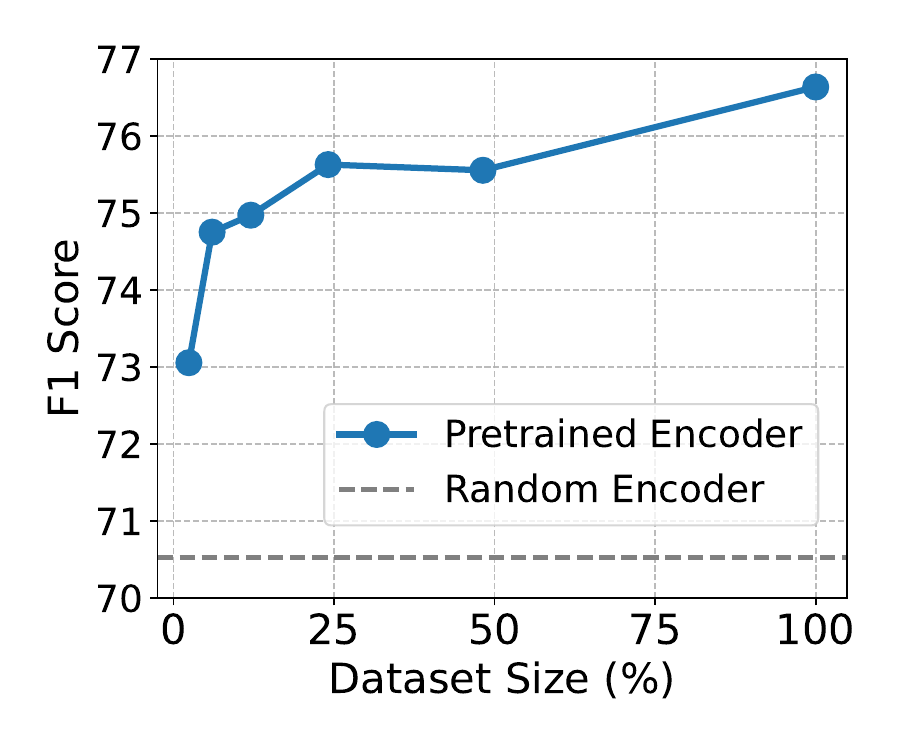}
                \caption{EnMAP-CORINE}
                \label{fig:ablation_pretraining_dataset_size_corine}
            \end{subfigure}
            \hfill 
            \begin{subfigure}{0.49\linewidth}
                \includegraphics[width=\linewidth]{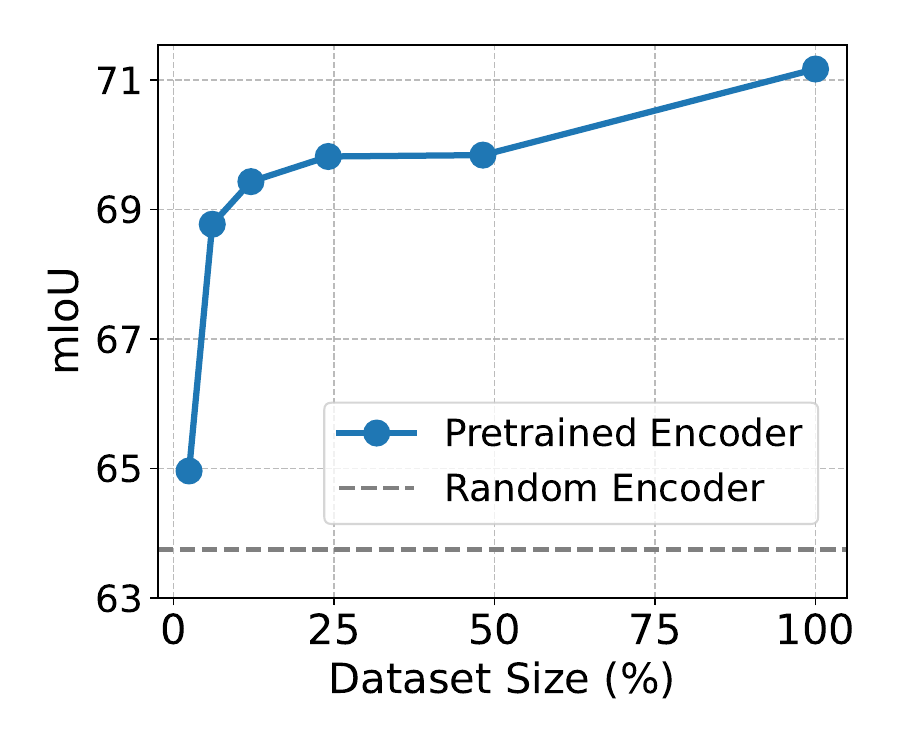}
                \caption{EnMAP-CDL}
                \label{fig:ablation_pretraining_dataset_size_cdl}
            \end{subfigure}
            \caption{Frozen encoder evaluation of Spec. RN50 on EnMAP-CORINE and EnMAP-CDL tasks with varying pre-training dataset sizes.}
            \label{fig:ablation_pretraining_dataset_size}
        \end{figure}

        \paragraph{How much data is needed for pre-training?} To answer this question, we explore the impact of dataset size by pre-training a Spec. RN50 with DINO on randomly sampled subsets of SpectralEarth with sizes 10K, 25K, 50K, 100K, and 200K locations. We conduct frozen-encoder evaluation on EnMAP-CORINE and EnMAP-CDL. Figure~\ref{fig:ablation_pretraining_dataset_size} summarizes our findings. We observe a consistent improvement in performance metrics as the scale of the pre-training dataset increases from 10k samples to the full size of SpectralEarth. This highlights the importance of large-scale datasets for hyperspectral self-supervised learning — even with small models such as Spec. RN50.

        We also compare SpectralEarth with HySpecNet-11k~\cite{fuchs2023hyspecnet} using the Spec. ViT-B backbone pre-trained with MAE. Table~\ref{tab:comparison_hyspec} reports performance under both frozen encoder and full fine-tuning settings. Pre-training on SpectralEarth consistently outperforms pre-training on HySpecNet-11k, with the largest gains observed in the frozen encoder evaluation. Under complete fine-tuning, the performance gap narrows, as the downstream training offset differences in pretraining.

        \begin{table}[ht]
            \small
            \centering
            \caption{Comparison of SpectralEarth and HySpecNet-11k. A Spec. ViT-B is pretrained using MAE and evaluated on EnMAP-CORINE and EnMAP-CDL.}
            \label{tab:comparison_hyspec}
            \begin{tabular}{@{}lcccc@{}}
            \toprule
            \textbf{Model} & \multicolumn{2}{c}{\textbf{EnMAP-CORINE (F1)}} & \multicolumn{2}{c}{\textbf{EnMAP-CDL (mIoU)}} \\
            \cmidrule(r){2-3} \cmidrule(l){4-5}
             & \textbf{Frozen} & \textbf{Fine-tune} & \textbf{Frozen} & \textbf{Fine-tune} \\
            \midrule
            HyspecNet-11k & $72.39$ & $78.20$ & $65.73$ & $75.20$ \\
            SpectralEarth & $\mathbf{74.98}$ & $\mathbf{79.65}$ & $\mathbf{72.42}$ & $\mathbf{77.46}$ \\
            \bottomrule
            \end{tabular}
        \end{table}
        
\begin{figure}[htbp]
            \centering
            \begin{subfigure}[b]{0.49\linewidth}
                \includegraphics[width=\linewidth]{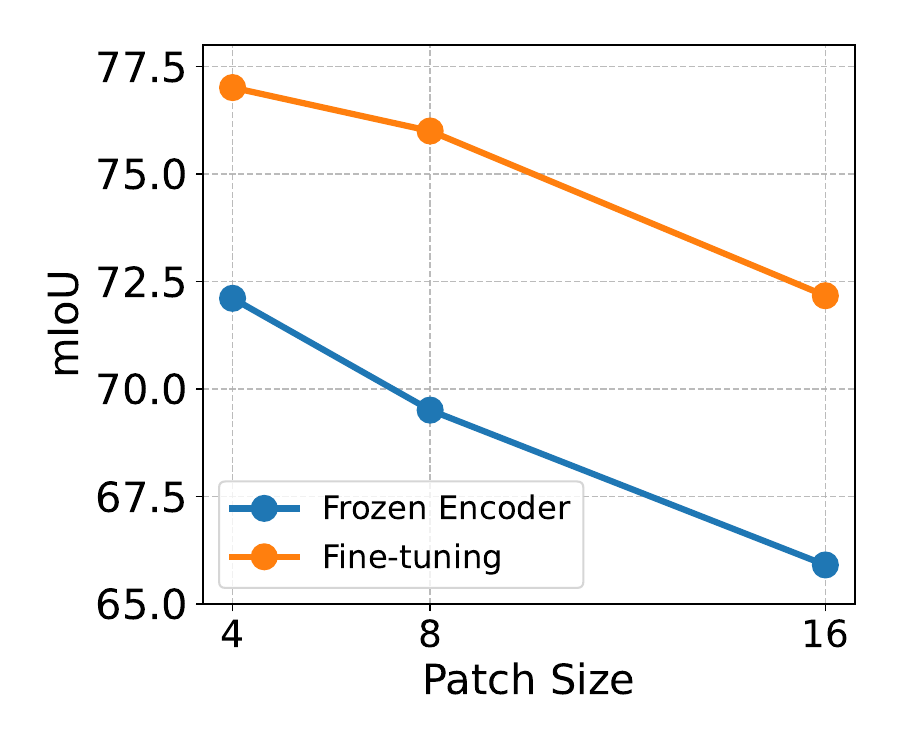}
                \caption{EnMAP-CDL}
                \label{fig:cdl_ablation_patch_size}
            \end{subfigure}
            \hfill 
            \begin{subfigure}[b]{0.49\linewidth}
                \includegraphics[width=\linewidth]{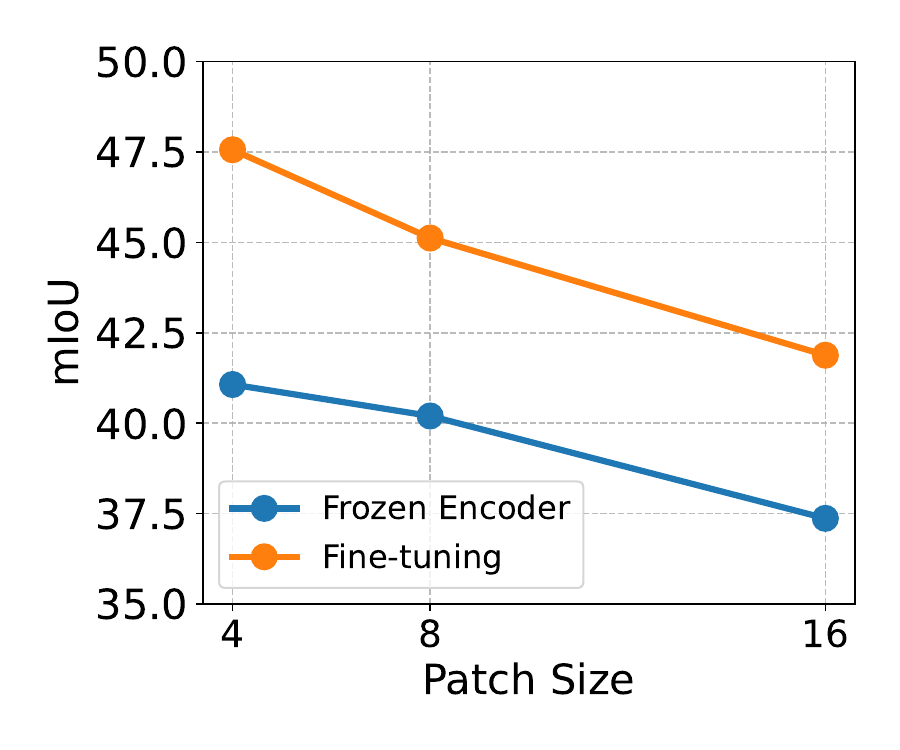}
                \caption{EnMAP-NLCD}
                \label{fig:nlcd_ablation_patch_size}
            \end{subfigure}
            \caption{Impact of the patch size for a pre-trained Spec. ViT-S evaluated on the EnMAP-CDL and EnMAP-NLCD datasets.}
            \label{fig:patch_size_ablation}
        \end{figure}

\begin{figure}[htbp]
        \centering
        \begin{tabular}{ccccc}
        \textbf{Image} & \textbf{Mask} & \textbf{4$\times$4} & \textbf{8$\times$8} & \textbf{16$\times$16} \\ 
        \includegraphics[width=0.155\linewidth]{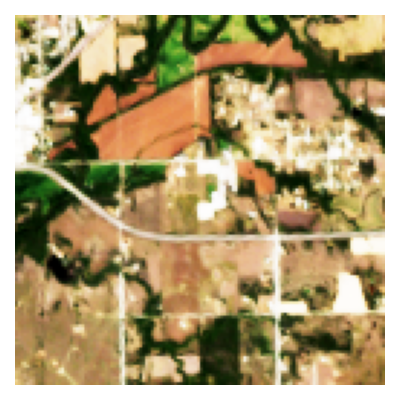} &
        \includegraphics[width=0.155\linewidth]{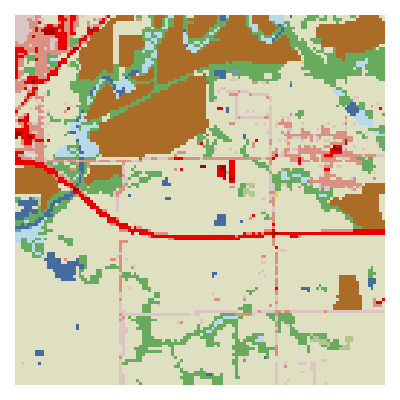} &
        \includegraphics[width=0.155\linewidth]{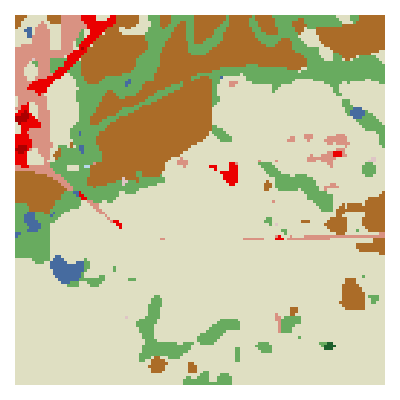} &
        \includegraphics[width=0.155\linewidth]{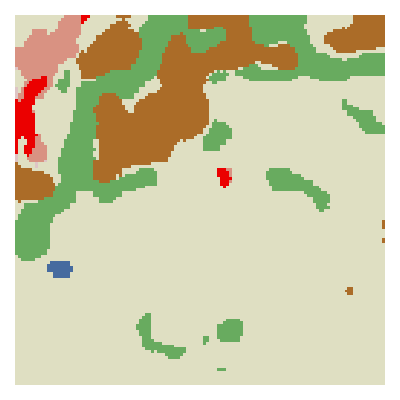} &
        \includegraphics[width=0.155\linewidth]{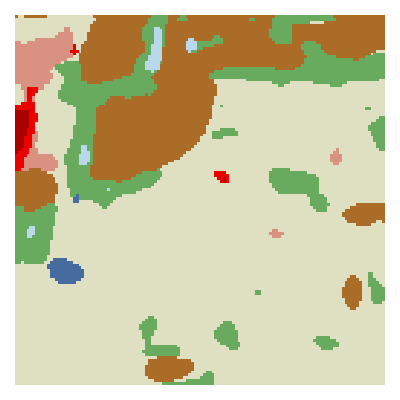} \\ 
        \includegraphics[width=0.155\linewidth]{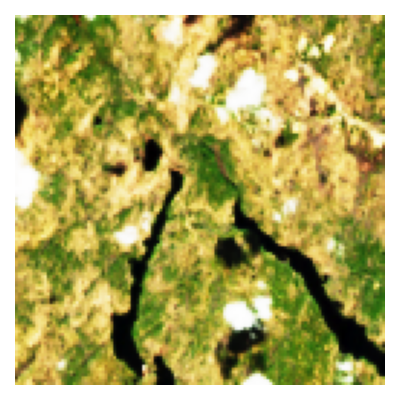} &
        \includegraphics[width=0.155\linewidth]{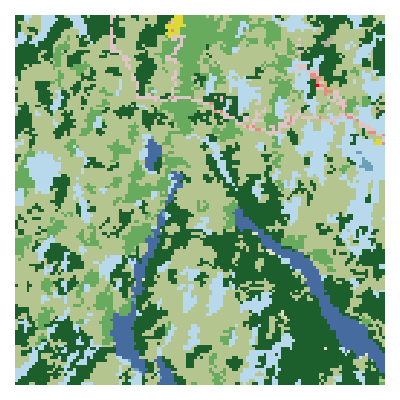} &
        \includegraphics[width=0.155\linewidth]{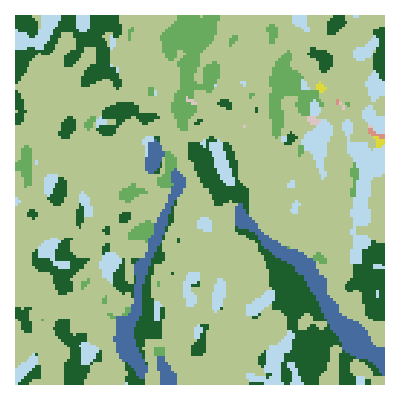} &
        \includegraphics[width=0.155\linewidth]{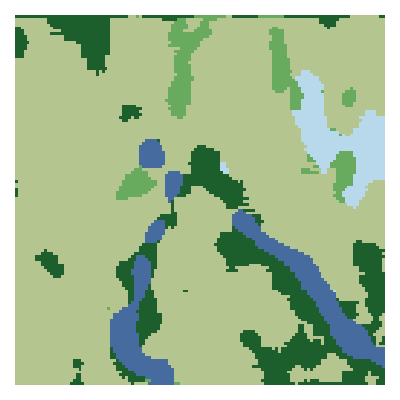} &
        \includegraphics[width=0.155\linewidth]{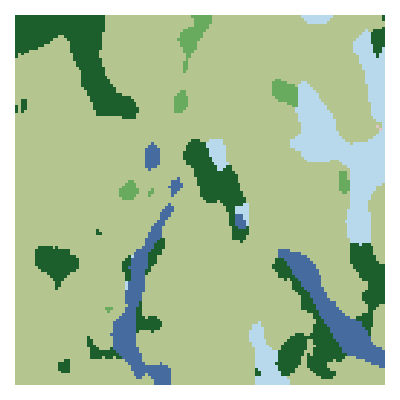} \\ 
        \includegraphics[width=0.155\linewidth]{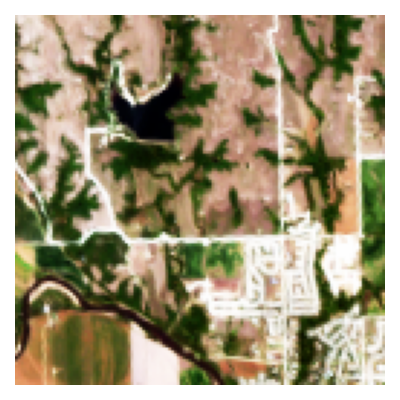} &
        \includegraphics[width=0.155\linewidth]{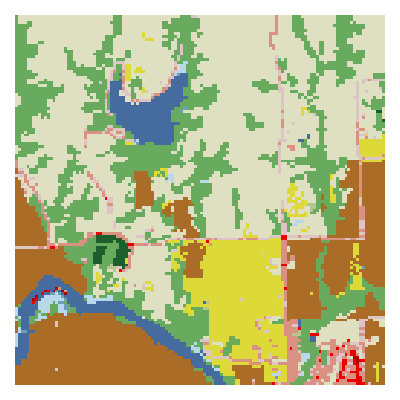} &
        \includegraphics[width=0.155\linewidth]{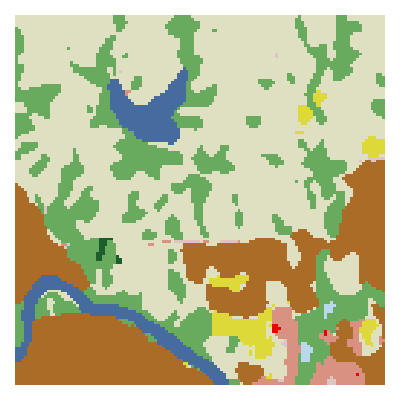} &
        \includegraphics[width=0.155\linewidth]{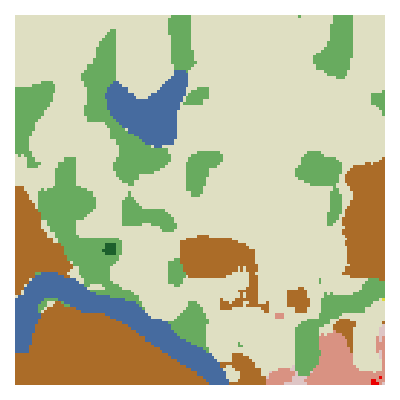} &
        \includegraphics[width=0.155\linewidth]{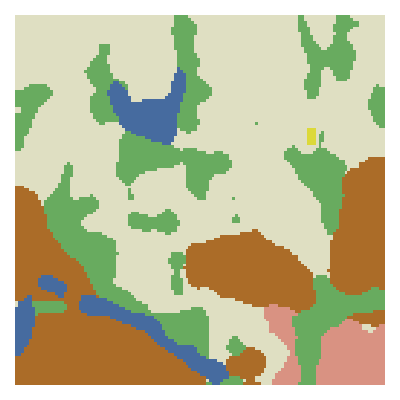} \\
        \includegraphics[width=0.155\linewidth]{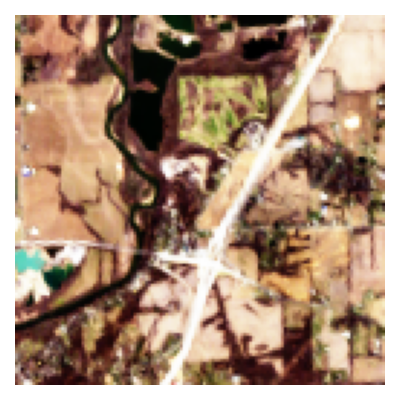} &
        \includegraphics[width=0.155\linewidth]{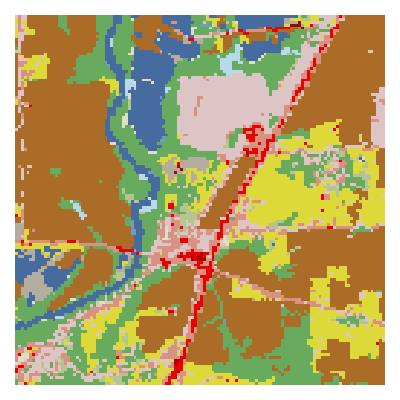} &
        \includegraphics[width=0.155\linewidth]{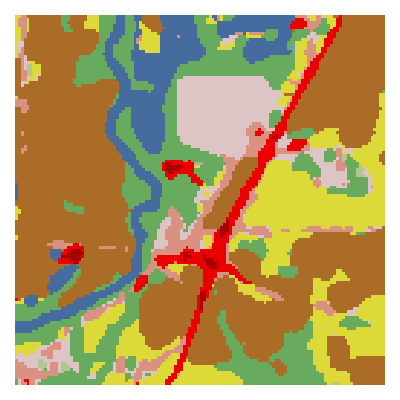} &
        \includegraphics[width=0.155\linewidth]{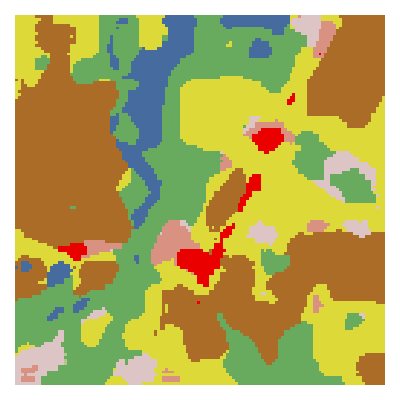} &
        \includegraphics[width=0.155\linewidth]{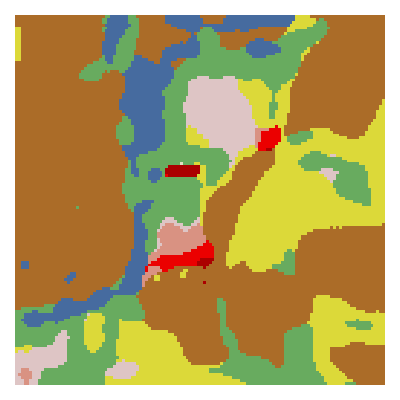}
        \end{tabular}
        \caption{Impact of the patch size in the Spec. ViT-S model on EnMAP-NLCD. From left to right respectively: RGB visualization of the input image, ground truth (GT) mask, and predictions with 4$\times$4, 8$\times$8, and 16$\times$16 patch sizes.}
        \label{fig:ablation_qualitative}

        \end{figure}

        \paragraph{Importance of the patch size} Medium-resolution remote sensing imagery requires more fine-grained detail preservation than natural imagery, especially for segmentation tasks. This aspect is sometimes overlooked in the literature, where vision transformers with a patch size of 16$\times$16 or 8$\times$8 are commonly used as backbones to pretrain foundation models to reduce the computational cost ~\cite{jakubik2023foundationmodelsgeneralistgeospatial,cong2022satmae,xiong2024neural,wang2024hypersigmahyperspectralintelligencecomprehension}. We argue that a smaller image size with a smaller patch size can lead better trade-offs, particularly for HSI imagery where pixel information is rich and should not be heavily compressed in the embedding space.
        To analyze the impact of the patch size, we pre-trained Spec. ViT-S using MAE with patch sizes of 4$\times$4, 8$\times$8 and 16$\times$16. Results in Figure~\ref{fig:patch_size_ablation} show that smaller patches consistently improve performance for segmentation tasks. This is further supported by qualitative results in Figure~\ref{fig:ablation_qualitative}, where models with smaller patches produce more refined segmentation maps. These results are consistent with the performance gap between ViT-based foundation models and ConvNet baselines observed in previous studies \cite{xie2024foundation}. In practice, using additional trainable parameters to extract high-resolution features in combination with a pre-trained ViT can mitigate this issue.

\begin{figure}[htbp]
            \centering
            \begin{subfigure}[b]{0.49\linewidth}
                \includegraphics[width=\linewidth]{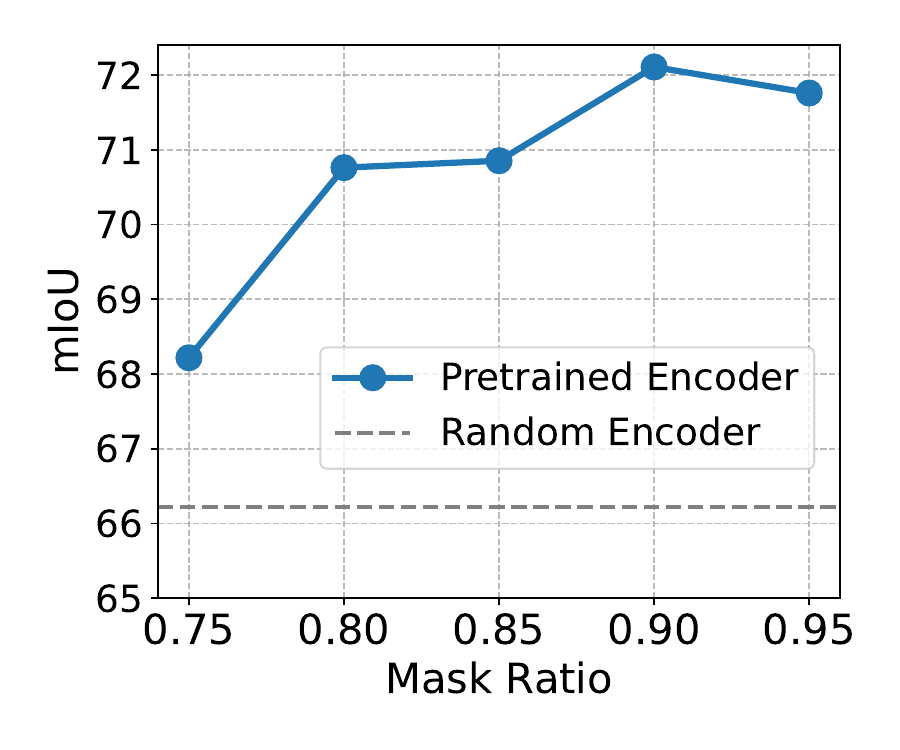}
                \caption{EnMAP-CDL}
                \label{fig:cdl_ablation_mask_ratio}
            \end{subfigure}
            \hfill 
            \begin{subfigure}[b]{0.49\linewidth}
                \includegraphics[width=\linewidth]{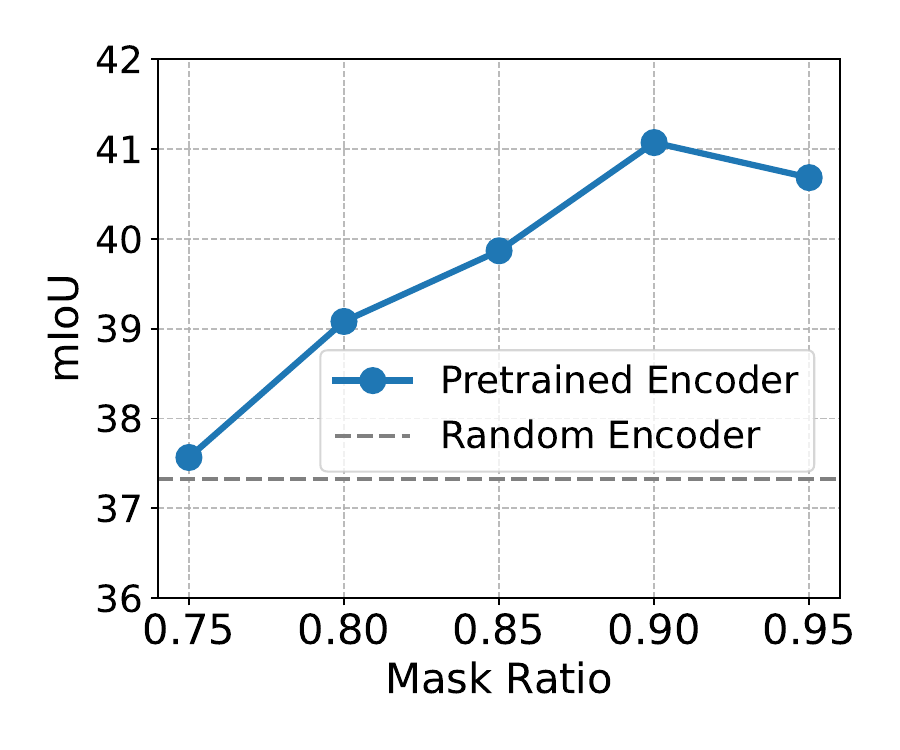}
                \caption{EnMAP-NLCD}
                \label{fig:nlcd_ablation_mask_ratio}
            \end{subfigure}
            \caption{Impact of the masking ratio in MAE evaluated on EnMAP-CDL and EnMAP-NLCD for the Spec. ViT-S model.}
            \label{fig:ablation_mask_ratio}
        \end{figure}

        \paragraph{Masking ratio in MAE} Given our small 4$\times$4 patch size for Spec. ViT models, we expect a higher masking ratio is needed for MAE compared to the usual $75\%$ used in computer vision~\cite{he2022masked}. To confirm this hypothesis, we pre-train Spec. ViT-S using MAE with different masking ratios, from $75\%$ to $95\%$. Given MAE's poor linear probing results for the classification task, we focus on the segmentation downstream tasks. Results in Figure \ref{fig:ablation_mask_ratio} show that we obtain good representations even with an aggressive $95\%$ masking, which enables faster pre-training as fewer tokens need to go through the encoder.  

        \begin{table}[ht]
            \small
            \centering
            \caption{Impact of the spectral adapter in the Spec. RN50 architecture pre-trained on SpectralEarth with MoCo-V2.}
            \label{tab:ablation_study_spec_adapter}
            \begin{tabular}{@{}lcccc@{}}
            \toprule
            \textbf{Model} & \multicolumn{2}{c}{\textbf{EnMAP-CORINE (F1)}} & \multicolumn{2}{c}{\textbf{EnMAP-CDL (mIoU)}} \\
            \cmidrule(r){2-3} \cmidrule(l){4-5}
             & \textbf{Frozen} & \textbf{Fine-tune} & \textbf{Frozen} & \textbf{Fine-tune} \\
            \midrule
            w/o Adapter & $72.82$ & $77.10$ & $67.50$ & $74.31$ \\
            w/ Adapter & $\mathbf{73.97}$ & $\mathbf{78.57}$ & $\mathbf{71.15}$ & $\mathbf{77.66}$ \\
            \bottomrule
            \end{tabular}
        \end{table}

        \paragraph{Impact of the spectral adapter}
        To assess the impact of the spectral adapter, we compare the classical ResNet-50 with our modified architecture, Spec. RN50. We adjust the first convolutional layer of the ResNet-50 model to accommodate the 202 input bands. Both models are pretrained with MoCo-V2 and evaluated on the EnMAP-CORINE and EnMAP-CDL datasets. The results are summarized in Table~\ref{tab:ablation_study_spec_adapter}.
        Spec. RN50 consistently outperforms the standard ResNet-50 in both frozen encoder evaluation and full fine-tuning. This result confirms the importance of a dedicated spectral feature extractor when dealing with hyperspectral images, particularly for tasks requiring fine-grained spectral discrimination.

%% file: sec/7_conclusion.tex
\section{Conclusion}
\label{sec:conclusion}

We introduce SpectralEarth, a large-scale dataset derived from EnMAP imagery as a basis for SSL methodologies in the hyperspectral domain. We leveraged SpectralEarth to pre-train foundation models based on popular SSL algorithms. Extensive evaluation on several downstream tasks benchmarked the pre-trained models. Our results demonstrate the effectiveness of models pre-trained on SpectralEarth to improve performance and reduce computational costs for downstream applications. We believe that the insights derived from this study serve as a basis for future development in SSL for hyperspectral imagery.  

As it stands, we encourage further effort to construct additional hyperspectral benchmark datasets. Covering tasks such as unmixing enables a more comprehensive evaluation of hyperspectral foundation models. In particular, transferability across different hyperspectral sensors remains a challenging problem.
Moreover, exploration of a wider range of architectures and SSL algorithms, beyond the Spectral ResNet-50 and ViT models employed in this work, may provide a more comprehensive toolbox for practical deep learning on HSI. Last but not least, exploration and benchmarking of multi-sensor SSL methods is a promising avenue of research toward more general geospatial foundation models.